\documentclass{article}
\usepackage[final, main]{neurips_2026}
\usepackage[utf8]{inputenc}
\usepackage[T1]{fontenc}
\usepackage{hyperref}
\usepackage{url}
\usepackage{booktabs}
\usepackage{amsfonts}
\usepackage{nicefrac}
\usepackage{microtype}
\usepackage{xcolor}
\usepackage{physics}
\usepackage{framed}
\usepackage{graphicx}
\graphicspath{{./}{../}}
\usepackage{subcaption}
\usepackage[bbgreekl]{mathbbol}
\usepackage{amsmath}
\usepackage{amssymb}
\usepackage{mathtools}
\usepackage{bm}
\usepackage{amsthm}
\usepackage{enumitem}
\usepackage{wrapfig}
\usepackage{multirow}
\usepackage[capitalize,noabbrev]{cleveref}
\usepackage[textsize=tiny]{todonotes}

\theoremstyle{plain}

\theoremstyle{definition}

\theoremstyle{remark}

\newcommand{\E}{\mathbb{E}}

\title{An Analytical Model of Compute-limited Multistage Training Pipelines}

\author{%
  Nishil Patel \\
  Gatsby Computational Neuroscience Unit \\
  University College London \\
  \texttt{nishil.patel@ucl.ac.uk}
  \And
  Jin Hwa Lee \\
  Sainsbury Wellcome Centre \\
  University College London
  \AND
  Basile Confavreux \\
  Gatsby Computational Neuroscience Unit \\
  University College London
  \And
  Andrew Saxe \\
  Gatsby Computational Neuroscience Unit \\
  Sainsbury Wellcome Centre \\
  University College London
}

\begin{document}

\maketitle

\begin{abstract}

Modern foundation models are trained through multiple stages, yet how these stages interact and how compute should be allocated across them remain poorly understood. We introduce an analytically tractable high-dimensional model combining unsupervised pretraining with a linear autoencoder, supervised fine-tuning (SFT) with noisy stepwise labels, and reinforcement learning (RL) with sparse episodic rewards. Using random matrix theory and an operator-projection closure, we derive approximate low-dimensional learning dynamics that closely match simulations. We show that finite pretraining data imposes a representation-dependent performance limit, with Bayes classification error scaling as the inverse square root of the pretraining sample ratio, and observe the same exponent in asymptotic SFT and RL errors. Better pretraining also reduces sensitivity to noisy supervision. SFT initialization shortens the initial RL plateau induced by sparse rewards, an effect we also observe in OLMo-2 1B on GSM8K. Longer SFT warm starts additionally improve asymptotic RL performance. Optimizing compute across all three stages reveals that the optimal allocation depends on supervision quality and total budget, with a non-monotonic dependence of the RL compute share on budget. Finally, a complementary compositional model shows how RL selectively strengthens task-relevant pretrained skills while largely preserving unused skills. Together, these results identify mechanisms linking representation quality, feedback structure, and compute allocation in multistage training.

\end{abstract}

\section{Introduction}

Modern foundation models exhibit transformative capabilities from human-level language comprehension and production~\cite{radford2019language,brown2020language, touvron2023llama}, to reasoning across modalities and mathematical problem-solving~\cite{jaech2024openai, shao2024deepseekmath,chervonyi2025gold,team2025gemini}. Multistage training is a critical aspect of the training paradigm of such models, which consists of unsupervised auto-regressive pretraining followed by several stages of supervised finetuning and reinforcement learning on smaller, domain-specific datasets~\cite{ouyang2022training, achiam2023gpt, liu2024deepseek,yang2025qwen3,deepseekai2025deepseekr1}. Yet, the consequences of different training stages and their interactions remain poorly understood. Empirical investigation of these questions is difficult, as it requires substantial compute and costly human annotation, constraining the scope of hypotheses that can be explored~\cite{zhao2025echo, shao2025spurious, akter2025front, liu2025prorl}.

Alongside the transition to multistage training pipelines, foundation models reflect a second shift from a sample-complexity–centric view of learning to a compute-centric one. Whereas classical theory treats data as the limiting resource and implicitly assumes unbounded compute, modern foundation models invert this assumption: data is abundant, compute is limited, and performance scales predictably with model size and training compute, as evidenced by large-scale empirical scaling laws~\cite{kaplan2020scaling, hoffmann2022training}. In this compute-limited regime, models are rarely trained to convergence, and performance is instead determined by how a fixed computational budget is allocated across model capacity and multiple training stages.

Together, these two shifts---toward multistage pipelines and the compute-limited regime---pose a challenge for existing theory, which offers limited guidance for analyzing learning dynamics, performance trade-offs, or compute allocation in modern foundation model training. ---

\paragraph{Contributions.}
---To this end, we develop an analytically tractable high-dimensional model based on statistical physics~\cite{seung1992statistical,SaadSolla,zdeborova2016statistical}. The model isolates the effects of pretrained representation quality, dense stepwise supervision, and sparse episodic rewards. We derive deterministic learning dynamics in the high-dimensional limit and obtain a tractable, approximately closed system of ODEs, allowing us to study how pretrained
representations, supervision quality, and compute allocation jointly
determine downstream performance. Our contributions are:
\begin{itemize}
    \item We combine random matrix theory with an operator-projection
    closure to obtain a low-dimensional approximation of the SFT and
    RL dynamics, validated against simulations. We derive the
    $\alpha^{-1/2}$ scaling of the representation-dependent Bayes
    error with pretraining sample size and observe the same exponent
    in asymptotic finetuning error.

    \item We characterize how SFT and RL complement each other.
    SFT shortens the initial sparse-reward plateau, an effect also
    observed in OLMo-2 1B on GSM8K. Under spherical normalization
    and locally optimal learning rates, longer SFT warm starts
    additionally improve asymptotic RL performance. Better
    pretraining also reduces the excess SFT error caused by
    label noise.

    \item We optimize compute allocation across pretraining, SFT,
    and RL. The optimal split depends on both supervision quality
    and the total budget: noisier labels favour RL, while the
    optimal RL share can vary non-monotonically with total compute.

    \item We extend the model to compositional tasks, showing how
    prior skill proficiency accelerates RL and how finetuning
    selectively improves task-relevant skills while largely
    preserving unused skills.
\end{itemize}

%Concretely, our contributions are:
%\begin{enumerate}[itemsep=.2mm]
    %\item Constructing an analytically tractable model of pretraining, SFT, and RL from bulk feedback, allowing us to derive exact ODEs for learning dynamics;
    %\item Studying how different design choices, such as finetuning strategies, data quality and pretraining level, impact learning dynamics and performance of the model;
    %\item Analyzing the optimal multistage training design given a limited compute budget.
%\end{enumerate}

\section{Related Work}
\label{section:related_work}

\paragraph{Empirical interactions} across training stages. Controlled studies have investigated how pretraining shapes the effectiveness of downstream reinforcement learning. \cite{shen2026understanding} examine a complete pretraining-–SFT-–RL pipeline and relate pretraining loss and data scale to post-RL performance and learning speed, including a pretraining–-RL compute frontier. \cite{zhang2026interplay} isolate the effects of pretraining, intermediate training, and RL on compositional reasoning tasks. \cite{zhao2024can} investigate the amplification of pretrained behaviors under RL. Empirical work compares the generalization behavior of SFT and outcome-based RL \citep{chu2025sft} and demonstrates that higher standalone SFT accuracy need not yield better subsequent RL performance \citep{zhang2026goodsft, kang2026quagmires}.

\paragraph{Closure of structured-data learning dynamics.}
The problem of anisotropic data covariance generating an infinite hierarchy of order parameters has been studied in the context of online learning with structured data \cite{yoshida2019data}. In that work, the authors show that the ODE system can only be closed if the covariance has a finite number of unique eigenvalues, where there is a pair of order parameters associated to every unique eigenvalue. There are several works that address this feature: approaches include replacing the infinite discrete hierarchy by a closed functional object --- a spectral order parameter \citep{goldt2020modeling, goldt2022gaussian} or a resolvent \citep{collins2024hitting}. However, this is non-trivial to implement in the case of a random data covariance. Another more recent approach most similar in spirit to ours is \cite{xu2026high}, where the authors show the infinite hierarchy can be exponentially well approximated by a finite truncated system. Ours is merely a different method of truncation, without a proof of exponential convergence.

\begin{figure*}[t!]
    \centering
        \includegraphics[width=\linewidth]{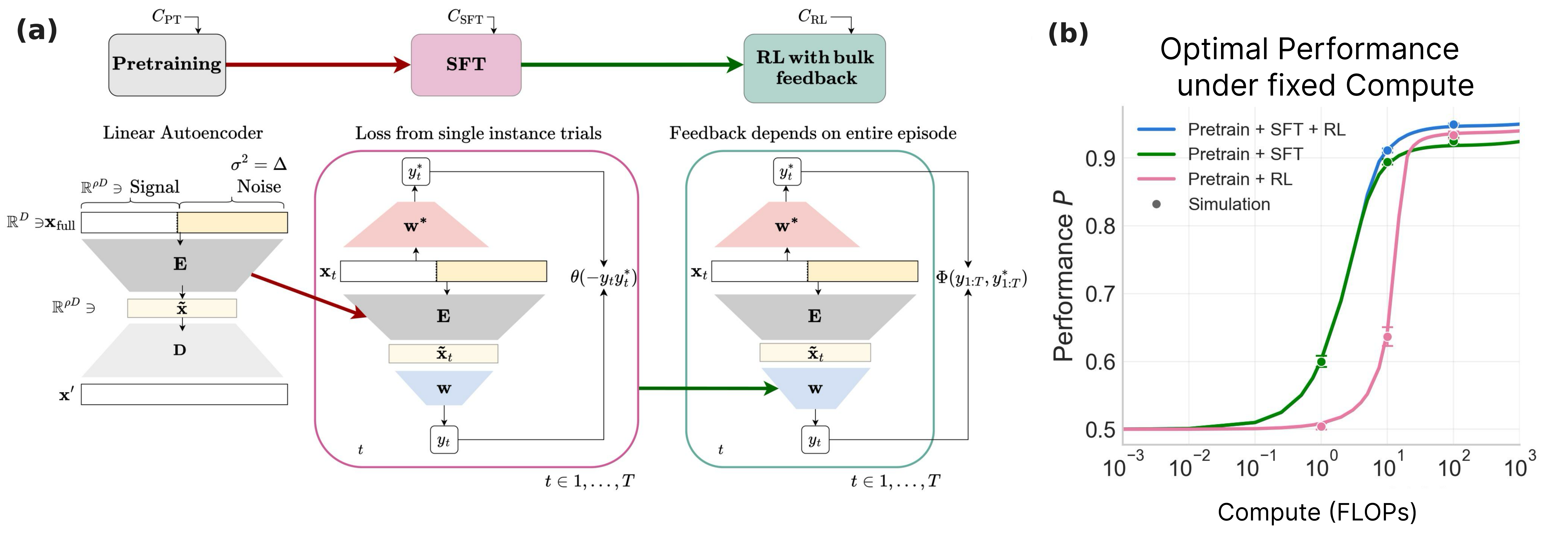}
    \vspace{-3mm}
    \caption{(a) \textbf{A Schematic of multistage training pipeline.} Each solid trapezoid indicates a single linear neural network layer. Pretraining is a Linear Autoencoder (LAE) that approximately learns the principal subspace of data $\mathbf{x} \sim \mathcal{N}(0, \Sigma)$, where $\Sigma$ is equal to the identity apart from a noise subspace, with identity covariance scaled by $\Delta < 1$. $C_\text{PT}$ is the compute allocated to pretraining. In the subsequent finetuning (SFT and RL), a teacher network $\mathbf{w}^*$ gives labels $y^*$ from the signal subspace and it is used to train a student network $\mathbf{w}$ learning in the subspace of the LAE encoder projection with feedback. The student predicts label $y$ using the effective features $\tilde{\mathbf{x}}$. Importantly, SFT is modelled by every observation at each $t$ updating the student, while RL is modelled by bulk feedback through reward at the end of a `correct' sequence through the entire $t=1,\dots,T$ episode. We denote the compute allocations to SFT and RL, $C_\text{SFT}$ and $C_\text{RL}$ respectively. (b) Optimal generalisation performance against total Compute flops under three protocols: SFT only, RL only, and SFT with RL. Compute is allocated as described in \ref{sec:budget_allocation}, parameters are set as in \ref{fig:5}d.}
    \label{fig:1}
     \vspace{-5mm}
\end{figure*}

\section{Setup}
\label{section:model_setup}

\subsection*{Pretraining}
\label{sec:pretraining}

We draw a fixed pretraining dataset
$\mathbf X\in\mathbb R^{N\times D}$ with independent rows distributed as
$\mathcal N(0,\mathbf\Sigma)$, where
$\mathbf\Sigma=\mathbf M_s+\Delta\mathbf M_n$,
$\mathbf M_s$ projects onto a signal subspace of dimension $\rho D$, and
$\mathbf M_n=\mathbf I-\mathbf M_s$.
We take $0<\rho<1$, $0<\Delta<1$, and the high-dimensional limit
$N,D\to\infty$ at fixed $\alpha=N/D > \rho$.
A linear autoencoder (LAE) with hidden width $\rho D$ learns the empirical
principal subspace of this dataset when trained to convergence
\citep{baldi1989neural,saxe2019information}.
We represent its learned subspace by the orthogonal projector $\mathbf P$.
Downstream observations are $\tilde{\mathbf x}=\mathbf P\mathbf x$,
with fresh $\mathbf x\sim\mathcal N(0,\mathbf\Sigma)$. $\mathbf P$ remains fixed throughout finetuning. Thus $\alpha$ determines how much task-relevant signal is available for the downstream task. The LAE learns a compressed feature subspace from \textit{unlabelled} data, abstracting the role of pretraining as representation learning: discovering reusable structure and organizing inputs into a feature space~\citep{bengio2012representation,lecun2015deep,lecun2022path,bommasani2021opportunities}. This frozen-feature assumption is motivated by evidence that finetuning preserves pretrained representations~\citep{nadipalli2025layer} and the success of PEFT~\citep{hu2022lora,han2024parameterefficientfinetuninglargemodels}

\begin{figure*}[t]
    \centering
        \includegraphics[width=\linewidth]{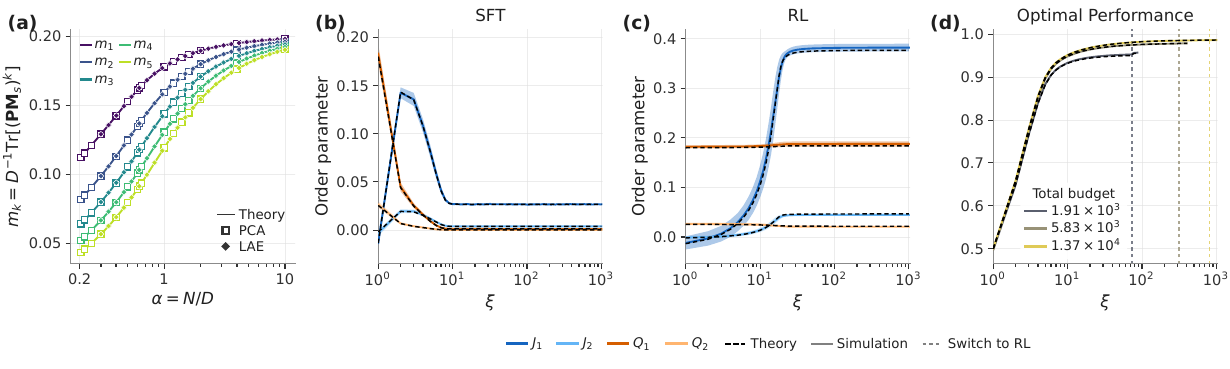}
    \caption{\textbf{ODEs accurately capture distinct learning dynamics of SFT and RL.} (a) Good agreement of normalised traces of powers of projection-signal overlap matrices for: RMT (solid), PCA (open square), and LAE (filled diamond). Good agreement of ODE theory (dashed) in time ($\xi$) and simulated evolution (solid) of order parameters for: (b) the SFT model, (c) the RL model. (d) The evolution of probability of correct classification ($P$) under optimal compute budget allocation for a range of total compute budgets in FLOPs. \textit{Parameters}: $\Delta=0.1$, $\rho=0.2$. (a) $D=2000$. (b) $\epsilon=0.05$, spherically unconstrained weights, $\eta=1$. (c) spherically constrained weights, locally optimal learning rate. (b), (c) and (d) $T=6$, $D=1000$. (d) $\epsilon=0.2$. }
    \vspace{-.5cm}
    \label{fig:2}
\end{figure*}

\subsection*{Finetuning}
\label{sec:Finetuning}

The finetuning task is organized in episodes of length $T$, where a \textit{student} perceptron $\mathbf w \in \mathrm{Range}(\mathbf P)$ makes a series of binary decisions $y_t = \text{sgn}\left(\frac{\mathbf{\tilde{x}_t}^{\intercal}\mathbf{w}}{\sqrt{D}}\right)$ given the raw observations $\mathbf{x}_{1\dots T} \sim \mathcal{N}(\mathbf 0, \mathbf \Sigma)$ projected onto the encoder subspace by $\mathbf P$. Data labels for supervision/reward $y_t^* = \text{sgn}\left(\frac{\mathbf{x}^{\intercal}\mathbf{w^*}}{\sqrt{D}}\right)$ are specified by a \textit{teacher} perceptron $\mathbf w^* \in R^{D}$, which has access to the raw data $\mathbf x$, importantly, it sees the entire signal subspace. The teacher is a fixed-weight oracle that generates the target labels. During SFT and RL, the weights of the student are updated with feedback from the teacher, but the core difference lies in how the feedback from the output is given to the student. This setting is inspired by the teacher--student framework~\citep{gardner1989three, seung1992statistical, biehl1993learning}. We discuss related teacher--student models in Appendix~\ref{app:teacher_student}.

\paragraph{SFT:}
Each individual sample in the episode contributes separately to the update by error feedback with the corresponding teacher label (\cref{fig:1}a, pink box). The update follows the Perceptron rule ~\citep{rosenblatt1958perceptron}.
The update at the $\mu_{\text{th}}$ episode can be written as

\begin{align}
    \mathbf{w}^{(\mu + 1)} = \mathbf{w}^{(\mu)} -  \frac{\eta}{T\sqrt{D}}\sum_{t=1}^T y_t^{(\mu)}
    \mathbf{\tilde{x}}_t^{(\mu)} \theta(-{y^*_{\mathrm{SFT}}}_t^{(\mu)}y_t^{(\mu)}),
\label{eq:sft_update}
\end{align}

where $\theta(\cdot)$ is the heaviside step function and $\eta$ is the learning rate. This can be interpreted as the linearized gradient descent update on the mean squared loss $\Sigma_{t=1}^T (y_t - {y^*_{\mathrm{SFT}}}_t)^2$. The teacher labels for SFT are generated with noise, i.e. ${y^*_{\mathrm{SFT}}}_t = y^*_tz_t$, where $z_t$ takes the values of 1 or -1 with probability $(1-\epsilon)$ or $\epsilon$ respectively. This is a simple model of imperfect supervision, which is common in real-world datasets. N.B. this label noise is unique to SFT in our model. Importantly, the student receives an individual update for every $t$. For example, in multi-step math tasks, this is equivalent to getting a supervised learning signal at \textit{every step} of a solution.

\paragraph{RL:} We use the RL perceptron update of \citet{RLperceptron}, which extends the teacher-student framework to online policy learning under a sequential decision-making task. Given the same finetuning task, the critical difference with SFT lies in the bulk feedback: the student is updated using a delayed, sparse reward signal at the end of the episode, rather than at every step, as in SFT. The update at $\mu_{\text{th}}$ episode can be written as
\begin{align}
    \mathbf{w}^{(\mu + 1)} = \mathbf{w}^{(\mu)} +  \frac{\eta}{T}\left( \sum_{t=1}^T \frac{y_t^{(\mu)}
    \mathbf{\tilde{x}}_t^{(\mu)}}{\sqrt{D}} \right) R(y_{1:T}^{(\mu)}, {y^*}^{(\mu)}_{1:T}),
\label{eq:rl_update}
\end{align}
where the reward rule is specified by $R(y_{1:T}, y^*_{1:T})$. In this work, we mainly use the reward condition, \textit{every action in an episode must be correct}, (i.e $y_t = y^*_t,~ \forall t \in 1,\dots T$, or $R = \Pi_{t=1}^T\theta(y_ty_t^*)$). This is akin to RLVR  where reward is 
received based on the correctness of a final result, where the final result requires multiple previous steps to be correct in order to reach it.
The episode length $T$ controls the difficulty of the task. Larger values of $T$ make the task more difficult as the model needs to make a longer sequence of correct decisions, giving an idealized analogue of verifiable-reward methods such as GRPO~\citep{deepseekai2025deepseekr1}. This task structure reflects the nature of multi-step problems, for example, a math problem requiring multiple steps of reasoning to be correct.
When we consider multiple stages of finetuning (e.g. SFT followed by RL), each stage trains the same student and uses the final student state of the previous stage as the initial student state of the subsequent stage.

We use it not as a literal model of full finetuning, but to isolate how representation quality constrains later learning.
\section{Learning Dynamics of SFT and RL}
\label{subsec:dynamics}
We can derive a set of ordinary differential equations (ODEs) that describe the learning dynamics of our model using the approach of~\citeauthor{SaadSolla}. In the thermodynamic limit ($D \rightarrow \infty$), the stochastic evolution of the high-dimensional student weights reduces to the deterministic evolution of a set of low-dimensional \textit{order parameters}:

\begin{align}
    J_{s/n} = \frac{\mathbf{w}^\top \mathbf{P} \mathbf{M}_{s/n}\mathbf{P}\mathbf{w}^*}{D},~Q_{s/n} = \frac{\mathbf{w}^\top  \mathbf{P}\mathbf{M}_{s/n}  \mathbf{P}\mathbf{w}}{D},~S_{s/n} = \frac{\mathbf{w^*}^\top  \mathbf{P}\mathbf{M}_{s/n}  \mathbf{P}\mathbf{w^*}}{D}~S = \frac{\mathbf{w^*}^\top \mathbf{w^*}}{D}
    \label{eq:order_params}
\end{align}

Intuitively, $Q_{s/n}$ is the size of the projected student in the signal/noise subspace, $J_{s/n}$ is the overlap between the projected student and the projected teacher in signal/noise subspace, and $(S_{s/n}, S)$ are constant throughout training. 

By appropriate multiplication of \ref{eq:sft_update} and \ref{eq:rl_update}, we can derive a set of ODEs that govern the evolution of order parameters in the thermodynamic limit.

For the SFT:
\begin{equation}
\begin{aligned}
    \frac{dJ_{s/n}}{d\xi} &= -\eta \left \langle y \frac{{\mathbf{w}^*}^\top \mathbf{P} \mathbf{M}_{s/n}\mathbf{P} \mathbf{x}}{\sqrt{D}} \theta(-yy^*) \right \rangle, \\
    \frac{dQ_{s/n}}{d\xi} &= -2 \eta \left \langle y \frac{{\mathbf{w}}^\top\mathbf{P} \mathbf{M}_{s/n}\mathbf{P} \mathbf{x}}{\sqrt{D}}  \theta(-yy^*) \right \rangle + \frac{\eta^2}{T}\frac{\text{Tr}(\mathbf{P} \mathbf{M}_{s/n}\mathbf{P} \Sigma)}{D} \langle \theta(-yy^*) \rangle, 
\end{aligned}
\label{eq:sft_ode}
\end{equation}

For the RL:
\begin{equation}
\begin{aligned}
    \frac{dJ_{s/n}}{d\xi} &= \frac{\eta}{T} \left \langle \sum_{t=1}^T y_t \frac{{\mathbf{w}^*}^\top \mathbf{P} \mathbf{M}_{s/n}\mathbf{P} \mathbf{x}_t}{\sqrt{D}} R \right \rangle, \\
    \frac{dQ_{s/n}}{d\xi} &= \frac{2 \eta}{T} \left \langle \sum_{t=1}^T y_t \frac{{\mathbf{w}}^\top\mathbf{P} \mathbf{M}_{s/n}\mathbf{P} \mathbf{x}_t}{\sqrt{D}}R \right \rangle + \frac{\eta^2}{T}\frac{\text{Tr}(\mathbf{P} \mathbf{M}_{s/n}\mathbf{P} \Sigma)}{D} \langle R \rangle.
\end{aligned}
\label{eq:rl_ode}
\end{equation}

Here, $\xi$ denotes continuous training time and the angled bracket expectations are taken over the online samples $\mathbf{x}$, and can be approximated in terms of the order parameters, allowing approximate closure (see \cref{sec:closing_odes}). The trace term is self-averaging in the proportional $D, N \rightarrow \infty$ limit, and can be expressed self-consistently (see \cref{App:TheoryProjected}). The expected generalization ability $P$ (i.e. the probability of a single decision being correct) can be expressed in terms of the order parameters:

\begin{align}
      P = \frac{1}{2} + \frac{1}{\pi} \text{sin}^{-1}\left( \frac{J_s + J_n}{\sqrt{S(Q_s + \Delta Q_n)}} \right ),
\end{align}

allowing us to compute generalization performance at any point during training. See~\cref{App:TheoryProjected} for the full derivation and explicit calculation of the expectations and traces in \cref{eq:sft_ode} and \cref{eq:rl_ode}. In~\cref{fig:2}, we see that the evolution of $P$ found from the numerical integration of the ODEs (dashed) shows an excellent match with the simulation of the updates (solid). We consider both unconstrained dynamics and a spherical constraint fixing
$Q_s+Q_n$. Where specified, the learning rate is chosen to maximize the
instantaneous predicted increase in clean generalization probability,
using the rule in \cref{App:LocallyOptimalLR}.

\subsection{Closing the ODEs}
\label{sec:closing_odes}

Evaluating the expectations in \cref{eq:sft_ode,eq:rl_ode} introduces
higher-order products of the pretraining and signal/noise projectors,
such as $\mathbf U_{ij}=\mathbf P\mathbf M_i\mathbf P\mathbf M_j\mathbf P$.
Their associated order parameters in turn depend on higher-order
products, generating a hierarchy of dynamical observables. We truncate
this hierarchy by approximating higher-order operators through
projection onto the retained lower-order operator
basis. This is a polynomial least-squares approximation and can be
applied at any chosen truncation order. Here, we retain the four order
parameters $(J_s,J_n,Q_s,Q_n)$, using
\[
\mathbf U_{ij}\approx
\Omega_{ij}^{(s)}\mathbf P\mathbf M_s\mathbf P+
\Omega_{ij}^{(n)}\mathbf P\mathbf M_n\mathbf P,
\]
with coefficients chosen to minimize the squared Frobenius norm of the operator approximation error. A six-order-parameter closure is considered in the appendix.

The projection coefficients are completely determined by normalized trace moments
\[
m_k=\frac{1}{D}\operatorname{Tr}
\bigl[(\mathbf P\mathbf M_s)^k\bigr].
\]
For the four-order-parameter closure, only $m_1$, $m_2$, and $m_3$ are
required.
Random matrix theory determines the exact high-dimensional limits of
these moments, which agree with empirical PCA and trained-LAE
measurements in \cref{fig:2}a. The resulting dynamical closure remains approximate because it
truncates the order-parameter hierarchy. The derivation and
random-matrix calculations are given in
\cref{App:TheoryProjected,App:ExpectedTraceConstants}. We empirically assess accuracy of the higher order closure in \cref{app:self_averaging}.

\section{Results}
\label{sec:results_sft_rl_dynamics}

We examine how pretraining sample size $\alpha=N/D$ and SFT
label noise $\epsilon$ affect learning dynamics, asymptotic
performance, and the benefit of SFT before RL. We then optimize
compute allocation across the three training stages and extend
the model to study how RL modifies individual skills in
compositional tasks.

\paragraph{SFT and RL exhibit distinct learning dynamics.}
The closed ODEs closely match simulated order-parameter
trajectories for both SFT and RL (\cref{fig:2}b--c).
SFT begins learning immediately from stepwise supervision,
whereas RL exhibits an initial plateau followed by rapid
improvement. Under the all-correct reward criterion, an episode
produces an update with probability $P^T$. Rewards are therefore
rare when initial accuracy is low, but become more frequent as
learning progresses. This accounts for the longer plateaus at
larger episode lengths $T$ reported in
\citet{RLperceptron}. Figure~\ref{fig:2}d shows accuracy throughout SFT followed by RL, when the pretraining level and time durations of SFT/RL are jointly optimized for different total compute budgets. We examine the benefits of SFT initialization below and the allocation problem in \cref{sec:budget_allocation}.

Unless otherwise stated, subsequent experiments use a locally
optimal learning rate $\eta_{\mathrm{opt}}$ that maximizes the
instantaneous predicted increase in generalization accuracy
(see \cref{App:LocallyOptimalLR}).

\begin{figure*}[t]
    \centering
        \includegraphics[width=\linewidth]{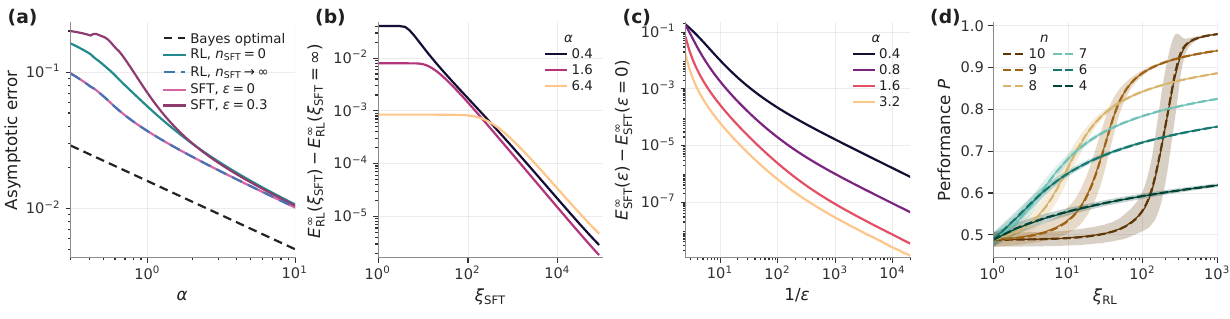}
    \caption{\textbf{Asymptotic performance predictably scales with batch size (a), degree of SFT warm start (b), label noise (c) and reward stringency (d).} (a) Power law scaling in asymptotic error reached by RL and SFT with higher/lower intermediate label noises $\epsilon$, and Bayes optimal scaling (black dashed). (b) SFT warm start improves the RL asymptotic performance. The residual asymptotic RL error after infinite-time vs $\xi_{SFT}$ SFT warm start is plotted against $\xi_{SFT}$, showing diminishing returns in accuracy. (c) The residual SFT asymptotic error for $\epsilon$-noise vs 0-noise against inverse $\epsilon$ shows diminishing returns in label quality and robustness to label-noise. (d) Evolution of generalisation performance ($P$) is plotted for different levels of RL reward stringency, where `$n$ or more' actions in an episode are required correct for reward. We observe a speed accuracy tradeoff mediated by reward stringency. \textit{Parameters}:  (a), (b) and (c) Locally optimal learning rate, spherically constrained, $T=6$, $\rho=0.2$, $\Delta=0.1$. (d) $T=10$, $\alpha=7$, $\eta=1$, spherically unconstrained.} 
    \label{fig:3}
    \vspace{-5mm}
\end{figure*}

% \paragraph{SFT performance is sensitive to training data noise.}
% In commonly used verifiable reward setting with bulk feedback (e.g., coding, math), RL receives feedback only on the final outcome's correctness, and therefore does not rely on noisy step-by-step supervision. By contrast, SFT learns from direct supervision at each intermediate output, making its performance sensitive to noise in the training trajectories. We parameterize this intermediate-step noise by $\epsilon$, which controls the corruption level of the supervised targets provided to SFT. Our model shows that this noise is an important determinant of SFT efficiency. \Cref{fig:3}a shows that, at the same pretraining level $\alpha$, SFT can be less efficient than RL when the supervision is noisy ($\epsilon = 0.35$, magenta), but more efficient than RL when the supervision is cleaner ($\epsilon = 0.1$, pink). Moreover, the performance gap between SFT and RL becomes more pronounced as the pretraining level decreases. This observation recapitulates the empirical findings of SFT efficiency dependency on the SFT data quality studied in~\citeauthor{ye2025limo}~(\cref{fig:3}b-c). 

\paragraph{SFT and RL attain the Bayes scaling exponent.}
Figure~\ref{fig:3}a shows that the asymptotic SFT and RL errors
decrease as $\alpha^{-1/2}$, matching the scaling of the
representation-dependent Bayes error despite remaining above it.
For an isotropic teacher in the signal subspace, at large $\alpha$, \cref{app:large_alpha_scaling} gives
\[
e_B(\alpha)=
\frac{\Delta\sqrt{1-\rho}}{\pi(1-\Delta)}\alpha^{-1/2}
+O(\alpha^{-1}).
\]
This result identifies a mechanism by which increasing pretraining
data produces power-law improvements in downstream performance. The $\alpha^{-1/2}$ scaling follows from the accuracy of the
pretrained subspace: finite-sample covariance fluctuations leave
an unrecoverable fraction of teacher-field variance of order
$\alpha^{-1}$, and the Bayes classification error scales as
the square root of this fraction.

\paragraph{SFT warm starts improve asymptotic RL performance.}
Under the spherical constraint and locally optimal learning rate used
in \cref{fig:3}a--b, increasing the duration of SFT reduces the
asymptotic error of subsequent RL. Figure~\ref{fig:3}b reports
$e_{\rm RL}^{\infty}(\xi_{\rm SFT})-
e_{\rm RL}^{\infty}(\infty)$,
the excess error relative to RL initialized from converged SFT.
For short warm starts, this excess error depends strongly on
pretraining quality. At longer SFT durations, the curves approach
a common power-law decay, indicating similar convergence towards
their respective fully warm-started RL asymptotes.

\paragraph{Pretraining improves SFT robustness to label noise.}
Figure~\ref{fig:3}c shows the excess asymptotic error due to label
corruption,
$\Delta e^\infty(\alpha,\epsilon)
=e^\infty_{\mathrm{SFT}}(\alpha,\epsilon)
-e^\infty_{\mathrm{SFT}}(\alpha,0)$.
At fixed $\epsilon$, increasing $\alpha$ reduces this excess error,
showing that better pretraining reduces sensitivity to corrupted
supervision in addition to improving clean-label performance.
For small $\epsilon$, the approximately parallel curves indicate
power-law scaling with a similar exponent across pretraining levels
and a prefactor that decreases with $\alpha$. For the unconstrained SFT dynamics, the \cref{app:large_alpha_scaling} derives asymptotic SFT performance
$e_{\mathrm{SFT}}^\infty(\alpha,\epsilon=0)\sim\sqrt{2}\,e_B(\alpha)$.
For fixed $0<\epsilon<1/2$, the SFT error instead approaches
$e_\epsilon=\pi^{-1}\arccos(1-2\epsilon)$, with
$e_{\mathrm{SFT}}^\infty(\alpha,\epsilon)-e_\epsilon=O(\alpha^{-1})$.

\paragraph{Reward stringency mediates a speed accuracy tradeoff.}
We consider the reward function that rewards an episode when at least $n$ of its $T$ decisions are correct, $R = \mathbf{1}(\sum_{t=1}^T \theta(y_t y_t^*) \geq n)$.
As noted by \cite{RLperceptron}, \cref{fig:3}d shows that relaxing the all-correct criterion can
accelerate early learning at the expense of asymptotic accuracy. 
With constant learning rate and unconstrained weights, the initial accuracy slope is larger by a factor
$\binom{T-1}{n-1}$ than under the all-correct rule. However, the asymptotic error decays as
$e_g(\xi)\propto(\log\xi)^{-1/(T-n)}$ for $n<T$, compared with
$e_g(\xi)\propto\sqrt{\log\xi/\xi}$ for $n=T$. Less stringent reward therefore allows faster initial learning but slower asymptotic convergence. Intuitively, this is because early in learning rewards are more frequent with less stringent reward, but conversely near convergence episodes containing incorrect decisions are increasingly rewarded. With spherical normalization and constant learning rate, the
small-error stationary solution instead scales as
$e_g^\infty\propto(\eta/Q)^{1/(T-n+2)}$ as $\eta/Q\to0$,
where $Q=Q_s + Q_n$ is fixed. The derivations are provided in \cref{app:nmore_isotropic}.

% \paragraph{Pretraining quality sets the performance ceiling and power-law scaling behavior.}
% While SFT and RL exhibit different learning dynamics, both are ultimately constrained by the quality of the pretrained representation. The pretraining quality $\alpha$ sets the asymptotic performance for both fine-tuning methods (\cref{fig:2}a-b), that is, \textit{even if the fine-tuning compute budget is scaled, the reachable performance remains bounded by the pretraining quality}. The interpretation follows directly from our model design, where $\alpha$ controls the quality of the learned feature space. Since both methods operate on this feature space, $\alpha$ determines how much task-relevant information is available for downstream learning. Thus, fine-tuning can improve the use of learned features, but cannot fully recover task-relevant structure that was not captured during pretraining. This allows us to characterize power-law scaling~\citep{hoffmann2022training,kaplan2020scaling}. In \cref{fig:3}a, the asymptotic test error decreases as a power law with pretraining compute, indicating that improvements in the learned representation translate systematically into downstream performance gains. Thus, the model connects the performance ceiling of fine-tuning to a measurable scaling relationship with pretraining quality. \textcolor{red}{This is a reasurring sign the model captures AN aspect of the scalign behaviour observed in practice.}

\begin{wrapfigure}{R}{0.48\columnwidth}
    \centering
    \vspace{-8pt}
    \includegraphics[width=\linewidth]{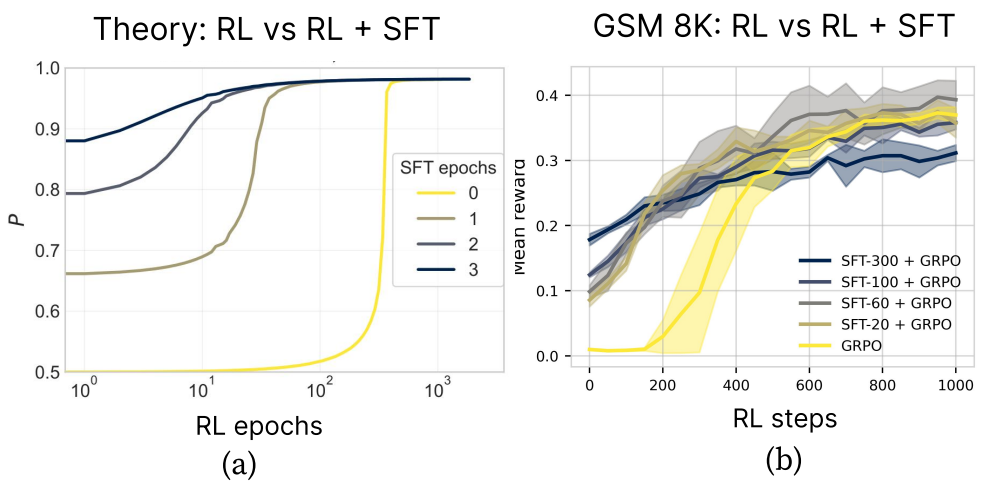}
    \caption{\textbf{SFT overcomes the RL Plateau.} (a) Pure RL exhibits a plateau due to reward sparsity (yellow). It is beneficial to first run SFT to overcome this plateau (navy) since SFT provides dense supervision without waiting for a successful episode. (b) We replicate that early SFT complementing RL by removing plateau in a realistic model and task using OlMo2 1B~\citep{olmo20242} and GSM8K~\citep{cobbe2021training}. We show training curves during RL after different amount of SFT over 3 seeds. \textit{Parameters}: $\alpha=4$, $T=12$, $\rho=0.2$, $\Delta=0.1$. }
    \label{fig:4}
    \vspace{-10pt}
\end{wrapfigure}

\paragraph{SFT overcomes the RL plateau.}
Figure~\ref{fig:4}a compares RL trajectories initialized after different
amounts of SFT. Increasing the duration of preceding SFT raises the initial probability of
a successful episode to a higher degree, allowing subsequent RL to quickly improve performance. A corresponding qualitative
effect appears in OLMo-2 1B~\citep{olmo2} finetuned on GSM8K~\citep{cobbe2021gsm8k}
(\cref{fig:4}b): models initialized from SFT checkpoints show faster reward
improvement while the RL-only baseline initially remains near zero.
Experimental details are in \cref{app:olmo_exp}. We next include the
costs of all stages to evaluate the complete pipeline.

% \paragraph{SFT prior to RL removes reward plateau and improves RL sample efficiency.}
% While end-to-end RL finetuning showed success for foundation models in reasoning tasks~\cite{deepseekai2025deepseekr1}, instantiating RL after SFT is common in practice~\cite{ouyang2022training,kirk2023understanding} and sometimes more efficient~\cite{chu2025sft}. We consider this setup, pretraining followed by SFT and then RL,  with our model. In \cref{fig:4}a, we observe that applying brief SFT before RL removes the plateau (navy) compared to finetuning with RL only (yellow), which requires a significantly larger number of steps to escape the plateau. We replicate this finding in a realistic model and a task using OlMo 2 1B~\citep{olmo20242} and GSM8k~\citep{cobbe2021training}, where we see that brief SFT prior to RL removes the characteristic initial plateau period (\cref{fig:4}b). See~\cref{app:olmo_exp} for more experimental details.

\subsection{Optimal compute budget allocation}
\label{sec:budget_allocation}

The previous section showed that fine-tuning performance is shaped by both the quality of the pretrained representation and the reliability of SFT supervision. We therefore extend our compute-allocation analysis to the full training pipeline, including pretraining, and ask how the optimal allocation changes as the SFT data quality varies. The compute costs for SFT and RL are computed as defined in \cref{app:pretraining_cost}.

\paragraph{Compute model and allocation objective.}
We count a multiply-add as two FLOPs. For an LAE of hidden width
$\rho D$, trained for $M$ updates on a fixed batch of size $\alpha D$,
the assigned pretraining cost is $C_{\rm PT}=10M\alpha\rho D^3$.
The assigned SFT and RL cost per episode is
$c_{\rm ep}=2\rho TD^2$, including the frozen encoder
projection and readout update. Thus a finetuning duration $\xi$
costs $D c_{\rm ep}\xi$ when $\xi=\mu/D$.
The resulting budget constraint is
\[
2\rho D^3 (5M\alpha + T(\xi_{\rm SFT}+\xi_{\rm RL})) \leq \mathrm B_{\rm FLOPs}.
\]
For convenience, we express compute in normalized units,
$\widetilde B = B_{\rm FLOPs}/(2\rho D^3)$,
so that the budget constraint becomes
\(
5M\alpha+T(\xi_{\rm SFT}+\xi_{\rm RL})
\leq \widetilde B.
\) All compute budgets reported in the following
figures are in these normalized units unless
otherwise stated.
We numerically optimize the final accuracy over the pretraining level and the
SFT/RL durations, with SFT preceding RL. Costs and optimization details are given in
\cref{app:pretraining_cost,App:BudgetAllocation}.

%For zero label noise in SFT ($\epsilon=0$), better pretraining quality (lower $\rho$, darker line) results in more compute allocated to SFT than RL, especially in the compute-limited condition (\cref{fig:3b}). As the compute budget increases, the optimal relative compute share for RL increases. This reflects our previous finding that the RL is compute-intensive due to the long initial plateau but is less affected by poor pretraining quality than SFT. We comment on SFT with noisy data ($\epsilon>0$) in the next section.

\begin{figure*}[t]
    \centering
\includegraphics[width=\linewidth]{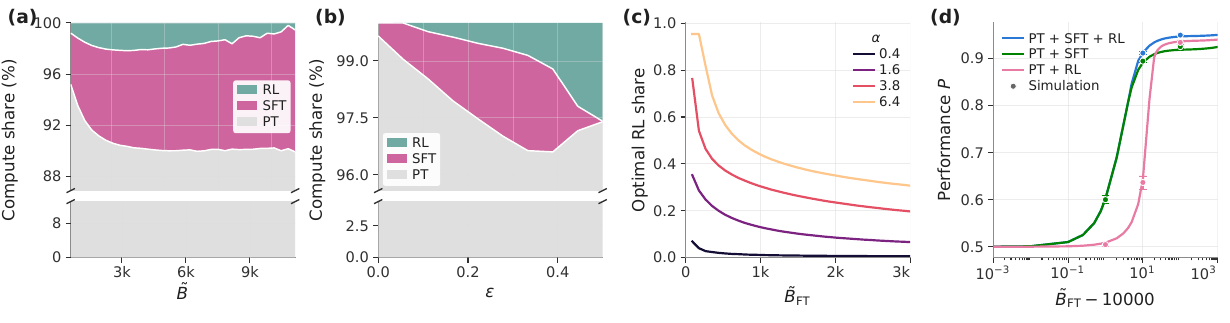}
    \vspace{-0.3cm}
    \caption{\textbf{Optimal allocation of compute across training stages}, (a) Optimal compute shares allocated to pretraining (gray), SFT (pink), and RL (teal) as the total budget varies. Most compute is allocated to pretraining; within the fine-tuning budget there is non-monotonicity in the optimal RL share. (b) Optimal compute shares at a fixed budget of 50000 flops as the SFT label noise $\epsilon$ varies. Higher SFT noise reduces the optimal share allocated to SFT and increases the share allocated to RL. (c) The optimal finetuning compute share to be allocated to RL under a given pretraining level $\alpha$. With a strong pretrained model, more share of the RL compute is beneficial, but its optimal share decreases as the finetuning budget increases. (d) Best achievable performance under three training paradigms plotted against Total budget, offset by 18000 FLOPs: pretraining + SFT + RL (blue), pretraining + SFT only (green), and pretraining + RL only (pink).
    The full multistage pipeline performs best across all of the compute-limited regime, SFT only outperforms RL only for smaller budgets, reflecting the overcoming of the RL plateau, before RL overtakes at larger budgets. \textit{Parameters}: $T=6$, $\rho=0.2$, $\Delta=0.1$, 1000 pretraining update steps, spherically constrained, locally optimal learning rate. (a)  $\epsilon=0.2$. (c) $\epsilon=0.3$. (d) $\epsilon=0.3$, Simulation $D=1000$.}
    \label{fig:5}
    \vspace{-0.5cm}
\end{figure*}

\paragraph{Pretraining quality and budget determine the optimal SFT--RL split.}
Figure~\ref{fig:5}c shows the optimal fraction of finetuning compute
allocated to RL at fixed pretraining level $\alpha$. At a given
budget, better pretraining increases the optimal RL fraction,
whereas at fixed $\alpha$, increasing the finetuning budget shifts
the optimal allocation towards SFT. The optimality of increasing the proportion of preceding SFT compute is consistent with the warm-start dependence
in \cref{fig:3}b: longer SFT can improve the asymptotic performance
of subsequent RL, in addition to shortening its initial plateau.

\paragraph{Optimal budget allocation under varying conditions}
 
When compute is allocated jointly across all three stages, pretraining
receives the largest share (\cref{fig:5}a--b). The optimal RL share
initially increases with the total budget before decreasing
(\cref{fig:5}a), demonstrating that additional compute need not favour
a progressively larger RL allocation. Within the remaining fine-tuning budget, SFT remains important at low noise because it helps overcome the initial RL plateau through dense supervision (\cref{fig:5}a). At a fixed total budget, increasing intermediate-label noise
increases the optimal RL share in
\cref{fig:5}b. Intuitively, at zero label noise, SFT is compute optimal, and increasing noise reduces its effectiveness.

In \cref{fig:5}d, we compare the maximum achievable performance under three training paradigms: the full pretraining + SFT + RL pipeline (blue), pretraining + SFT only (green), and pretraining + RL only (pink).  The full three-stage pipeline is compute-optimal the entire compute-limited regime, showing the benefit of combining SFT and RL as we demonstrated previously by showing SFT removes the early reward plateau from RL. At smaller compute budgets, SFT-only outperforms RL-only, and is compute optimal in the multi-stage case too, since it can overcome the RL plateau. But as compute budgets increase, the optimality of RL-only surpasses the SFT-only, since it avoids noisy SFT supervision. However, it still falls short of the full three-stage pipeline, due the increase in asymptotic performance afforded by SFT warm-start.

\subsection{RL finetuning and skill amplification}

\label{skill-composition}

So far, we have treated pretraining quality as a single scalar controlling the overall usefulness of the learned representation. In practice, however, pretrained models may contain many latent skills, only some of which are relevant to a given downstream task. Fine-tuning on a domain specific task, such as on math or coding, may therefore selectively amplify task-relevant skills rather than uniformly improve the full representation~\cite{zhao2025echo,shao2025spurious}. This motivates the question: how does the fine-tuning task specification influence the skills acquired during pretraining?

We address this question by extending our analytical model to study RL finetuning as learning to linearly compose multiple `pretrained' skills into a more complex task~\citep{lee2024animals}. Concretely, each skill is represented by one of $K$ independent teacher-student pairs ($\mathbf{w}^*_{1...K}, \mathbf{w}_{1...K}$), and the downstream task is defined by a context teacher-student vector ($V^*, V \in \mathbb{R}^{K}$) that selects and combines these skills. During fine-tuning, both the student skills and the context are updated. This setup allows some pretrained skills to be task-relevant, while others are present in the base model but unused by the downstream task. We provide the ODE derivation for this compositional RL setting in \cref{app:skill_composition}. The notion of pretraining in this setup is how well each skill is learned before the downstream task.

\begin{figure}[t]
    \centering
    \begin{subfigure}[b]{0.4\linewidth}
        \includegraphics[width=\linewidth]{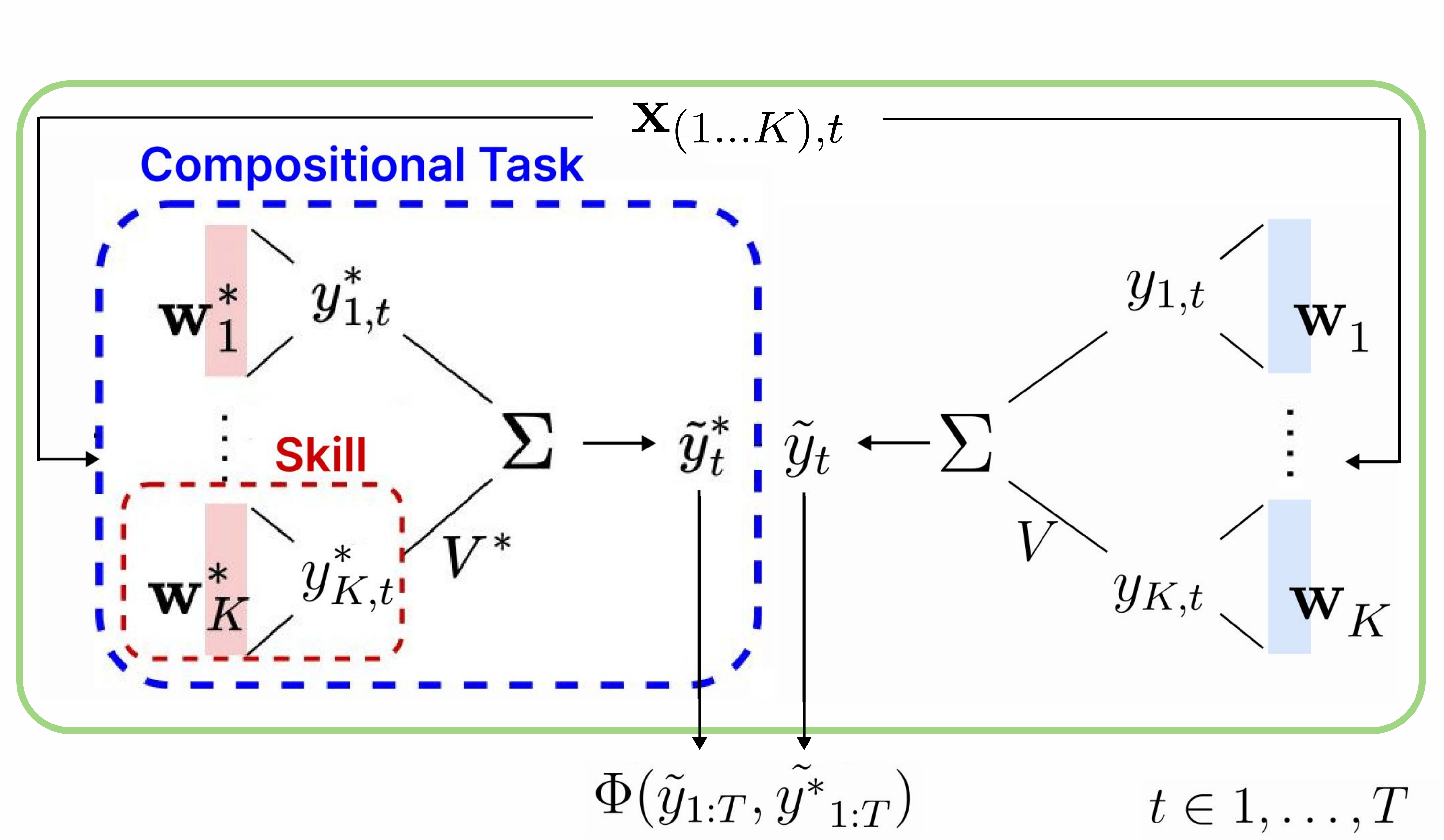}
        \caption{}
        \label{fig:6a}
    \end{subfigure}%
    \hfill
    \begin{subfigure}[b]{0.3\linewidth}
        \includegraphics[width=\linewidth]{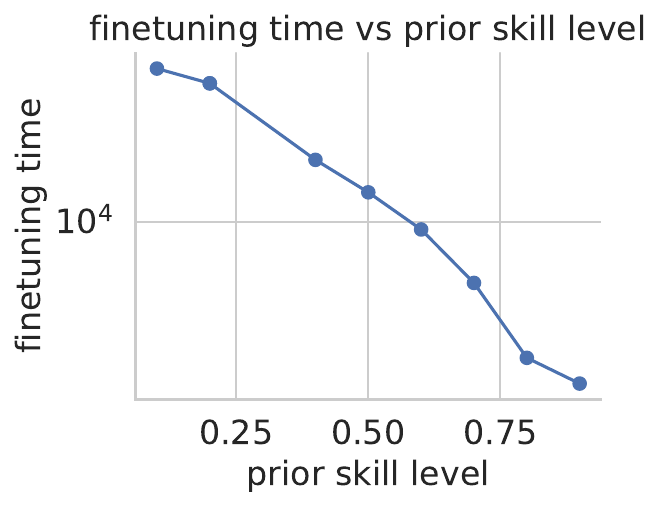}
        \caption{}
        \label{fig:6b}
    \end{subfigure}%
    \hfill
    \begin{subfigure}[b]{0.3\linewidth}
    \includegraphics[width=\linewidth]{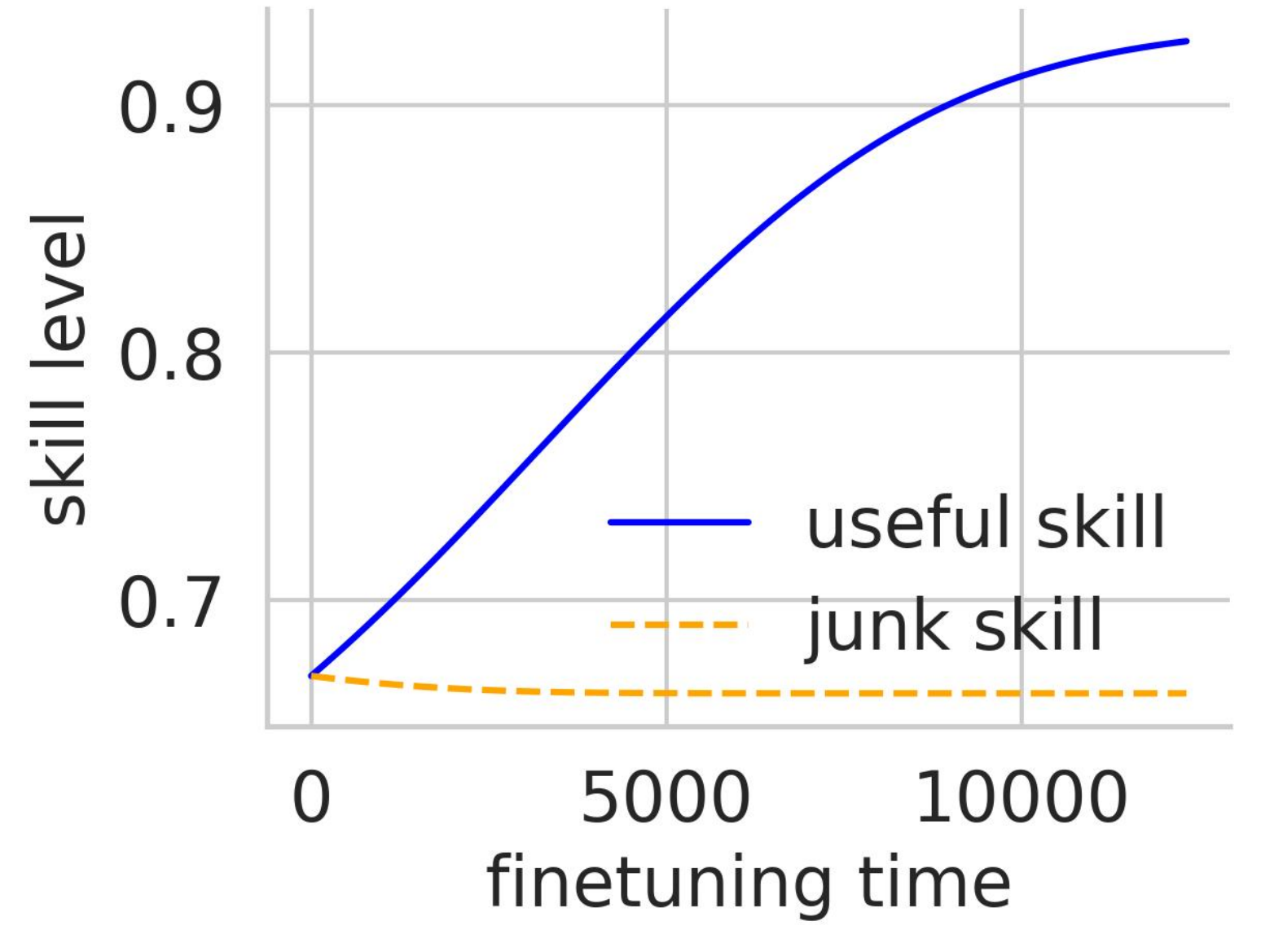}
        \caption{}
        \label{fig:6c}
    \end{subfigure}
   \caption{\textbf{Skill composition in RL finetuning.} (a) Schematic of a compositional RL perceptron where individual skill is modeled as a single RL perceptron and compositional task is defined as a linear combination of those skills. (b) The fine-tuning time required to get the compositional task correct decreases as the pre-trained level of individual skill becomes better. (c) During the finetuning, the skill used in the finetuned compositional task further improves and is amplified. The junk skill previously acquired during pretraining but not used for the compositional task persists, not forgotten. \textit{Parameters}: $\eta=1$, $T=8$, $K=8$, $J_{\text{init}} = 0.5$.}
    \label{fig:6}
    \vspace{-5mm}
\end{figure}

\paragraph{Prior skill proficiency accelerates relevant RL finetuning.}
We first vary the initial proficiency of each skill and measure the fine-tuning time required to reach 99\% of the maximal task performance. Fine-tuning converges exponentially faster as prior skill proficiency improves (\cref{fig:6b}), echoing the importance of a strong pretrained base model for RL efficiency.

\paragraph{RL amplifies task-relevant skills while preserving unused skills.}
We then study how task specification shapes individual skills during fine-tuning. We distinguish \textit{useful skills}, which contribute to the downstream task, from \textit{junk skills}, which are present in the base model but unused for finetuned task. As shown in \cref{fig:6c}, useful skills are further amplified, while junk skills initially deteriorate only mildly and are not catastrophically forgotten. This aligns with previous empirical observations that RL amplifies useful patterns learned during pretraining~\citep{zhao2025echo,shao2025spurious}, and predicts that unused skills may remain largely recoverable for future tasks where they become relevant.

\section{Discussion}

Our contribution is an analytically tractable starting point for
studying multistage training pipelines in foundation models.
Despite its simplicity, the model captures power-law error scaling,
SFT--RL complementarity, warm-start dependence, and selective skill
amplification. The dependence of downstream learning on pretraining quality is consistent with controlled experiments showing that stronger pretrained models improve faster and reach higher performance under RL~\citep{shen2026understanding}, and that pretraining quality is a central bottleneck for downstream performance and compute efficiency~\citep{akter2025front,yue2025does, ye2025limo}. The benefit of supervised preparation is consistent with empirical findings that stronger SFT initialization can improve subsequent RL~\citep{liu2025acereason}.

Our allocation analysis excludes the cost of acquiring high-quality
supervision, which can substantially improve post-training
\citep{zhou2023lima,ye2025limo} but is difficult to obtain at scale. In our model, proportional reductions
in label noise yield diminishing absolute improvements in SFT error,
while better pretraining reduces sensitivity to noise. Accounting
for data acquisition and verification costs could provide more practical
insights into when improving supervision is preferable to additional pretraining or RL.

Broadly, by making the interactions between pretraining, SFT, and RL analytically tractable, our model provides a starting point for understanding how these stages jointly impact learning performance.

\textbf{Limitations}: Our model is a simplified and abstract setup, where each component is modeled as a simple single-layer network. The real-world models rely on complex architectures and heavy engineering, which we do not account for. Our proposed model of RL captures an aspect of bulk feedback in RL, but it assumes independence between all observations in an episode. In practice, the cost of SFT is not just a compute one, but the cost of being able to source enough high quality supervision data we do not take into account the difficulty of obtaining the labels of the data for defining compute budget in our analysis. Using frozen encoder weights is a simplification, in practice all weights are updated during fine-tuning.

\section*{Acknowledgements}

We thank the following funding
sources: Gatsby Charitable Foundation (GAT4058) to AS; Sainsbury Wellcome
Centre Core Grant from Wellcome (219627/Z/19/Z) to AS; Schmidt Science Polymath Award to
AS. AS is a CIFAR Azrieli Global Scholar in the Learning in Machines \& Brains program. The authors declare no competing interests.

\bibliography{icml2026_references}
\bibliographystyle{plainnat}

%%%%%%%%%%%%%%%%%%%%%%%%%%%%%%%%%%%%%%%%%%%%%%%%%%%%%%%%%%%%%%%%%%%%%%%%%%%%%%%
% APPENDIX
%%%%%%%%%%%%%%%%%%%%%%%%%%%%%%%%%%%%%%%%%%%%%%%%%%%%%%%%%%%%%%%%%%%%%%%%%%%%%%%
\newpage
\appendix

All code and instructions to reproduce the figures are available at \url{https://github.com/nishp99/multistage_pipeline.git}.
\subsection{Teacher-student framework}
\label{app:teacher_student}
There is a growing body of neural network theory of learning and generalization based on a teacher-student framework that applies tools from statistical physics, which allows us to characterize the typical behavior in high-dimensional stochastic systems~\cite {gardner1989three,seung1992statistical,engel2001statistical, bahri2020statistical,zdeborova2016statistical}. Previous works studied diverse learning paradigms and setups, including vanilla supervised learning~\cite{du2018gradient, goldt2019dynamics,aubin2018committee,advani2020high}, continual learning~\cite{lee2021continual,lee2022maslow,asanuma2021statistical}, curriculum learning~\cite{saglietti2022analytical,lee2024animals}, transfer learning~\cite{gerace2022probing,lampinen2018analytic} and online policy learning~\cite{RLperceptron}.

\section{Derivation of ODEs}
\label{App:TheoryProjected}

\subsection{Thermodynamic Limit}

We follow the standard procedure of \citeauthor{SaadSolla}, but the main difference is the random projection which requires averaging over. It was shown by \citet{yoshida2019data} that macroscopically describing the dynamics when data is sampled from a centred Gaussian generates an infinite hierarchy of order parameters to track, and the ODE system can only be closed if the covariance has a finite number of unique eigenvalues, where there is a pair of order parameters associated to every unique eigenvalue. In our case, the projection onto the empirical principal subspace introduces an effective data covariance with, in general, $\rho D$ unique eigenvalues. We prevent the infinite cascade of ODEs by approximating the higher-order order parameters in terms of lower order ones using the Hilbert Schmidt Operator projection.

We denote by \(\mathbf{M}_s\) and \(\mathbf{M}_n=\mathbf{I}-\mathbf{M}_s\) the orthogonal projectors onto the signal and noise subspaces, respectively. The data covariance is
\begin{align}
    \Sigma = \mathbf{M}_s + \Delta \mathbf{M}_n,
\end{align}
where \(0<\Delta<1\). The teacher \(\mathbf{w}^*\) is assumed to lie entirely in the signal subspace, so that \(\mathbf{M}_n\mathbf{w}^*=0\). The student does not observe \(\mathbf{x}\) directly, but instead observes the projected input \(\mathbf{P}\mathbf{x}\), where \(\mathbf{P}\) is the orthogonal projector onto the PCA subspace learned during pretraining.

We define the following order parameters:
\begin{align}
    J_i &= \frac{\mathbf{w}^{\top}\mathbf{P}\mathbf{M}_i\mathbf{P}\mathbf{w}^*}{D}, 
    &
    Q_i &= \frac{\mathbf{w}^{\top}\mathbf{P}\mathbf{M}_i\mathbf{P}\mathbf{w}}{D},
    &
    S_i &= \frac{{\mathbf{w}^*}^{\top}\mathbf{P}\mathbf{M}_i\mathbf{P}\mathbf{w}^*}{D},
\end{align}
for \(i\in\{s,n\}\), together with
\begin{align}
    S = \frac{{\mathbf{w}^*}^{\top}\mathbf{w}^*}{D}.
\end{align}
The total student-teacher overlap observed through the projected representation is
\begin{align}
    J = J_s + J_n
    = \frac{\mathbf{w}^{\top}\mathbf{P}\mathbf{w}^*}{D},
\end{align}
and the variance of the student field is
\begin{align}
    \widehat Q = Q_s + \Delta Q_n.
\end{align}

We define the local fields
\begin{align}
    \lambda &= \lambda_s + \lambda_n = \frac{\mathbf{w}^{\top}\mathbf{P}\mathbf{x}}{\sqrt{D}},
    &
    \nu &= \frac{{\mathbf{w}^*}^{\top}\mathbf{x}}{\sqrt{D}},
\end{align}
and the projected subfields
\begin{align}
    \lambda_i &= \frac{\mathbf{w}^{\top}\mathbf{P}\mathbf{M}_i\mathbf{P}\mathbf{x}}{\sqrt{D}},
    &
    \tilde{\nu}_i &= \frac{{\mathbf{w}^*}^{\top}\mathbf{P}\mathbf{M}_i\mathbf{P}\mathbf{x}}{\sqrt{D}},
    \qquad i\in\{s,n\}.
\end{align}
The student and teacher actions are
\begin{align}
    y = \mathrm{sgn}(\lambda),
    \qquad
    y^* = \mathrm{sgn}(\nu),
\end{align}
with noisy labels in SFT obtained by flipping \(y^*\) with probability \(\epsilon\).

The fields are jointly Gaussian with zero mean. The covariances needed below are
\begin{align}
    \mathrm{Var}(\lambda) &= \widehat Q = Q_s + \Delta Q_n,
    &
    \mathrm{Var}(\nu) &= S,
    &
    \mathrm{Cov}(\lambda,\nu) &= J_s + J_n.
\end{align}
We also need
\begin{align}
    \mathrm{Cov}(\nu,\lambda_i) &= J_i,
    &
    \mathrm{Cov}(\nu,\tilde{\nu}_i) &= S_i.
\end{align}
Because \(\mathbf{P}\) does not in general commute with \(\mathbf{M}_s\) and \(\mathbf{M}_n\), the remaining covariances involve higher-order operator products:
\begin{align}
    Q_{ij}
    &= \frac{\mathbf{w}^{\top}
    \mathbf{P}\mathbf{M}_i\mathbf{P}\mathbf{M}_j\mathbf{P}
    \mathbf{w}}{D},
    \\
    J_{ij}
    &= \frac{\mathbf{w}^{\top}
    \mathbf{P}\mathbf{M}_i\mathbf{P}\mathbf{M}_j\mathbf{P}
    \mathbf{w}^*}{D}.
\end{align}
Using these, we have
\begin{align}
    \mathrm{Cov}(\lambda,\lambda_s)
    &= Q_{ss} + \Delta Q_{sn},
    &
    \mathrm{Cov}(\lambda,\lambda_n)
    &= Q_{sn} + \Delta Q_{nn},
    \\
    \mathrm{Cov}(\lambda,\tilde{\nu}_s)
    &= J_{ss} + \Delta J_{ns},
    &
    \mathrm{Cov}(\lambda,\tilde{\nu}_n)
    &= J_{sn} + \Delta J_{nn}.
\end{align}

\subsection{Closure of Higher-Order Projector Products}
\label{app:closure}

The non-commutativity of \(\mathbf{P}\) with \(\mathbf{M}_s\) and \(\mathbf{M}_n\) generates terms such as
\[
    \mathbf{P}\mathbf{M}_i\mathbf{P}\mathbf{M}_j\mathbf{P}.
\]
To close the dynamics, we project these operators onto the two-dimensional operator basis
\[
    \mathbf{B}_s = \mathbf{P}\mathbf{M}_s\mathbf{P},
    \qquad
    \mathbf{B}_n = \mathbf{P}\mathbf{M}_n\mathbf{P}.
\]
Thus we approximate
\begin{align}
    \mathbf{P}\mathbf{M}_i\mathbf{P}\mathbf{M}_j\mathbf{P}
    \approx
    \Omega_{ij}^{(s)} \mathbf{P}\mathbf{M}_s\mathbf{P}
    +
    \Omega_{ij}^{(n)} \mathbf{P}\mathbf{M}_n\mathbf{P}.
    \label{eq:omega_projection}
\end{align}
This gives the closed approximations
\begin{align}
    Q_{ij}
    &\approx
    \Omega_{ij}^{(s)} Q_s
    +
    \Omega_{ij}^{(n)} Q_n,
    \\
    J_{ij}
    &\approx
    \Omega_{ij}^{(s)} J_s
    +
    \Omega_{ij}^{(n)} J_n.
\end{align}

The coefficients \(\Omega_{ij}^{(s)}\) and \(\Omega_{ij}^{(n)}\) are obtained by Hilbert-Schmidt projection, which we will detail now. The purpose of the Hilbert-Schmidt projection is not to assert that $\mathbf P\mathbf M_i\mathbf P\mathbf M_j\mathbf P$ is exactly a linear combination of \(\mathbf P\mathbf M_s\mathbf P\) and \(\mathbf P\mathbf M_n\mathbf P\). Rather, it gives the unique best approximation to this operator within the space already represented by the order parameters. More precisely, define
\[
    \mathbf U_{ij}
    =
    \mathbf P\mathbf M_i\mathbf P\mathbf M_j\mathbf P,
    \qquad
    \mathbf V_s
    =
    \mathbf P\mathbf M_s\mathbf P,
    \qquad
    \mathbf V_n
    =
    \mathbf P\mathbf M_n\mathbf P.
\]
We choose \(\Omega_{ij}^{(s)}\) and \(\Omega_{ij}^{(n)}\) to minimize
\[
    \frac{1}{D}
    \mathrm{Tr}
    \left[
    \left(
    \mathbf U_{ij}
    -
    (\Omega_{ij}^{(s)}\mathbf V_s
    +
    \Omega_{ij}^{(n)}\mathbf V_n)
    \right)^2
    \right].
\]
The discarded component, or residual operator, is therefore
\begin{align}
    \mathbf \Delta_{ij}
    =
    \mathbf U_{ij}
    -
    (\Omega_{ij}^{(s)}\mathbf V_s
    +
    \Omega_{ij}^{(n)}\mathbf V_n).
    \label{eq:residual_operator}
\end{align}

By construction, this residual is orthogonal, in trace average, to all tracked operators:
$\frac{1}{D}\mathrm{Tr}(\mathbf \Delta_{ij}\mathbf V_s)=0, \qquad \frac{1}{D}\mathrm{Tr}(\mathbf \Delta_{ij}\mathbf V_n)=0.$ This orthogonality does not ensure that the residual contractions with the student and teacher vanish: training can generate correlations with the eigendirections of \(\mathbf \Delta_{ij}\). Vanishing residual contractions would yield an exact quadratic-form closure, whereas residuals that remain \(O(1)\) require additional macroscopic order parameters. We therefore treat the four-order-parameter dynamics as approximate. Appendix~\ref{app:self_averaging} tests this approximation numerically, finding small relative residuals whose disorder-averaged magnitudes decrease with dimension over the tested range, without establishing self-averaging to zero.

Define the normalized trace constants
\begin{align}
    t &= \frac{1}{D}\mathrm{Tr}(\mathbf{P}),
    &
    t_s &= \frac{1}{D}\mathrm{Tr}(\mathbf{P}\mathbf{M}_s),
    \\
    T_{ss}
    &= \frac{1}{D}\mathrm{Tr}(\mathbf{P}\mathbf{M}_s\mathbf{P}\mathbf{M}_s),
    &
    \tau_{sss}
    &= \frac{1}{D}
    \mathrm{Tr}(\mathbf{P}\mathbf{M}_s\mathbf{P}\mathbf{M}_s\mathbf{P}\mathbf{M}_s).
\end{align}
Since \(\mathbf{M}_s+\mathbf{M}_n=\mathbf{I}\) and \(\mathbf{P}^2=\mathbf{P}\), all other second- and third-order trace constants can be written as
\begin{align}
    T_{sn} &= t_s - T_{ss},
    \\
    T_{nn} &= t - 2t_s + T_{ss},
\end{align}
and
\begin{align}
    V_{ss} &= \tau_{sss},
    \\
    V_{sn} &= T_{ss} - \tau_{sss},
    \\
    V_{nn} &= t_s - 2T_{ss} + \tau_{sss},
\end{align}
where
\begin{align}
    T_{ij}
    &= \frac{1}{D}\mathrm{Tr}
    \left(
    \mathbf{P}\mathbf{M}_i\mathbf{P}\mathbf{M}_j
    \right),
    \\
    V_{ij}
    &= \frac{1}{D}\mathrm{Tr}
    \left(
    \mathbf{P}\mathbf{M}_s\mathbf{P}\mathbf{M}_i\mathbf{P}\mathbf{M}_j
    \right).
\end{align}

Let
\begin{align}
    \Delta_G = T_{ss}t - t_s^2.
\end{align}
Then the projection coefficients are
\begin{align}
    \Omega_{ij}^{(s)}
    &=
    \frac{(t-t_s)V_{ij} - (t_s-T_{ss})T_{ij}}{\Delta_G},
    \\
    \Omega_{ij}^{(n)}
    &=
    \frac{-t_sV_{ij} + T_{ss}T_{ij}}{\Delta_G}.
    \label{eq:omega_coefficients}
\end{align}
Explicitly,
\begin{align}
    \Omega_{ss}^{(s)}
    &=
    \frac{(t-t_s)\tau_{sss} - (t_s-T_{ss})T_{ss}}{\Delta_G},
    \\
    \Omega_{sn}^{(s)}
    &=
    \frac{(t-t_s)(T_{ss}-\tau_{sss}) - (t_s-T_{ss})(t_s-T_{ss})}{\Delta_G},
    \\
    \Omega_{nn}^{(s)}
    &=
    \frac{(t-t_s)(t_s-2T_{ss}+\tau_{sss}) - (t_s-T_{ss})(t-2t_s+T_{ss})}{\Delta_G},
\end{align}
and
\begin{align}
    \Omega_{ss}^{(n)}
    &=
    \frac{-t_s\tau_{sss} + T_{ss}^2}{\Delta_G},
    \\
    \Omega_{sn}^{(n)}
    &=
    \frac{-t_s(T_{ss}-\tau_{sss}) + T_{ss}(t_s-T_{ss})}{\Delta_G},
    \\
    \Omega_{nn}^{(n)}
    &=
    \frac{-t_s(t_s-2T_{ss}+\tau_{sss}) + T_{ss}(t-2t_s+T_{ss})}{\Delta_G}.
\end{align}
A useful check is
\begin{align}
    \Omega_{nn}^{(s)} = \Omega_{ss}^{(s)} - 1,
    \qquad
    \Omega_{nn}^{(n)} = \Omega_{ss}^{(n)} + 1,
\end{align}
which indeed holds, and follows from
\[
    \mathbf{P}\mathbf{M}_n\mathbf{P}\mathbf{M}_n\mathbf{P}
    =
    \mathbf{P}
    -2\mathbf{P}\mathbf{M}_s\mathbf{P}
    +
    \mathbf{P}\mathbf{M}_s\mathbf{P}\mathbf{M}_s\mathbf{P}.
\]

The trace constants are static during finetuning. They depend only on the pretraining projector \(\mathbf{P}\) and on the true signal/noise decomposition. In the typical-case theory, these constants are self-averaging and are averaged over the randomness in the pretraining batch.

\subsection{Projected RL Dynamics}

The RL update is
\begin{align}
    \mathbf{w}^{(\mu+1)}
    =
    \mathbf{w}^{(\mu)}
    +
    \frac{\eta}{T}
    \left(
    \sum_{t=1}^T
    \frac{y_t^{(\mu)}\mathbf{P}\mathbf{x}_t^{(\mu)}}{\sqrt{D}}
    \right)
    R(y_{1:T}^{(\mu)}, {y^*}_{1:T}^{(\mu)}),
    \label{eq:rl_update_macroscopic}
\end{align}

\begin{align}
    \mathbf{w}^{(\mu+1)}
    =
    \mathbf{w}^{(\mu)}
    +
    \frac{\eta}{TG} \sum_{g=1}^G \left[
    \left(
    \sum_{t=1}^T
    \frac{y_t^{g,(\mu)}\mathbf{P}\mathbf{x}_t^{g,(\mu)}}{\sqrt{D}}
    \right)
    (R(y_{1:T}^{g,(\mu)}, {y^*}_{1:T}^{g,(\mu)}) - A^{(\mu)}) \right],
\end{align}

with $A = \frac{1}{G} \sum_{g=1}^G (R(y_{1:T}^{g,(\mu)}, {y^*}_{1:T}^{g,(\mu)}))$ and $R =  2 \left(\prod_{t=1}^T \theta(\nu_t\lambda_t)\right) - 1$

where for the all-correct reward rule,
\begin{align}
    R
    =
    \prod_{t=1}^T \theta(\nu_t\lambda_t).
\end{align}

Taking the inner product of the update with \(\mathbf{P}\mathbf{M}_i\mathbf{P}\mathbf{w}^*\), summing over a window of episodes, and taking the limit \(D\to\infty\), gives
\begin{align}
    \frac{dJ_i}{d\xi}
    =
    \frac{\eta}{T}
    \left\langle
    \sum_{t=1}^T
    \mathrm{sgn}(\lambda_t)
    \tilde{\nu}_{i,t}
    R
    \right\rangle.
\end{align}
Similarly, squaring the update gives
\begin{align}
    \frac{dQ_i}{d\xi}
    =
    \frac{2\eta}{T}
    \left\langle
    \sum_{t=1}^T
    \mathrm{sgn}(\lambda_t)
    \lambda_{i,t}
    R
    \right\rangle
    +
    \frac{\eta^2}{T}
    \frac{1}{D}
    \mathrm{Tr}
    \left(
    \mathbf{P}\mathbf{M}_i\mathbf{P}\Sigma
    \right)
    \left\langle R \right\rangle
    +
    \mathcal{O}\left(\frac{1}{D}\right).
\end{align}
The trace terms are
\begin{align}
    G_s
    &=
    \frac{1}{D}
    \mathrm{Tr}
    \left(
    \mathbf{P}\mathbf{M}_s\mathbf{P}\Sigma
    \right)
    =
    T_{ss}+\Delta T_{sn},
    \\
    G_n
    &=
    \frac{1}{D}
    \mathrm{Tr}
    \left(
    \mathbf{P}\mathbf{M}_n\mathbf{P}\Sigma
    \right)
    =
    T_{sn}+\Delta T_{nn}.
\end{align}

Let
\begin{align}
    P
    =
    \left\langle \theta(\nu\lambda)\right\rangle
    =
    \frac{1}{2}
    +
    \frac{1}{\pi}
    \sin^{-1}
    \left(
    \frac{J_s+J_n}{\sqrt{S(Q_s+\Delta Q_n)}}
    \right).
\end{align}

This is the probability that a single step is correct (the generalisation performance). And is trivially computed from the Gaussian statistics of the fields.
For the all-correct reward rule, \(\langle R\rangle=P^T\). Also,
\begin{align}
    \mathrm{sgn}(\lambda)\theta(\nu\lambda)
    =
    \frac{1}{2}
    \left[
    \mathrm{sgn}(\lambda)+\mathrm{sgn}(\nu)
    \right].
\end{align}
Therefore,
\begin{align}
    \frac{dJ_i}{d\xi}
    =
    \frac{\eta}{2}
    P^{T-1}
    \left(
    \left\langle \tilde{\nu}_i\mathrm{sgn}(\lambda)\right\rangle
    +
    \left\langle \tilde{\nu}_i\mathrm{sgn}(\nu)\right\rangle
    \right),
    \label{eq:rl_Ji_pre_average}
\end{align}
and
\begin{align}
    \frac{dQ_i}{d\xi}
    =
    \eta
    P^{T-1}
    \left(
    \left\langle \lambda_i\mathrm{sgn}(\lambda)\right\rangle
    +
    \left\langle \lambda_i\mathrm{sgn}(\nu)\right\rangle
    \right)
    +
    \frac{\eta^2}{T}G_i P^T.
    \label{eq:rl_Qi_pre_average}
\end{align}

\subsection{Computing the Gaussian Averages}

For zero-mean jointly Gaussian variables \(X,Y\),
\begin{align}
    \left\langle X\mathrm{sgn}(Y)\right\rangle
    =
    \sqrt{\frac{2}{\pi}}
    \frac{\mathrm{Cov}(X,Y)}{\sqrt{\mathrm{Var}(Y)}}.
\end{align}
Using the covariance structure above, define
\begin{align}
    C_s^Q &= Q_{ss}+\Delta Q_{sn},
    &
    C_n^Q &= Q_{sn}+\Delta Q_{nn},
    \\
    C_s^J &= J_{ss}+\Delta J_{sn},
    &
    C_n^J &= J_{sn}+\Delta J_{nn}.
\end{align}
Then
\begin{align}
    \left\langle \tilde{\nu}_i\mathrm{sgn}(\lambda)\right\rangle
    &=
    \sqrt{\frac{2}{\pi}}
    \frac{C_i^J}{\sqrt{\widehat Q}},
    \\
    \left\langle \tilde{\nu}_i\mathrm{sgn}(\nu)\right\rangle
    &=
    \sqrt{\frac{2}{\pi}}
    \frac{S_i}{\sqrt{S}},
    \\
    \left\langle \lambda_i\mathrm{sgn}(\lambda)\right\rangle
    &=
    \sqrt{\frac{2}{\pi}}
    \frac{C_i^Q}{\sqrt{\widehat Q}},
    \\
    \left\langle \lambda_i\mathrm{sgn}(\nu)\right\rangle
    &=
    \sqrt{\frac{2}{\pi}}
    \frac{J_i}{\sqrt{S}}.
\end{align}
Here \(Q_{ij}\) and \(J_{ij}\) are replaced by their closed forms from \ref{eq:omega_projection}:
\begin{align}
    Q_{ij}
    &=
    \Omega_{ij}^{(s)}Q_s+\Omega_{ij}^{(n)}Q_n,
    \\
    J_{ij}
    &=
    \Omega_{ij}^{(s)}J_s+\Omega_{ij}^{(n)}J_n.
\end{align}

Substituting these averages into \ref{eq:rl_Ji_pre_average} and \ref{eq:rl_Qi_pre_average} gives the closed RL ODEs:
\begin{align}
    \frac{dJ_i}{d\xi}
    &=
    \frac{\eta}{2}
    P^{T-1}
    \sqrt{\frac{2}{\pi}}
    \left(
    \frac{C_i^J}{\sqrt{\widehat Q}}
    +
    \frac{S_i}{\sqrt{S}}
    \right),
    \\
    \frac{dQ_i}{d\xi}
    &=
    \eta
    P^{T-1}
    \sqrt{\frac{2}{\pi}}
    \left(
    \frac{C_i^Q}{\sqrt{\widehat Q}}
    +
    \frac{J_i}{\sqrt{S}}
    \right)
    +
    \frac{\eta^2}{T}G_i P^T,
    \qquad i\in\{s,n\}.
    \label{eq:projected_rl_odes}
\end{align}

\subsection{Projected SFT Dynamics}

For SFT, the update is
\begin{align}
    \mathbf{w}^{(\mu+1)}
    =
    \mathbf{w}^{(\mu)}
    -
    \frac{\eta}{T\sqrt{D}}
    \sum_{t=1}^T
    y_t^{(\mu)}
    \mathbf{P}\mathbf{x}_t^{(\mu)}
    \theta\left(-y_t^{*(\mu)}y_t^{(\mu)}\right).
\end{align}
The label \(y^*\) is noisy: with probability \(1-\epsilon\), \(y^*=\mathrm{sgn}(\nu)\), and with probability \(\epsilon\), \(y^*=-\mathrm{sgn}(\nu)\).

Following the same thermodynamic-limit procedure,
\begin{align}
    \frac{dJ_i}{d\xi}
    =
    -\eta
    \left\langle
    \mathrm{sgn}(\lambda)
    \tilde{\nu}_i
    \theta(-yy^*)
    \right\rangle,
\end{align}
and
\begin{align}
    \frac{dQ_i}{d\xi}
    =
    -2\eta
    \left\langle
    \mathrm{sgn}(\lambda)
    \lambda_i
    \theta(-yy^*)
    \right\rangle
    +
    \frac{\eta^2}{T}G_i
    \left\langle \theta(-yy^*)\right\rangle.
\end{align}
Using
\begin{align}
    \mathrm{sgn}(\lambda)\theta(-yy^*)
    =
    \frac{1}{2}
    \left[
    \mathrm{sgn}(\lambda)-y^*
    \right],
\end{align}
and averaging over the label noise gives
\begin{align}
    \left\langle X y^* \right\rangle
    =
    (1-2\epsilon)
    \left\langle X\mathrm{sgn}(\nu)\right\rangle.
\end{align}
The probability of matching the noisy label is
\begin{align}
    P_\epsilon
    =
    (1-2\epsilon)P+\epsilon.
\end{align}
Therefore,
\begin{align}
    \frac{dJ_i}{d\xi}
    &=
    -\frac{\eta}{2}
    \left(
    \left\langle \tilde{\nu}_i\mathrm{sgn}(\lambda)\right\rangle
    -
    (1-2\epsilon)
    \left\langle \tilde{\nu}_i\mathrm{sgn}(\nu)\right\rangle
    \right),
    \\
    \frac{dQ_i}{d\xi}
    &=
    -\eta
    \left(
    \left\langle \lambda_i\mathrm{sgn}(\lambda)\right\rangle
    -
    (1-2\epsilon)
    \left\langle \lambda_i\mathrm{sgn}(\nu)\right\rangle
    \right)
    +
    \frac{\eta^2}{T}G_i(1-P_\epsilon).
\end{align}
Substituting the Gaussian averages gives the closed SFT ODEs:
\begin{align}
    \frac{dJ_i}{d\xi}
    &=
    -\frac{\eta}{2}
    \sqrt{\frac{2}{\pi}}
    \left(
    \frac{C_i^J}{\sqrt{\widehat Q}}
    -
    (1-2\epsilon)
    \frac{S_i}{\sqrt{S}}
    \right),
    \\
    \frac{dQ_i}{d\xi}
    &=
    -\eta
    \sqrt{\frac{2}{\pi}}
    \left(
    \frac{C_i^Q}{\sqrt{\widehat Q}}
    -
    (1-2\epsilon)
    \frac{J_i}{\sqrt{S}}
    \right)
    +
    \frac{\eta^2}{T}G_i(1-P_\epsilon),
    \qquad i\in\{s,n\}.
    \label{eq:projected_sft_odes}
\end{align}

\subsection{Magnitude and dimension scaling of closure residuals}
\label{app:self_averaging}

Only two residual contractions need to be tracked, one for $J$ and one for $Q$. To see this, recall $\mathbf U_{ij}=\mathbf P\mathbf M_i\mathbf P\mathbf M_j\mathbf P$ and $\mathbf V_i=\mathbf P\mathbf M_i\mathbf P$. Since $\mathbf M_n=\mathbf I-\mathbf M_s$ and $\mathbf P^2=\mathbf P$,
\[
    \mathbf U_{sn}=\mathbf U_{ns}=\mathbf V_s-\mathbf U_{ss},
    \qquad
    \mathbf U_{nn}=\mathbf V_n-\mathbf V_s+\mathbf U_{ss}.
\]
The Hilbert--Schmidt projection onto $\mathrm{span}\{\mathbf V_s,\mathbf V_n\}$ is linear and leaves both basis operators unchanged. Subtracting the projected operators therefore gives $\mathbf\Delta_{sn}=\mathbf\Delta_{ns}=-\mathbf\Delta_{ss}$ and $\mathbf\Delta_{nn}=\mathbf\Delta_{ss}$. Consequently, the signed contraction residuals $r^{(J)}_{ij}=D^{-1}\mathbf w^{\top}\mathbf\Delta_{ij}\mathbf w^*$ and $r^{(Q)}_{ij}=D^{-1}\mathbf w^{\top}\mathbf\Delta_{ij}\mathbf w$ satisfy
\[
    r^{(C)}_{ss}=r^{(C)}_{nn}=-r^{(C)}_{sn}=-r^{(C)}_{ns},
    \qquad C\in\{J,Q\}.
\]
Thus $r^{(J)}_{ss}$ and $r^{(Q)}_{ss}$ determine all eight contraction residuals exactly, at every training time. These identities concern the signed residuals before relative normalization; the relative residuals can differ because their denominators depend on the contraction.

\begin{figure}[t]
    \centering
    \includegraphics[width=\textwidth]{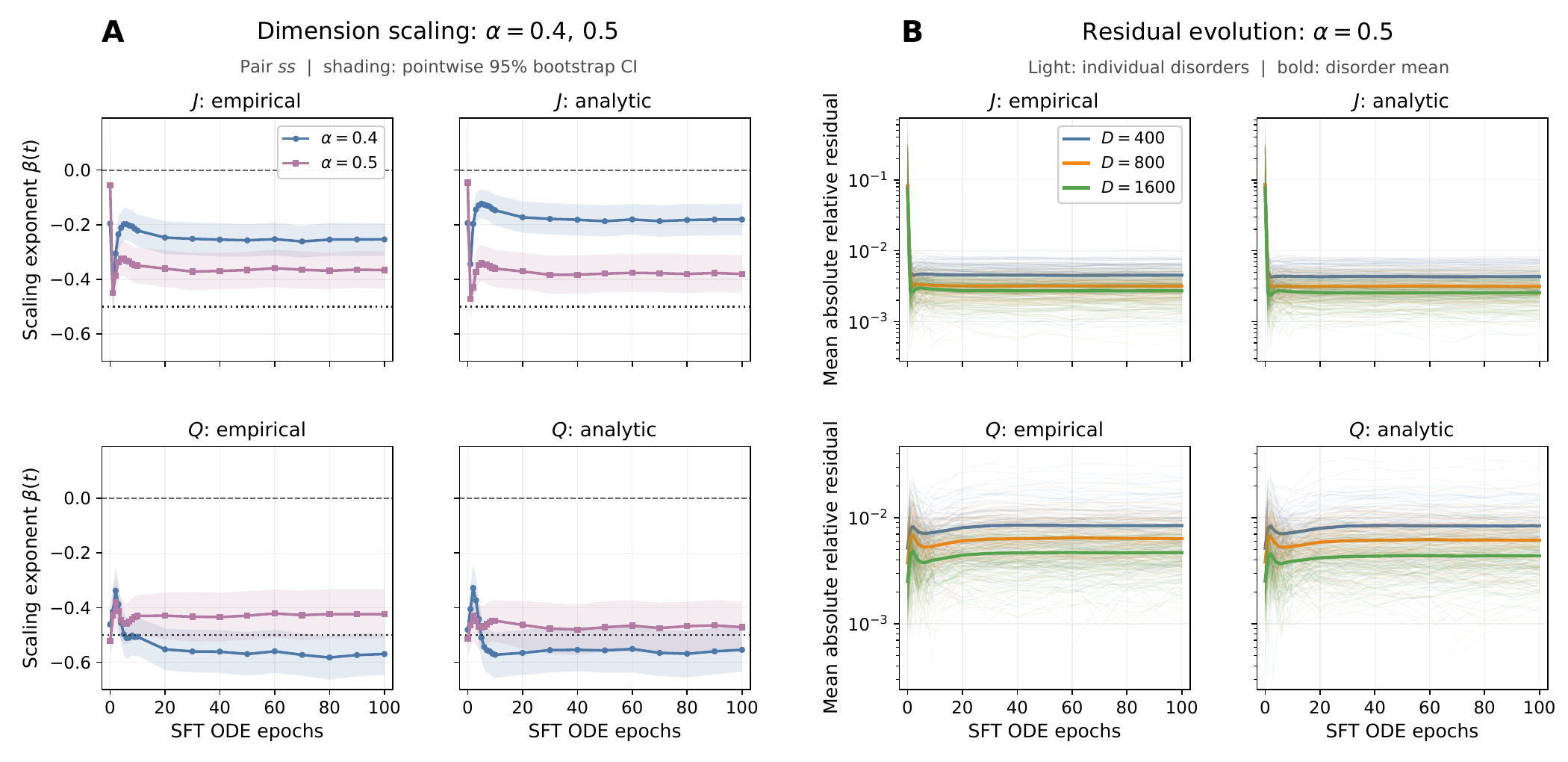}
    \caption{\textbf{Evolution and dimension scaling of closure residuals during spherical SFT in the projector limit.}
    Rows show the $J_{ss}$ and $Q_{ss}$ contractions; columns within each block use empirical Hilbert--Schmidt or analytic RMT coefficients.
    \textbf{(A)} Exponents fitted using $D=400,800,1600$ at $\alpha=0.4,0.5$, with pointwise 95\% bootstrap confidence intervals over disorders. Dotted and dashed lines mark $\beta=-1/2$ and $0$.
    \textbf{(B)} Absolute relative residuals at $\alpha=0.5$ (logarithmic axis). Faint curves average four PCA projections per fixed teacher/student disorder; bold curves additionally average over 128 disorders at each $D$.
    Parameters: $\epsilon=0$, $\eta=1$, $T=5$, $\rho_{\mathrm{noise}}=0.8$, $\Delta=0.1$. One ODE epoch comprises $D$ microscopic updates; no finite-pretraining dynamics are included.}
    \label{fig:projector-closure-residuals}
\end{figure}

To assess the Hilbert--Schmidt closure, we compare the exact signal--signal contractions $C_{\mathrm{exact}}\in\{J_{ss},Q_{ss}\}$ with their closed approximations $C_{\mathrm{closed}}=\Omega_{ss}^{(s)}C_s+\Omega_{ss}^{(n)}C_n$, using the symmetric relative residual
\begin{equation}
    r_{\mathrm{rel}}(\xi;D)
    = \frac{C_{\mathrm{exact}}-C_{\mathrm{closed}}}
    {|C_{\mathrm{exact}}|+|C_{\mathrm{closed}}|+10^{-12}}.
\end{equation}
Figure~\ref{fig:projector-closure-residuals}B shows that, after the initial transient, the mean absolute relative residual remains small and decreases with dimension. This motivates fitting $\langle|r_{\mathrm{rel}}(\xi;D)|\rangle\propto D^{\beta(\xi)}$ at each training time (panel A), with absolute values taken before averaging over realizations. Negative $\beta$ indicates decreasing residual magnitude over the tested dimensions; $\beta=-1/2$ is the usual $D^{-1/2}$ self-averaging benchmark. The late-time $Q$ scaling is broadly compatible with this benchmark, while $J$ shows a persistently shallower decrease. These finite-size results support the accuracy of the approximate closure in the simulated regime, but do not establish that the residuals self-average to zero as $D\to\infty$.

\subsection{Generalization Performance}

At any point in training, the probability that the student agrees with the clean teacher on a single example is
\begin{align}
    P
    =
    \frac{1}{2}
    +
    \frac{1}{\pi}
    \sin^{-1}
    \left(
    \frac{J_s+J_n}{\sqrt{S(Q_s+\Delta Q_n)}}
    \right).
\end{align}
This expression depends only on the total projected student-teacher overlap \(J_s+J_n\), the teacher norm \(S\), and the effective student variance \(Q_s+\Delta Q_n\).

\subsection{Optional Spherical Constraint}
\label{app:spherical_constraint}

If the student norm in the projected feature space is constrained to remain fixed,
\[
    Q_s+Q_n = R,
\]
then the unconstrained increments may be projected back onto the constraint manifold. If
\[
    dQ_{\mathrm{unc}} = dQ_s+dQ_n,
\]
then the constrained increments used in the implementation are
\begin{align}
    dJ_i
    &\leftarrow
    dJ_i
    -
    \frac{J_i}{2R}
    dQ_{\mathrm{unc}},
    \\
    dQ_i
    &\leftarrow
    dQ_i
    -
    \frac{Q_i}{R}
    dQ_{\mathrm{unc}},
    \qquad i\in\{s,n\}.
\end{align}
This is a Lagrange-multiplier projection that removes the radial component of the update while preserving the projected order-parameter dynamics tangential to the fixed-norm surface.

\subsection{Locally Optimal Learning Rate}
\label{App:LocallyOptimalLR}

In some experiments, rather than using a fixed learning rate, we choose a locally optimal learning rate at each integration step. The learning rate is chosen to maximize the instantaneous predicted increase in the clean single-step generalization probability
\begin{align}
    P
    =
    \frac{1}{2}
    +
    \frac{1}{\pi}
    \sin^{-1}
    \left(
    \frac{J_s+J_n}
    {\sqrt{S(Q_s+\Delta Q_n)}}
    \right).
\end{align}
For SFT with label noise, the dynamics use the noisy-label probability
\begin{align}
    P_\epsilon = (1-2\epsilon)P+\epsilon,
\end{align}
but the locally optimal learning rate is chosen to improve the clean probability \(P\). This corresponds to choosing the learning rate that gives the greatest predicted improvement in true teacher-student alignment, rather than in the noisy labels.

Let
\begin{align}
    \widehat Q = Q_s+\Delta Q_n,
    \qquad
    J = J_s+J_n.
\end{align}
The derivatives of \(P\) with respect to the order parameters are
\begin{align}
    \frac{\partial P}{\partial J_s}
    =
    \frac{\partial P}{\partial J_n}
    &=
    \frac{1}
    {\pi\sqrt{S\widehat Q}
    \sqrt{1-\frac{J^2}{S\widehat Q}}},
    \\
    \frac{\partial P}{\partial Q_s}
    &=
    -
    \frac{J}
    {2\pi \sqrt{S}\widehat Q^{3/2}
    \sqrt{1-\frac{J^2}{S\widehat Q}}},
    \\
    \frac{\partial P}{\partial Q_n}
    &=
    -
    \frac{\Delta J}
    {2\pi \sqrt{S}\widehat Q^{3/2}
    \sqrt{1-\frac{J^2}{S\widehat Q}}}.
\end{align}
For compactness, define
\begin{align}
    p_J
    &=
    \frac{\partial P}{\partial J_s}
    =
    \frac{\partial P}{\partial J_n},
    &
    p_{Q_s}
    &=
    \frac{\partial P}{\partial Q_s},
    &
    p_{Q_n}
    &=
    \frac{\partial P}{\partial Q_n}.
\end{align}

For either SFT or RL, the order-parameter ODEs can be written in the form
\begin{align}
    \frac{dJ_i}{d\xi}
    &=
    \eta A_{J_i},
    \\
    \frac{dQ_i}{d\xi}
    &=
    \eta A_{Q_i}+\eta^2 B_{Q_i},
    \qquad i\in\{s,n\},
\end{align}
where \(A_{J_i}\), \(A_{Q_i}\), and \(B_{Q_i}\) are the drift and diffusion coefficients appropriate to the current finetuning rule. Explicitly, for SFT,
\begin{align}
    A_{J_i}^{\mathrm{SFT}}
    &=
    -\frac{1}{2}
    \left(
    \left\langle \tilde{\nu}_i\mathrm{sgn}(\lambda)\right\rangle
    -
    (1-2\epsilon)
    \left\langle \tilde{\nu}_i\mathrm{sgn}(\nu)\right\rangle
    \right),
    \\
    A_{Q_i}^{\mathrm{SFT}}
    &=
    -
    \left(
    \left\langle \lambda_i\mathrm{sgn}(\lambda)\right\rangle
    -
    (1-2\epsilon)
    \left\langle \lambda_i\mathrm{sgn}(\nu)\right\rangle
    \right),
    \\
    B_{Q_i}^{\mathrm{SFT}}
    &=
    \frac{G_i}{T}(1-P_\epsilon),
\end{align}
whereas for RL,
\begin{align}
    A_{J_i}^{\mathrm{RL}}
    &=
    \frac{1}{2}
    \left(
    \left\langle \tilde{\nu}_i\mathrm{sgn}(\lambda)\right\rangle
    +
    \left\langle \tilde{\nu}_i\mathrm{sgn}(\nu)\right\rangle
    \right)
    P^{T-1},
    \\
    A_{Q_i}^{\mathrm{RL}}
    &=
    \left(
    \left\langle \lambda_i\mathrm{sgn}(\lambda)\right\rangle
    +
    \left\langle \lambda_i\mathrm{sgn}(\nu)\right\rangle
    \right)
    P^{T-1},
    \\
    B_{Q_i}^{\mathrm{RL}}
    &=
    \frac{G_i}{T}P^T.
\end{align}
Here
\begin{align}
    G_s = T_{ss}+\Delta T_{sn},
    \qquad
    G_n = T_{sn}+\Delta T_{nn}.
\end{align}

To choose the learning rate, we locally linearize \(p\) in the order parameters while retaining the quadratic dependence on \(\eta\) coming from the \(Q_i\) diffusion terms. For a small integration step \(d\xi\),
\begin{align}
    \Delta P
    &\approx
    d\xi
    \left[
    p_J(dJ_s+dJ_n)
    +
    p_{Q_s}dQ_s
    +
    p_{Q_n}dQ_n
    \right].
\end{align}
Substituting the unconstrained increments gives
\begin{align}
    \Delta P(\eta)
    &\approx
    d\xi
    \left[
    b\eta+a\eta^2
    \right],
\end{align}
where
\begin{align}
    b
    &=
    p_J(A_{J_s}+A_{J_n})
    +
    p_{Q_s}A_{Q_s}
    +
    p_{Q_n}A_{Q_n},
    \\
    a
    &=
    p_{Q_s}B_{Q_s}
    +
    p_{Q_n}B_{Q_n}.
\end{align}
Thus the locally optimal learning rate is obtained by maximizing the quadratic
\begin{align}
    f(\eta)=b\eta+a\eta^2
\end{align}
over the allowed interval \(0<\eta\leq \eta_{\max}\). In the implementation we take \(\eta_{\max}=1\), and use a small positive floor \(\eta_{\min}\) to avoid exactly zero updates.

If \(a<0\) and \(b>0\), the quadratic is locally concave and initially increasing, so the unconstrained maximizer is
\begin{align}
    \eta_* = -\frac{b}{2a}.
\end{align}
If the quadratic is convex, or if the vertex is not a useful positive maximizer, the optimum over a bounded interval lies on the boundary. The implemented rule is
\begin{align}
    \eta_{\mathrm{opt}}
    =
    \min\left\{
    \eta_{\max},
    \begin{cases}
        -\dfrac{b}{2a},
        & a<0 \text{ and } b>0, \\[1.2em]
        \max\left\{-\dfrac{b}{a},0\right\}+\eta_{\min},
        & a>0 \text{ and } b<0, \\[1.2em]
        \eta_{\max},
        & b>0, \\[0.4em]
        \eta_{\min},
        & \text{otherwise}.
    \end{cases}
    \right.
\end{align}
This is the learning rate used for the next Euler step.

\paragraph{Spherical constraint.}
When the spherical constraint is enabled, the projected norm
\begin{align}
    Q_s+Q_n=R
\end{align}
is held fixed by removing the radial component of the update. Let the unconstrained \(Q\)-increment be
\begin{align}
    dQ_{\mathrm{unc}}
    =
    dQ_s+dQ_n.
\end{align}
The projected increments are
\begin{align}
    dJ_i
    &\leftarrow
    dJ_i
    -
    \frac{J_i}{2R}dQ_{\mathrm{unc}},
    \\
    dQ_i
    &\leftarrow
    dQ_i
    -
    \frac{Q_i}{R}dQ_{\mathrm{unc}},
    \qquad i\in\{s,n\}.
\end{align}
Since
\begin{align}
    dQ_{\mathrm{unc}}
    =
    \eta(A_{Q_s}+A_{Q_n})
    +
    \eta^2(B_{Q_s}+B_{Q_n}),
\end{align}
the spherical projection modifies both the linear and quadratic coefficients of the local objective.

Define
\begin{align}
    A_Q^{\mathrm{tot}}
    &=
    A_{Q_s}+A_{Q_n},
    &
    B_Q^{\mathrm{tot}}
    &=
    B_{Q_s}+B_{Q_n}.
\end{align}
The constrained linear coefficient is
\begin{align}
    b_{\mathrm{con}}
    &=
    p_J
    \left[
    A_{J_s}
    -
    \frac{J_s}{2R}A_Q^{\mathrm{tot}}
    \right]
    +
    p_J
    \left[
    A_{J_n}
    -
    \frac{J_n}{2R}A_Q^{\mathrm{tot}}
    \right]
    \nonumber \\
    &\quad
    +
    p_{Q_s}
    \left[
    A_{Q_s}
    -
    \frac{Q_s}{R}A_Q^{\mathrm{tot}}
    \right]
    +
    p_{Q_n}
    \left[
    A_{Q_n}
    -
    \frac{Q_n}{R}A_Q^{\mathrm{tot}}
    \right].
\end{align}
The constrained quadratic coefficient is
\begin{align}
    a_{\mathrm{con}}
    &=
    -
    p_J
    \frac{J_s}{2R}B_Q^{\mathrm{tot}}
    -
    p_J
    \frac{J_n}{2R}B_Q^{\mathrm{tot}}
    \nonumber \\
    &\quad
    +
    p_{Q_s}
    \left[
    B_{Q_s}
    -
    \frac{Q_s}{R}B_Q^{\mathrm{tot}}
    \right]
    +
    p_{Q_n}
    \left[
    B_{Q_n}
    -
    \frac{Q_n}{R}B_Q^{\mathrm{tot}}
    \right].
\end{align}
When the spherical constraint is used, the same quadratic maximization rule is applied with
\begin{align}
    b \leftarrow b_{\mathrm{con}},
    \qquad
    a \leftarrow a_{\mathrm{con}}.
\end{align}
Otherwise, the unconstrained coefficients \(b\) and \(a\) are used.

\begin{figure}
    \centering
\includegraphics[width=\linewidth]{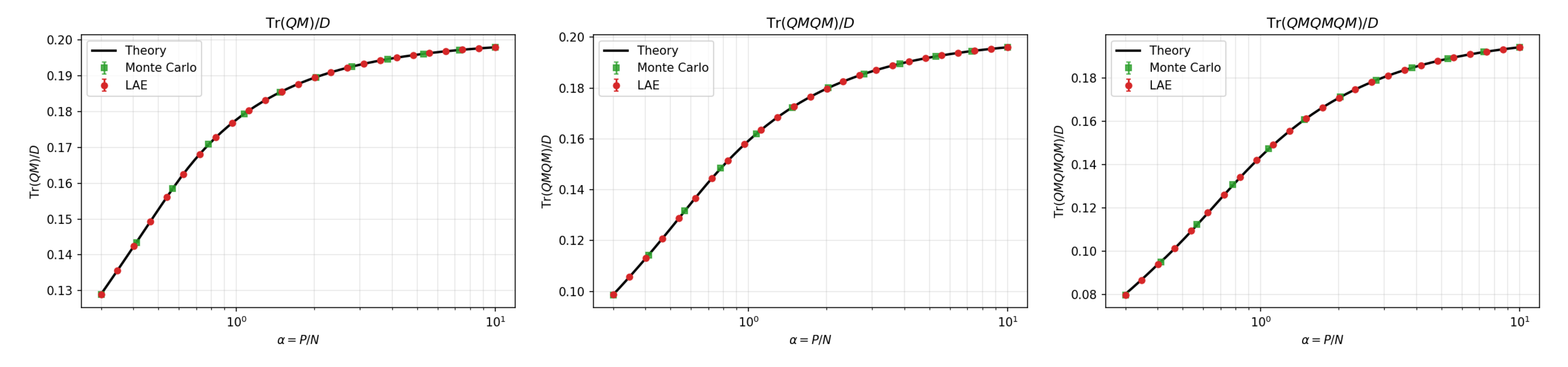}
        \caption{The expected normalised traces of powers of $\mathbf{P}\mathbf{M}_s$, calculated by theory (solid), PCA Monte carlo simulation (orange) and trained LAE encoder projection for different batch sizes (blue). \textit{Parameters}: $N=2000,\quad (k = 0.2), \Delta=0.1$}
        \label{supp_fig:subspace_overlap}
\end{figure}

\section{Expected Trace Constants for the Random PCA Projector}
\label{App:ExpectedTraceConstants}

The projected dynamics in Appendix~\ref{App:TheoryProjected} depend on the
static trace constants
\[
    t,\qquad t_s,\qquad T_{ss},\qquad \tau_{sss},
\]
where
\begin{align}
    t
    &= \frac{1}{D}\Tr(\mathbf P),
    &
    t_s
    &= \frac{1}{D}\Tr(\mathbf P\mathbf M_s),
    \\
    T_{ss}
    &= \frac{1}{D}
    \Tr\!\left(
        \mathbf P\mathbf M_s\mathbf P\mathbf M_s
    \right),
    &
    \tau_{sss}
    &= \frac{1}{D}
    \Tr\!\left(
        \mathbf P\mathbf M_s\mathbf P\mathbf M_s\mathbf P\mathbf M_s
    \right).
\end{align}
These quantities depend only on the empirical PCA projector \(\mathbf P\)
learned during pretraining.  In the proportional high-dimensional limit they are
self-averaging, so in the typical-case ODEs we replace them by their deterministic
limits, or, equivalently, by their expectations over the pretraining dataset.

Let the ambient dimension be \(D\).  The signal and noise projectors satisfy
\begin{align}
    \rank(\mathbf M_s)=\rho D,
    \qquad
    \rank(\mathbf M_n)=(1-\rho)D,
    \qquad
    \mathbf M_s+\mathbf M_n=\mathbf I_D,
\end{align}
with \(0<\rho<1\).  The population covariance is
\begin{align}
    \Sigma=\mathbf M_s+\Delta \mathbf M_n,
    \qquad 0<\Delta<1.
\end{align}
The pretraining data matrix is
\begin{align}
    \mathbf X=\mathbf Z\Sigma^{1/2},
    \qquad
    Z_{\mu i}\sim \mathcal N(0,1),
    \qquad
    \mu=1,\dots,N,
\end{align}
and the empirical covariance is
\begin{align}
    \widehat\Sigma
    =
    \frac{1}{N}\mathbf X^\top\mathbf X.
\end{align}
The PCA projector \(\mathbf P\) is the orthogonal projector onto the top
\(\rho D\) empirical eigenvectors of \(\widehat\Sigma\).  Hence
\begin{align}
    t
    =
    \frac{1}{D}\Tr(\mathbf P)
    =
    \rho
\end{align}
for every realization of the pretraining data. We work in the Marchenko--Pastur scaling
\begin{align}
    D,N\to\infty,
    \qquad
    \gamma=\frac{D}{N} \text{, fixed.}
\end{align}
We assume that the selected top-\(\rho D\) empirical eigenspace lies in
the non-zero empirical spectrum.  If \(N<\rho D\), then the projector necessarily
includes directions from the null space of \(\widehat\Sigma\); in that case the
trace constants acquire additional convention-dependent null-space contributions (we do not consider this regime).

\subsection{Resolvent representation of the PCA projector}

Define the resolvent as 
\begin{align}
    \mathbf R_{\widehat\Sigma}(z)
    =
    (\widehat\Sigma-z\mathbf I)^{-1}.
\end{align}
If
\[
    \widehat\Sigma
    =
    \sum_{a=1}^D
    \widehat\lambda_a
    \hat{\mathbf u}_a\hat{\mathbf u}_a^\top,
\]
then we can write
\[
    \mathbf R_{\widehat\Sigma}(z)
    =
    \sum_{a=1}^D
    \frac{
        \hat{\mathbf u}_a\hat{\mathbf u}_a^\top
    }{
        \widehat\lambda_a-z
    }.
\]
Let \(I\) denote the limiting empirical eigenvalue window selected by the PCA
projector.  Equivalently, \(I\) is the upper spectral set satisfying
\begin{align}
    \int_I \varrho(x)\,dx=\rho,
    \label{eq:top_window_mass}
\end{align}
where \(\varrho\) is the limiting empirical spectral density.  If the top
spectral component is separated from the rest of the spectrum, then \(I\) is the
support of that top component.  If the spectrum is gapless, then \(I\) is the
upper \(\rho\)-quantile of the limiting spectrum.

By Cauchy's functional calculus, For an eigenvalue $\widehat\lambda_a$ inside the PCA window $I$,
\[
\oint_{\Gamma_a} \frac{dz}{\widehat\lambda_a-z} = -2\pi i,\]

where $\Gamma_a$ is a positively oriented contour enclosing $\widehat\lambda_a$ and no other eigenvalues.  Therefore, when the selected spectral set is separated by a macroscopic gap, this projector
has the contour representation

\begin{align}
    -\frac{1}{2\pi i}
    \oint_{\Gamma_I}
    \mathbf R_{\widehat\Sigma}(z)\,dz =
    \sum_{\widehat\lambda_a \in I}
    \hat{\mathbf u}_a\hat{\mathbf u}_a^\top= \mathbf P,
    \label{eq:projector_contour}
\end{align}
where \(\Gamma_I\) is a positively oriented contour enclosing the spectral set $I$, and no other part of the spectrum.  The minus sign appears because
we use the convention
\(\mathbf R_{\widehat\Sigma}(z)=(\widehat\Sigma-z\mathbf I)^{-1}\). 
This contour representation
is useful because each trace constant can be written as a contour integral of a
resolvent chain, as below:
\begin{align}
    t_s
    &=
    -\frac{1}{2\pi i}
    \oint_{\Gamma_I}
    \frac{1}{D}
    \Tr\!\left(
        \mathbf M_s\mathbf R_{\widehat\Sigma}(z)
    \right)
    dz,
    \\
    T_{ss}
    &=
    \frac{1}{(2\pi i)^2}
    \oint_{\Gamma_I}
    \oint_{\Gamma_I}
    \frac{1}{D}
    \Tr\!\left(
        \mathbf M_s\mathbf R_{\widehat\Sigma}(z_1)
        \mathbf M_s\mathbf R_{\widehat\Sigma}(z_2)
    \right)
    dz_1dz_2,
    \\
    \tau_{sss}
    &=
    -\frac{1}{(2\pi i)^3}
    \oint_{\Gamma_I}
    \oint_{\Gamma_I}
    \oint_{\Gamma_I}
    \frac{1}{D}
    \Tr\!\left(
        \mathbf M_s\mathbf R_{\widehat\Sigma}(z_1)
        \mathbf M_s\mathbf R_{\widehat\Sigma}(z_2)
        \mathbf M_s\mathbf R_{\widehat\Sigma}(z_3)
    \right)
    dz_1dz_2dz_3.
    \label{eq:tau_contour_resolvent}
\end{align}
The contour integral is simply a way of replacing the discontinuous operation “keep only eigenvectors in the PCA window” by an analytic operation on the resolvent. Random matrix theory can compute deterministic limits of resolvent traces much more directly than deterministic limits of eigenvector projectors. Thus the problem reduces to evaluating deterministic equivalents for one-,
two-, and three-resolvent traces with deterministic insertions
\(\mathbf M_s\).  This is the standard route in random matrix theory for
extracting eigenvector information from resolvents; see, for example,
\citet{silverstein1995analysis},
\citet{ledoit2011eigenvectors}. N.B. \textbf{the contour formula is clean only in the separated case}, i.e. where the spectrum is gapped ($\gamma < 1$). If the selected window $I$ lies inside a connected bulk component, there is no fixed macroscopic contour enclosing the selected eigenvalues and no other spectrum. At finite $D$, there is technically a tiny random gap between consecutive eigenvalues, but it vanishes as $D \rightarrow \infty$. In the gapless quantile-selection regime, the contour picture should be understood
as the limiting boundary-value representation obtained by replacing the contour
integral by the spectral discontinuity of the resolvent ($\mathbf R_{\widehat\Sigma}(x+i0) - \mathbf R_{\widehat\Sigma}(x-i0)$) across the interval \(I\), we detail this in the subsequent sections.

In the Marchenko--Pastur limit, the resolvent traces self-average, so the trace constants can be replaced by their deterministic limits.  The purpose of the Matrix Dyson Equation (next section) is to compute these deterministic limits.

\subsection{Matrix Dyson Equation}

The random matrix
\begin{align}
    \widehat{\Sigma}
    =
    \frac{1}{N}\mathbf X^\top \mathbf X
\end{align}
depends quadratically on the Gaussian data matrix \(\mathbf X\).  Consequently,
its resolvent
\begin{align}
    \mathbf R_{\widehat\Sigma}(z)
    =
    (\widehat{\Sigma}-z\mathbf I)^{-1}
\end{align}
is a nonlinear function of the Gaussian randomness.  The purpose of the Matrix
Dyson Equation, or MDE, is to replace this random resolvent, in the proportional
high-dimensional limit, by a deterministic matrix-valued function (\(\mathbf R_{\widehat\Sigma}(z) \rightarrow \boldsymbol\Pi_{\widehat{\Sigma}}(z)\)). (More general methods already exist for doing this \cite{atanasov2024scaling}, but we require only a specific case of this for our simplified block structure).  In this
sense, the MDE plays the role of a law of large numbers for resolvents.

It is convenient not to apply the MDE directly to the quadratic matrix
\(\widehat{\Sigma}\), but instead to linearise it.  Define
\begin{align}
    \mathcal H
    =
    \begin{pmatrix}
        0 & \mathbf X^\top/\sqrt N \\
        \mathbf X/\sqrt N & 0
    \end{pmatrix}
    \in
    \mathbb R^{(D+N)\times(D+N)} .
\end{align}
The matrix \(\mathcal H\) is linear in the Gaussian data.  Moreover,
\(\mathcal H^2\) has upper-left block
\begin{align}
    \frac{1}{N}\mathbf X^\top\mathbf X
    =
    \widehat{\Sigma}.
\end{align}
Thus the nonzero eigenvalues of \(\widehat{\Sigma}\) are the squares of the
corresponding singular-value eigenvalues of the upper block of \(\mathcal H\).  This is the reason
for introducing \(\mathcal H\): it turns the quadratic covariance problem into a
linear random-matrix problem.

Let
\begin{align}
    \mathbf R_{\mathcal H}(\zeta)
    =
    (\mathcal H-\zeta\mathbf I)^{-1}
\end{align}
be the resolvent of the linearised matrix.  We denote its deterministic
equivalent by \(\boldsymbol\Pi_{\mathcal H}(\zeta)\). Heuristically, this deterministic
equivalent is obtained by averaging the resolvent identity and solving for the expected resolvent:
\begin{align}
    (\mathcal H-\zeta\mathbf I)
    \mathbf R_{\mathcal H}(\zeta)
    &=
    \mathbf I \\
    \mathbb E[\mathcal H \mathbf R_{\mathcal H}(\zeta)] -\zeta\mathbb E [\mathbf R_{\mathcal H}(\zeta)] &= 
    \mathbf I.
\end{align}
If \(\mathcal H \) were deterministic we could just solve for 
\(\boldsymbol\Pi_{\mathcal H}(\zeta)\), but \(\mathcal H\) is random and \(\boldsymbol R_{\mathcal H} \) depends on \(\mathcal H\). Hence, we are left with the problem of evaluating the correlation
\(\mathbb E[\mathcal H\mathbf R_{\mathcal H}]\).

\subsection{Derivation of the Matrix Dyson Equation}

Let
\begin{align}
\mathbf{R}(\zeta)
&:=
\mathbf{R}_{\mathcal{H}}(\zeta)
=
\left(
\mathcal{H}-\zeta\mathbf{I}_{D+N}
\right)^{-1},
\\
\mathbf{m}(\zeta)
&:=
\mathbb{E}\!\left[\mathbf{R}(\zeta)\right].
\end{align}
The deterministic equivalent
\(\boldsymbol{\Pi}(\zeta)\) is defined below as the solution of the
corresponding Matrix Dyson Equation. In the proportional
high-dimensional limit, one expects
\begin{align}
\mathbf{m}(\zeta)-\boldsymbol{\Pi}(\zeta)
\longrightarrow 0.
\end{align}

Define
\begin{align}
A_{\mu i}
&:=
\frac{X_{\mu i}}{\sqrt{N}},
\\
\mathbf{B}_{\mu i}
&:=
\mathbf{E}_{i,D+\mu}
+
\mathbf{E}_{D+\mu,i},
\end{align}
where \(\mathbf{E}_{ab}\) denotes the matrix with a \(1\) in entry
\((a,b)\) and zeros elsewhere. Then
\begin{align}
\mathcal{H}
=
\sum_{\mu=1}^{N}
\sum_{i=1}^{D}
A_{\mu i}\mathbf{B}_{\mu i}.
\end{align}

The random variables \(A_{\mu i}\) are independent,
centred Gaussian variables satisfying
\begin{align}
\mathbb{E}\!\left[A_{\mu i}^{2}\right]
&=
\frac{\sigma_i}{N},
\\
\sigma_i
&=
\begin{cases}
1,
&
 i\in\mathcal{I}_{\mathrm{signal}},
\\
\Delta,
&
 i\in\mathcal{I}_{\mathrm{noise}}.
\end{cases}
\end{align}

We begin by expanding
\begin{align}
\mathbb{E}\!\left[
\mathcal{H}\mathbf{R}(\zeta)
\right]
&=
\sum_{\mu=1}^{N}
\sum_{i=1}^{D}
\mathbb{E}\!\left[
A_{\mu i}
\mathbf{B}_{\mu i}
\mathbf{R}(\zeta)
\right]
\nonumber\\
&=
\sum_{\mu=1}^{N}
\sum_{i=1}^{D}
\mathbf{B}_{\mu i}
\mathbb{E}\!\left[
A_{\mu i}\mathbf{R}(\zeta)
\right],
\label{eq:expected-HR-expansion}
\end{align}
where we used the fact that \(\mathbf{B}_{\mu i}\) is deterministic.

Gaussian integration by parts gives
\begin{align}
\mathbb{E}\!\left[
A_{\mu i}\mathbf{R}(\zeta)
\right]
=
\frac{\sigma_i}{N}
\mathbb{E}\!\left[
\frac{\partial\mathbf{R}(\zeta)}
{\partial A_{\mu i}}
\right].
\label{eq:gaussian-ibp}
\end{align}

Since
\begin{align}
\mathbf{R}(\zeta)
=
\left(
\mathcal{H}-\zeta\mathbf{I}_{D+N}
\right)^{-1},
\end{align}
the derivative of the inverse matrix gives
\begin{align}
\frac{\partial\mathbf{R}(\zeta)}
{\partial A_{\mu i}}
&=
-
\mathbf{R}(\zeta)
\frac{\partial\mathcal{H}}
{\partial A_{\mu i}}
\mathbf{R}(\zeta)
\nonumber\\
&=
-
\mathbf{R}(\zeta)
\mathbf{B}_{\mu i}
\mathbf{R}(\zeta).
\label{eq:resolvent-derivative}
\end{align}
Therefore,
\begin{align}
\mathbb{E}\!\left[
A_{\mu i}\mathbf{R}(\zeta)
\right]
=
-
\frac{\sigma_i}{N}
\mathbb{E}\!\left[
\mathbf{R}(\zeta)
\mathbf{B}_{\mu i}
\mathbf{R}(\zeta)
\right].
\label{eq:ibp-resolvent}
\end{align}

Substituting \ref{eq:ibp-resolvent} into
\ref{eq:expected-HR-expansion}, we obtain
\begin{align}
\mathbb{E}\!\left[
\mathcal{H}\mathbf{R}(\zeta)
\right]
=
-
\sum_{\mu=1}^{N}
\sum_{i=1}^{D}
\frac{\sigma_i}{N}
\mathbf{B}_{\mu i}
\mathbb{E}\!\left[
\mathbf{R}(\zeta)
\mathbf{B}_{\mu i}
\mathbf{R}(\zeta)
\right].
\label{eq:expected-HR-after-ibp}
\end{align}

Define the linear self-energy operator by
\begin{align}
\mathcal{S}[\mathbf{T}]
:=
\sum_{\mu=1}^{N}
\sum_{i=1}^{D}
\frac{\sigma_i}{N}
\mathbf{B}_{\mu i}
\mathbf{T}
\mathbf{B}_{\mu i},
\label{eq:self-energy-definition}
\end{align}
for every
\(\mathbf{T}\in\mathbb{C}^{(D+N)\times(D+N)}\).

When \(\mathbf{T}\) is deterministic, this can equivalently be written as
\begin{align}
\mathcal{S}[\mathbf{T}]
=
\mathbb{E}\!\left[
\mathcal{H}\mathbf{T}\mathcal{H}
\right].
\end{align}

Using the definition of \(\mathcal{S}\), equation
\ref{eq:expected-HR-after-ibp} becomes
\begin{align}
\boxed{
\mathbb{E}\!\left[
\mathcal{H}\mathbf{R}(\zeta)
\right]
=
-
\mathbb{E}\!\left[
\mathcal{S}[\mathbf{R}(\zeta)]
\mathbf{R}(\zeta)
\right]
}
\label{eq:exact-self-energy-identity}
\end{align}

This is an exact identity for any centred Gaussian model. No deterministic-equivalent
approximation has yet been made.

The resolvent identity
\begin{align}
\left(
\mathcal{H}-\zeta\mathbf{I}_{D+N}
\right)
\mathbf{R}(\zeta)
=
\mathbf{I}_{D+N}
\end{align}
implies
\begin{align}
\mathbb{E}\!\left[
\mathcal{H}\mathbf{R}(\zeta)
\right]
-
\zeta\mathbf{m}(\zeta)
=
\mathbf{I}_{D+N}.
\label{eq:averaged-resolvent-identity}
\end{align}

Combining \ref{eq:exact-self-energy-identity} and
\ref{eq:averaged-resolvent-identity}, we find
\begin{align}
-
\mathbb{E}\!\left[
\mathcal{S}[\mathbf{R}(\zeta)]
\mathbf{R}(\zeta)
\right]
-
\zeta\mathbf{m}(\zeta)
=
\mathbf{I}_{D+N}.
\label{eq:exact-averaged-equation}
\end{align}

To isolate the fluctuation term, write
\begin{align}
\mathbf{R}(\zeta)
&=
\mathbf{m}(\zeta)
+
\mathbf{Q}(\zeta),
\\
\mathbb{E}\!\left[\mathbf{Q}(\zeta)\right]
&=
\mathbf{0}.
\end{align}
Since \(\mathcal{S}\) is deterministic and linear,
\begin{align}
\mathbb{E}\!\left[
\mathcal{S}[\mathbf{R}]\mathbf{R}
\right]
=
\mathcal{S}[\mathbf{m}]\mathbf{m}
+
\mathbb{E}\!\left[
\mathcal{S}[\mathbf{Q}]\mathbf{Q}
\right].
\label{eq:self-energy-fluctuation-decomposition}
\end{align}

Consequently,
\begin{align}
-
\left(
\zeta\mathbf{I}_{D+N}
+
\mathcal{S}[\mathbf{m}(\zeta)]
\right)
\mathbf{m}(\zeta)
=
\mathbf{I}_{D+N}
+
\mathbb{E}\!\left[
\mathcal{S}[\mathbf{Q}(\zeta)]
\mathbf{Q}(\zeta)
\right].
\label{eq:perturbed-MDE}
\end{align}

In the proportional high-dimensional limit, concentration and fluctuation
averaging are used to show that the final term in
\ref{eq:perturbed-MDE} is negligible in an appropriate sense. This
motivates defining \(\boldsymbol{\Pi}(\zeta)\) as the solution of
\begin{align}
-
\left(
\zeta\mathbf{I}_{D+N}
+
\mathcal{S}[\boldsymbol{\Pi}(\zeta)]
\right)
\boldsymbol{\Pi}(\zeta)
=
\mathbf{I}_{D+N}.
\label{eq:MDE-product-form}
\end{align}

Multiplying \ref{eq:MDE-product-form} from the right by
\(\boldsymbol{\Pi}(\zeta)^{-1}\) gives
\begin{align}
\boxed{
-
\boldsymbol{\Pi}(\zeta)^{-1}
=
\zeta\mathbf{I}_{D+N}
+
\mathcal{S}[\boldsymbol{\Pi}(\zeta)]
}
\label{eq:MDE}
\end{align}

This is the Matrix Dyson Equation (MDE), and we have just shown that its solution is the deterministic equivalent of the resolvent of the linearised covariance matrix \(\mathcal H\).

\subsubsection{Solving the Matrix Dyson Equation to find the deterministic Resolvent (In our block symmetric case)}
We now compute the action of the self-energy operator \(\mathcal S\) on
block-constant diagonal matrices of the form
\begin{align}
    \operatorname{diag}
    \bigl(
        x_s\mathbf I_{\rho D},
        x_n\mathbf I_{(1-\rho)D},
        x_0\mathbf I_N
    \bigr),
\end{align}
the operator \(\mathcal S\) acts as
\begin{align}
    \mathcal S:
    (x_s,x_n,x_0)
    \mapsto
    \left(
        x_0,\,
        \Delta x_0,\,
        \gamma
        \bigl[
            \rho x_s
            +
            (1-\rho)\Delta x_n
        \bigr]
    \right).
    \label{eq:self_energy_action}
\end{align}
This can be shown from the definition of the self-energy operator and the covariance structure of the problem.  A signal coordinate
has variance \(1/N\) against each sample index, a noise coordinate has variance
\(\Delta/N\), and the sample block receives the average contribution from all
signal and noise coordinates.  Since \(D/N\to\gamma\), the sample-block term is
\begin{align}
    \gamma
    \bigl[
        \rho x_s
        +
        (1-\rho)\Delta x_n
    \bigr].
\end{align}

Because the model is invariant under rotations inside the signal subspace,
inside the noise subspace, and inside the \(N\)-dimensional sample space, the
deterministic resolvent cannot distinguish individual directions within each of
these three blocks.  Hence the MDE solution has the form
\begin{align}
    \boldsymbol\Pi_{\mathcal H}(\zeta)
    =
    \operatorname{diag}
    \left(
        a_s(\zeta)\mathbf I_{\rho D},
        a_n(\zeta)\mathbf I_{(1-\rho)D},
        b(\zeta)\mathbf I_N
    \right).
\end{align}
Substituting this ansatz and using the action of the self-energy operator \ref{eq:self_energy_action} into the MDE \ref{eq:MDE} gives the three scalar
self-consistency equations
\begin{align}
    -\frac{1}{a_s}
    &=
    \zeta+b,
    \\
    -\frac{1}{a_n}
    &=
    \zeta+\Delta b,
    \\
    -\frac{1}{b}
    &=
    \zeta+
    \gamma
    \bigl[
        \rho a_s
        +
        (1-\rho)\Delta a_n
    \bigr].
    \label{eq:scalar_MDE}
\end{align}
These equations determine \(\boldsymbol\Pi_{\mathcal H}(\zeta)\), the deterministic resolvent of the linearised matrix
\(\mathcal H\). But all of this was in order to find the deterministic resolvent of the empirical covariance matrix \(\widehat{\Sigma}\). So we now translate this back to the covariance resolvent
\begin{align}
    \mathbf R_{\widehat{\Sigma}}(\zeta)
    =
    (\widehat{\Sigma}-\zeta^2\mathbf I)^{-1}.
\end{align}
The link between this quantity and \(\mathbf \Pi(\zeta)\) is the upper-left block of the linearised resolvent.  Applying the
Schur complement to \(\mathcal H-\zeta\mathbf I\) gives
\begin{align}
    \left[
        (\mathcal H-\zeta\mathbf I)^{-1}
    \right]_{\mathrm{upper}\text{-}\mathrm{left}}
    =
    \zeta
    \left(
        \widehat{\Sigma}
        -
        \zeta^2\mathbf I
    \right)^{-1}.
\end{align}
Thus, with \(z=\zeta^2\),
\begin{align}
    \left[
        (\mathcal H-\zeta\mathbf I)^{-1}
    \right]_{\mathrm{upper}\text{-}\mathrm{left}}
    =
    \zeta\mathbf R_{\widehat{\Sigma}}(z).
    \label{eq:linearized_to_covariance_resolvent}
\end{align}

The variables \(a_s\) and \(a_n\) are therefore not themselves the block values
of the covariance resolvent.  They are the signal and noise block values of the
upper-left block of the linearised resolvent, which is \(\zeta\) times the
covariance resolvent.

Since the upper-left block of \(\boldsymbol\Pi_{\mathcal H}(\zeta)\) is
\begin{align}
    a_s(\zeta)\mathbf M_s
    +
    a_n(\zeta)\mathbf M_n,
\end{align}
the deterministic equivalent of the covariance resolvent has the block form
\begin{align}
    \mathbf \Pi_{\widehat{\Sigma}}(z)
    =
    g_s(z)\mathbf M_s
    +
    g_n(z)\mathbf M_n,
    \label{eq:Resolvent}
\end{align}
where
\begin{align}
    g_s(z)
    &=
    \frac{a_s(\zeta)}{\zeta},
    \\
    g_n(z)
    &=
    \frac{a_n(\zeta)}{\zeta},
    \qquad
    z=\zeta^2 .
    \label{eq:block_resolvent_functions}
\end{align}

Finally, introduce the scalar variable
\begin{align}
    \xi(z)
    =
    -\zeta b(\zeta).
    \label{eq:xi_definition}
\end{align}
Using the first two scalar MDE equations, one obtains
\begin{align}
    a_s(\zeta)
    &=
    \frac{\zeta}{\xi(z)-z},
    \\
    a_n(\zeta)
    &=
    \frac{\zeta}{\Delta\xi(z)-z},
    \\
    b(\zeta)
    &=
    -\frac{\xi(z)}{\zeta}.
    \label{eq:linearized_block_variables}
\end{align}
Therefore
\begin{align}
    g_s(z)
    &=
    \frac{1}{\xi(z)-z},
    \\
    g_n(z)
    &=
    \frac{1}{\Delta\xi(z)-z}.
\end{align}
These are the signal and noise block values of the deterministic equivalent of
the covariance resolvent.  They are the one-resolvent quantities used in the
subsequent sections to compute the trace constants
\begin{align}
    t_s,
    \qquad
    T_{ss},
    \qquad
    \tau_{sss}.
\end{align}

\subsection{One-resolvent quantities and the local signal overlap}
\label{app:one_resolvent}

We use the previously derived deterministic equivalent of the resolvent to compute the limiting empirical spectral density and the expected signal overlap of empirical eigenvectors.  The local signal overlap is the squared projection of an empirical eigenvector onto the true signal subspace, as a function of the  corresponding empirical eigenvalue.  This is the continuum version of the Ledoit--Péché eigenvector-overlap law \citep{ledoit2011eigenvectors}.

Using \cref{eq:scalar_MDE} \ref{eq:block_resolvent_functions} \ref{eq:xi_definition}, we can write \(\xi(z)\) defined as the solution of
\begin{align}
    \frac{1}{\xi(z)}
    =
    1
    +
    \gamma
    \left(
        \frac{\rho}{\xi(z)-z}
        +
        \frac{(1-\rho)\Delta}{\Delta\xi(z)-z}
    \right).
    \label{eq:xi_equation_trace_appendix}
\end{align}
The physical branch is the solution satisfying \(\xi(z)\to1\) as
\(|z|\to\infty\), and such that the corresponding Stieltjes transform
\(m(z)\) (defined in \cref{eq:global_stieltjes_xi}) obeys \(\Im m(z)>0\) for
\(z\in\mathbb C^+\).

Equivalently, \(\xi\) is the appropriate root of the cubic
\begin{align}
    \Delta \xi^3
    &-
    \left[
        z(1+\Delta)+\Delta(1-\gamma)
    \right]\xi^2
    \nonumber\\
    &+
    z
    \left[
        z+1+\Delta
        -
        \gamma\bigl(\rho+(1-\rho)\Delta\bigr)
    \right]\xi
    -
    z^2
    =
    0.
    \label{eq:xi_cubic}
\end{align}

Then, using \cref{eq:Resolvent}, the deterministic equivalent of the normalized resolvent trace (the Stieltjes transform) is
\begin{align}
    m(z)
    =
    \frac{1}{D} \Tr \mathbf \Pi_{\widehat{\Sigma}}(z) = g_s(z)\frac{\Tr \mathbf M_s}{D} \mathbf M_s
    +
    g_n(z)\frac{\Tr \mathbf M_n}{D}
    =
    \rho g_s(z)+(1-\rho)g_n(z).
    \label{eq:global_stieltjes_xi}
\end{align}
The limiting empirical spectral density is recovered from the boundary value using the Stieltjes-Perron inversion formula
\[
    \varrho(x)
    =
    \frac{1}{\pi}
    \Im m(x+i0).
\]
Similarly, we can compute the limiting signal-resolved trace constant from the deterministic equivalent of the resolvent:

\begin{align}
    \frac{1}{D}
    \Tr\left(
        \mathbf M_s\mathbf R_{\widehat{\Sigma}}(z)
    \right) \longrightarrow \frac{1}{D}
    \Tr\left(
        \mathbf M_s\mathbf \Pi_{\widehat{\Sigma}}(z)
    \right)
    =
    \rho g_s(z),
\end{align}

so we define the signal-resolved density

\begin{align}
    \varrho_s(x)
    =
    \frac{1}{\pi}
    \Im g_s(x+i0).
\end{align}
The density \(\rho\varrho_s(x)\) is the empirical spectral density weighted by
the squared projection of the corresponding empirical eigenvector onto the true
signal subspace.  Therefore the local signal overlap of an empirical eigenvector
with eigenvalue \(x\) is
\begin{align}
    w_s(x)
    =
    \frac{
        \rho\varrho_s(x)
    }{
        \varrho(x)
    }.
    \label{eq:local_signal_overlap}
\end{align}
This is the continuum version of the Ledoit--Péché eigenvector-overlap law:
for an empirical eigenvector \(\hat{\mathbf u}_a\) with eigenvalue
\(\widehat\lambda_a\approx x\),
\[
    \hat{\mathbf u}_a^\top
    \mathbf M_s
    \hat{\mathbf u}_a
    \approx
    w_s(x).
\]

The first-order trace constant is therefore
\begin{align}
    t_s
    &=
    \frac{1}{D}
    \mathbb E
    \Tr(\mathbf P\mathbf M_s)
    \nonumber\\
    &=
    \int_I
    \varrho(x)w_s(x)\,dx
    \nonumber\\
    &=
    \rho
    \int_I
    \varrho_s(x)\,dx.
    \label{eq:ts_trace_expectation}
\end{align}
Notice that \(t_s/\rho\) is the average fraction of the selected PCA subspace
lying in the true signal subspace, whereas \(t_s\) is the normalization used in
the ODE closure.

\subsection{Two-body stability operator and the deterministic equivalent of a two-resolvent chain}
\label{sec:two_body_stability}

The one-resolvent Matrix Dyson Equation determines the deterministic matrix
\(\boldsymbol\Pi(\zeta)\) that approximates a single linearised resolvent
\begin{align}
    \mathbf R_a
    &:=
    \mathbf R_{\mathcal H}(\zeta_a)
    =
    (\mathcal H-\zeta_a\mathbf I)^{-1},
    &
    \boldsymbol\Pi_a
    &:=
    \boldsymbol\Pi(\zeta_a).
\end{align}
For a deterministic matrix \(\mathbf K\), however, the deterministic equivalent
of the chain
\begin{align}
    \mathbf R_1\mathbf K\mathbf R_2
    \label{eq:two_resolvent_random_chain}
\end{align}
is not, in general, obtained by replacing the two resolvents separately:
\begin{align}
    \mathbf R_1\mathbf K\mathbf R_2
    \not\sim
    \boldsymbol\Pi_1\mathbf K\boldsymbol\Pi_2.
\end{align}
The reason is that \(\mathbf R_1\) and \(\mathbf R_2\) are functions of the
same random matrix \(\mathcal H\).  Their fluctuations are therefore correlated.
The deterministic approximation to \ref{eq:two_resolvent_random_chain} must
retain the repeated contractions generated by this shared randomness.  The
operator that resums these contractions is the two-body stability operator.

We derive this operator directly from the MDE by introducing two replicas of the
linearised problem and coupling them by a deterministic source.  This gives, in
one calculation, both the definition of the two-body stability operator and the
formula for the deterministic equivalent of a two-resolvent chain.

\subsubsection{Replicated resolvent with an off-diagonal source}

Let
\begin{align}
    \mathbb H
    &:=
    \begin{pmatrix}
        \mathcal H & \mathbf 0\\
        \mathbf 0 & \mathcal H
    \end{pmatrix},
\end{align}
and, for a deterministic matrix
\(\mathbf K\in\mathbb C^{(D+N)\times(D+N)}\), define the upper-triangular
spectral-source matrix
\begin{align}
    \mathbb Z_{\varepsilon}
    =
    \begin{pmatrix}
        \zeta_1\mathbf I_{D+N} & \varepsilon\mathbf K\\
        \mathbf 0 & \zeta_2\mathbf I_{D+N}
    \end{pmatrix}.
    \label{eq:replicated_spectral_source}
\end{align}
The corresponding replicated resolvent is
\begin{align}
    \mathbb R_{\varepsilon}
    =
    \left(
        \mathbb H-\mathbb Z_{\varepsilon}
    \right)^{-1}.
    \label{eq:replicated_resolvent_definition}
\end{align}
Since
\begin{align}
    \mathbb H-\mathbb Z_{\varepsilon}
    =
    \begin{pmatrix}
        \mathcal H-\zeta_1\mathbf I & -\varepsilon\mathbf K\\
        \mathbf 0 & \mathcal H-\zeta_2\mathbf I
    \end{pmatrix},
\end{align}
the inverse of this block-upper-triangular matrix is
\begin{align}
    \mathbb R_{\varepsilon}
    =
    \begin{pmatrix}
        \mathbf R_1
        &
        \varepsilon\mathbf R_1\mathbf K\mathbf R_2
        \\
        \mathbf 0
        &
        \mathbf R_2
    \end{pmatrix}.
    \label{eq:replicated_resolvent_block_inverse}
\end{align}
In particular, the two-resolvent chain is exactly the linear response of the
upper-right block with respect to the source:
\begin{align}
    \left.
    \frac{\partial}{\partial\varepsilon}
    \mathbb R_{\varepsilon,12}
    \right|_{\varepsilon=0}
    =
    \mathbf R_1\mathbf K\mathbf R_2.
    \label{eq:chain_as_source_derivative}
\end{align}
This source construction is useful because the same derivative can be taken on
the deterministic MDE solution.

The two diagonal blocks of \(\mathbb H\) contain the same random matrix, not two
independent copies.  To make the resulting covariance structure explicit, recall
that
\begin{align}
    \mathcal H
    =
    \sum_{\mu=1}^{N}
    \sum_{i=1}^{D}
    A_{\mu i}\mathbf B_{\mu i}.
\end{align}
Then
\begin{align}
    \mathbb H
    =
    \sum_{\mu=1}^{N}
    \sum_{i=1}^{D}
    A_{\mu i}
    \begin{pmatrix}
        \mathbf B_{\mu i} & \mathbf 0\\
        \mathbf 0 & \mathbf B_{\mu i}
    \end{pmatrix}.
\end{align}
The same Gaussian coefficient \(A_{\mu i}\) appears in both replicas.  Hence the
self-energy operator of the replicated problem acts on a \(2\times2\) block
matrix \(\mathbb T=(\mathbf T_{ab})_{a,b=1}^{2}\) as
\begin{align}
    \mathbb S[\mathbb T]
    =
    \begin{pmatrix}
        \mathcal S[\mathbf T_{11}]
        &
        \mathcal S[\mathbf T_{12}]
        \\
        \mathcal S[\mathbf T_{21}]
        &
        \mathcal S[\mathbf T_{22}]
    \end{pmatrix}.
    \label{eq:replicated_self_energy}
\end{align}
The off-diagonal action
\(\mathbb S[\mathbb T]_{12}=\mathcal S[\mathbf T_{12}]\) is precisely where the
correlation between the two resolvents enters.  Had the two diagonal blocks of
\(\mathbb H\) contained independent random matrices, this off-diagonal covariance
term would vanish, and the leading deterministic approximation would reduce to
the naive product \(\boldsymbol\Pi_1\mathbf K\boldsymbol\Pi_2\).

\subsubsection{Linearising the replicated Matrix Dyson Equation}

Let \(\mathbb \Pi_{\varepsilon}\) denote the deterministic MDE solution for the
replicated resolvent \(\mathbb R_{\varepsilon}\).  With the same sign convention
as in \ref{eq:MDE}, it satisfies
\begin{align}
    -\mathbb\Pi_{\varepsilon}^{-1}
    =
    \mathbb Z_{\varepsilon}
    +
    \mathbb S[\mathbb\Pi_{\varepsilon}].
    \label{eq:replicated_MDE}
\end{align}
At \(\varepsilon=0\), the two replicas decouple at the deterministic level and
\begin{align}
    \mathbb\Pi_{0}
    =
    \begin{pmatrix}
        \boldsymbol\Pi_1 & \mathbf 0\\
        \mathbf 0 & \boldsymbol\Pi_2
    \end{pmatrix}.
\end{align}
We therefore expand the upper-right block to first order as
\begin{align}
    \mathbb\Pi_{\varepsilon}
    =
    \begin{pmatrix}
        \boldsymbol\Pi_1
        &
        \varepsilon\mathbf M_{12}[\mathbf K]
        \\
        \mathbf 0
        &
        \boldsymbol\Pi_2
    \end{pmatrix}
    +
    O(\varepsilon^2).
    \label{eq:replicated_MDE_expansion}
\end{align}
The matrix \(\mathbf M_{12}[\mathbf K]\) will be the deterministic equivalent of
\(\mathbf R_1\mathbf K\mathbf R_2\).

For an invertible matrix-valued function \(\mathbb Q_{\varepsilon}\),
\begin{align}
    \frac{\partial}{\partial\varepsilon}
    \mathbb Q_{\varepsilon}^{-1}
    =
    -
    \mathbb Q_{\varepsilon}^{-1}
    \left(
        \frac{\partial\mathbb Q_{\varepsilon}}
        {\partial\varepsilon}
    \right)
    \mathbb Q_{\varepsilon}^{-1}.
\end{align}
Consequently, the upper-right block of
\(-\mathbb\Pi_{\varepsilon}^{-1}\), to first order in \(\varepsilon\), is
\begin{align}
    \left[
        -\mathbb\Pi_{\varepsilon}^{-1}
    \right]_{12}
    =
    \varepsilon
    \boldsymbol\Pi_1^{-1}
    \mathbf M_{12}[\mathbf K]
    \boldsymbol\Pi_2^{-1}
    +
    O(\varepsilon^2).
    \label{eq:inverse_replicated_MDE_expansion}
\end{align}
On the right-hand side of \ref{eq:replicated_MDE}, the upper-right block is
\begin{align}
    \left[
        \mathbb Z_{\varepsilon}
        +
        \mathbb S[\mathbb\Pi_{\varepsilon}]
    \right]_{12}
    =
    \varepsilon
    \left(
        \mathbf K
        +
        \mathcal S[\mathbf M_{12}[\mathbf K]]
    \right)
    +
    O(\varepsilon^2).
    \label{eq:rhs_replicated_MDE_expansion}
\end{align}
Equating \ref{eq:inverse_replicated_MDE_expansion} and
\ref{eq:rhs_replicated_MDE_expansion} gives
\begin{align}
    \boldsymbol\Pi_1^{-1}
    \mathbf M_{12}[\mathbf K]
    \boldsymbol\Pi_2^{-1}
    =
    \mathbf K
    +
    \mathcal S[\mathbf M_{12}[\mathbf K]].
    \label{eq:two_resolvent_recursive_equation_inverse_form}
\end{align}
Multiplying from the left by \(\boldsymbol\Pi_1\) and from the right by
\(\boldsymbol\Pi_2\), we obtain the recursive Dyson equation
\begin{align}
    \mathbf M_{12}[\mathbf K]
    =
    \boldsymbol\Pi_1\mathbf K\boldsymbol\Pi_2
    +
    \boldsymbol\Pi_1
    \mathcal S[\mathbf M_{12}[\mathbf K]]
    \boldsymbol\Pi_2.
    \label{eq:two_resolvent_recursive_Dyson_equation}
\end{align}
This equation is linear in the unknown two-resolvent deterministic equivalent.
It is therefore natural to define the two-body stability operator
\begin{align}
    \boxed{
    \mathcal B_{12}[\mathbf T]
    :=
    \mathbf T
    -
    \boldsymbol\Pi_1
    \mathcal S[\mathbf T]
    \boldsymbol\Pi_2
    }
    \label{eq:two_body_stability_operator}
\end{align}
for every
\(\mathbf T\in\mathbb C^{(D+N)\times(D+N)}\).  In this notation,
\ref{eq:two_resolvent_recursive_Dyson_equation} becomes
\begin{align}
    \mathcal B_{12}
    [\mathbf M_{12}[\mathbf K]]
    =
    \boldsymbol\Pi_1\mathbf K\boldsymbol\Pi_2.
    \label{eq:two_body_stability_equation}
\end{align}
Whenever \(\mathcal B_{12}\) is invertible on the relevant matrix subspace,
\begin{align}
    \boxed{
    \mathbf M_{12}[\mathbf K]
    =
    \mathcal B_{12}^{-1}
    \left[
        \boldsymbol\Pi_1\mathbf K\boldsymbol\Pi_2
    \right]
    }
    \label{eq:two_resolvent_deterministic_equivalent_general}
\end{align}
is the deterministic equivalent of the chain
\(\mathbf R_1\mathbf K\mathbf R_2\), in the same weak, anisotropic, or
normalised-trace sense in which the corresponding multi-resolvent local law is
used.  For the global contour calculations considered here, the spectral
parameters remain away from the real spectrum, and the inverse is bounded.  If
one subsequently takes boundary values close to the spectrum, the size of
\(\mathcal B_{12}^{-1}\) must be controlled as part of the local-law analysis.

The word ``stability'' has the same meaning as for the one-body MDE.  The
operator \(\mathcal B_{12}\) is the linear operator that must be inverted to
turn a perturbation of the two-replica MDE into the corresponding perturbation
of its solution.  When \(\mathcal B_{12}^{-1}\) is large, a small source can
produce a large two-resolvent response.

\subsubsection{Solving to find the two-body deterministic equivalent}
\label{subsec:two_body_deterministic_equivalent}

We now take the deterministic insertion to be the signal projector lifted to the
linearised space,
\begin{align}
    \widehat{\mathbf M}_s
    =
    \operatorname{diag}
    \left(
        \mathbf I_{\rho D},
        \mathbf 0_{(1-\rho)D},
        \mathbf 0_N
    \right).
    \label{eq:lifted_signal_projector_two_body}
\end{align}
For \(\zeta_a^2=z_a\), write
\begin{align}
    \boldsymbol\Pi_a
    =
    \operatorname{diag}
    \left(
        a_{s,a}\mathbf I_{\rho D},
        a_{n,a}\mathbf I_{(1-\rho)D},
        b_a\mathbf I_N
    \right),
    \qquad
    a=1,2.
    \label{eq:two_body_block_Pi_notation}
\end{align}
The deterministic equivalent of
\(\mathbf R_1\widehat{\mathbf M}_s\mathbf R_2\) is therefore
\begin{align}
    \boxed{
    \mathbf A_{12}^{(s)}
    :=
    \mathcal B_{12}^{-1}
    \left[
        \boldsymbol\Pi_1
        \widehat{\mathbf M}_s
        \boldsymbol\Pi_2
    \right]
    }
    \label{eq:signal_two_resolvent_DE}
\end{align}
with
\begin{align}
    \boldsymbol\Pi_1
    \widehat{\mathbf M}_s
    \boldsymbol\Pi_2
    =
    \operatorname{diag}
    \left(
        a_{s,1}a_{s,2}\mathbf I_{\rho D},
        \mathbf 0_{(1-\rho)D},
        \mathbf 0_N
    \right).
    \label{eq:signal_two_body_input}
\end{align}
Because both the MDE solution and the input are invariant under rotations within
the signal, noise, and sample blocks, the solution has the form
\begin{align}
    \mathbf A_{12}^{(s)}
    =
    \operatorname{diag}
    \left(
        x_s\mathbf I_{\rho D},
        x_n\mathbf I_{(1-\rho)D},
        x_0\mathbf I_N
    \right).
    \label{eq:signal_two_body_block_ansatz}
\end{align}
Introduce the abbreviations
\begin{align}
    \alpha_s
    &:=
    a_{s,1}a_{s,2},
    &
    \alpha_n
    &:=
    a_{n,1}a_{n,2},
    &
    \beta
    &:=
    b_1b_2.
    \label{eq:two_body_scalar_abbreviations}
\end{align}
Using the block action of the self-energy operator from
\ref{eq:self_energy_action},
\begin{align}
    \mathcal S:
    (x_s,x_n,x_0)
    \mapsto
    \left(
        x_0,
        \Delta x_0,
        \gamma
        \left[
            \rho x_s
            +(1-\rho)\Delta x_n
        \right]
    \right),
\end{align}
we find
\begin{align}
    \boldsymbol\Pi_1
    \mathcal S[\mathbf A_{12}^{(s)}]
    \boldsymbol\Pi_2
    =
    \operatorname{diag}
    \left(
        \alpha_s x_0\mathbf I_{\rho D},
        \Delta\alpha_n x_0\mathbf I_{(1-\rho)D},
        \beta\gamma
        \left[
            \rho x_s
            +(1-\rho)\Delta x_n
        \right]
        \mathbf I_N
    \right).
\end{align}
The stability equation
\begin{align}
    \mathcal B_{12}
    [\mathbf A_{12}^{(s)}]
    =
    \boldsymbol\Pi_1
    \widehat{\mathbf M}_s
    \boldsymbol\Pi_2
\end{align}
therefore reduces to the three-dimensional linear system
\begin{align}
    x_s-\alpha_s x_0
    &=
    \alpha_s,
    \\
    x_n-\Delta\alpha_n x_0
    &=
    0,
    \\
    x_0
    -
    \beta\gamma
    \left[
        \rho x_s
        +(1-\rho)\Delta x_n
    \right]
    &=
    0.
    \label{eq:signal_two_body_linear_system}
\end{align}
Equivalently,
\begin{align}
    \begin{pmatrix}
        1 & 0 & -\alpha_s\\
        0 & 1 & -\Delta\alpha_n\\
        -\gamma\beta\rho
        &
        -\gamma\beta(1-\rho)\Delta
        &
        1
    \end{pmatrix}
    \begin{pmatrix}
        x_s\\
        x_n\\
        x_0
    \end{pmatrix}
    =
    \begin{pmatrix}
        \alpha_s\\
        0\\
        0
    \end{pmatrix}.
    \label{eq:signal_two_body_matrix_system}
\end{align}
The determinant controlling the inverse on this invariant subspace is
\begin{align}
    \mathfrak d_{12}
    :=
    1
    -
    \gamma\beta
    \left[
        \rho\alpha_s
        +(1-\rho)\Delta^2\alpha_n
    \right].
    \label{eq:two_body_scalar_stability_denominator}
\end{align}
Solving \ref{eq:signal_two_body_linear_system} gives
\begin{align}
    x_0
    &=
    \frac{
        \gamma\beta\rho\alpha_s
    }{
        \mathfrak d_{12}
    },
    \\
    x_n
    &=
    \frac{
        \gamma\beta\rho\Delta\alpha_s\alpha_n
    }{
        \mathfrak d_{12}
    },
    \\
    x_s
    &=
    \alpha_s
    \left(
        1
        +
        \frac{
            \gamma\beta\rho\alpha_s
        }{
            \mathfrak d_{12}
        }
    \right)
    \\
    &=
    \alpha_s
    \frac{
        1
        -
        \gamma\beta(1-\rho)\Delta^2\alpha_n
    }{
        \mathfrak d_{12}
    }.
    \label{eq:signal_two_body_scalar_solution}
\end{align}

\subsubsection{Returning to the covariance resolvent}

Let
\begin{align}
    \mathbf G_a
    :=
    \mathbf R_{\widehat\Sigma}(z_a)
    =
    (\widehat\Sigma-z_a\mathbf I)^{-1},
    \qquad
    z_a=\zeta_a^2.
\end{align}
From \ref{eq:linearized_to_covariance_resolvent}, the upper-left block of the
linearised resolvent is
\begin{align}
    [\mathbf R_a]_{DD}
    =
    \zeta_a\mathbf G_a.
\end{align}
Since \(\widehat{\mathbf M}_s\) is supported only in the upper-left block,
\begin{align}
    \Tr
    \left(
        \widehat{\mathbf M}_s
        \mathbf R_1
        \widehat{\mathbf M}_s
        \mathbf R_2
    \right)
    =
    \zeta_1\zeta_2
    \Tr
    \left(
        \mathbf M_s
        \mathbf G_1
        \mathbf M_s
        \mathbf G_2
    \right).
    \label{eq:linearized_two_chain_to_covariance_chain}
\end{align}
Therefore the deterministic equivalent of the normalised covariance-resolvent
chain is
\begin{align}
    \frac{1}{D}
    \Tr
    \left(
        \mathbf M_s
        \mathbf G_1
        \mathbf M_s
        \mathbf G_2
    \right)
    &\sim
    \frac{1}{D\zeta_1\zeta_2}
    \Tr
    \left(
        \widehat{\mathbf M}_s
        \mathbf A_{12}^{(s)}
    \right)
    \\
    &=
    \frac{\rho x_s}{\zeta_1\zeta_2}.
    \label{eq:two_covariance_resolvent_trace_from_A12}
\end{align}
Using
\begin{align}
    a_{s,a}
    &=
    \zeta_a g_s(z_a),
    &
    a_{n,a}
    &=
    \zeta_a g_n(z_a),
    &
    b_a
    &=
    -\frac{\xi(z_a)}{\zeta_a},
\end{align}
we have
\begin{align}
    \beta\alpha_s
    &=
    \xi(z_1)\xi(z_2)
    g_s(z_1)g_s(z_2),
    \\
    \beta\alpha_n
    &=
    \xi(z_1)\xi(z_2)
    g_n(z_1)g_n(z_2).
\end{align}
It is convenient to define
\begin{align}
    \kappa(z_1,z_2)
    :=
    \gamma\xi(z_1)\xi(z_2)
    \left[
        \rho g_s(z_1)g_s(z_2)
        +(1-\rho)\Delta^2g_n(z_1)g_n(z_2)
    \right].
    \label{eq:covariance_two_body_denominator}
\end{align}
Then \ref{eq:two_covariance_resolvent_trace_from_A12} becomes
\begin{align}
    \boxed{
    \frac{1}{D}
    \Tr
    \left(
        \mathbf M_s
        \mathbf R_{\widehat\Sigma}(z_1)
        \mathbf M_s
        \mathbf R_{\widehat\Sigma}(z_2)
    \right)
    \sim
    \rho g_s(z_1)g_s(z_2)
    \frac{
        1
        -
        \gamma(1-\rho)\Delta^2
        \xi(z_1)\xi(z_2)
        g_n(z_1)g_n(z_2)
    }{
        1-\kappa(z_1,z_2)
    }
    }
    \label{eq:explicit_two_covariance_resolvent_DE}
\end{align}

\subsubsection{Two-resolvent quantity \(T_{ss}\)}

The previous calculated quantity is the two-body deterministic equivalent required in the double-contour
representation of \(T_{ss}\).  In particular,
\begin{align}
    T_{ss}
    &=
    \frac{1}{(2\pi i)^2}
    \oint_{\Gamma_I}
    \oint_{\Gamma_I}
    \frac{1}{D}
    \Tr\!\left(
        \mathbf M_s\mathbf R_{\widehat\Sigma}(z_1)
        \mathbf M_s\mathbf R_{\widehat\Sigma}(z_2)
    \right)dz_1dz_2 \\
    T_{ss}
    &=
    \frac{1}{(2\pi i)^2}
    \oint_{\Gamma_I}
    \oint_{\Gamma_I}
    \frac{\rho x_s(z_1,z_2)}{\zeta_1\zeta_2}
    \,dz_1\,dz_2,
    \label{eq:Tss_from_two_body_DE}
\end{align}
where \(x_s(z_1,z_2)\) is given by
\ref{eq:signal_two_body_scalar_solution}, and the square roots
\(\zeta_a^2=z_a\) are chosen consistently along the contours.  In a gapless regime the same formula is interpreted through the limiting upper and lower
boundary values of the resolvents, as in the one-resolvent discussion. We detail this now.

The second-order trace constant is
\begin{align}
    T_{ss}
    =
    \frac{1}{D}
    \mathbb E
    \Tr\!\left(
        \mathbf P\mathbf M_s\mathbf P\mathbf M_s
    \right).
\end{align}
Expanding in empirical eigenvectors, with
\[
    m_{ab}
    =
    \hat{\mathbf u}_a^\top
    \mathbf M_s
    \hat{\mathbf u}_b,
\]
gives
\begin{align}
    T_{ss}
    =
    \frac{1}{D}
    \sum_{a,b:\,\widehat\lambda_a,\widehat\lambda_b\in I}
    m_{ab}^2.
    \label{eq:Tss_eigenvector_expansion}
\end{align}
The diagonal terms $(a=b)$ are controlled by the local overlap \(w_s\), but the
off-diagonal terms $(a \neq b)$ measure correlations between distinct empirical eigenvectors through the fixed signal projector $\mathbf{M}_s$. Even if each $m_{ab}$ is small, there are $\mathcal{O}(D^2)$ such terms. After normalisation by $1/D$, their total contribution remains $\mathcal{O}(1)$. Therefore we must keep track of the off-diagonal terms, and we require the deterministic equivalent of the two-resolvent chain:

\begin{align}
    \mathcal G_{ss}^{(D)}(z_1,z_2)
    =
    \frac{1}{D}
    \Tr\!\left(
        \mathbf M_s\mathbf R(z_1)
        \mathbf M_s\mathbf R(z_2)
    \right),
    \qquad
    \mathbf R(z)=(\widehat\Sigma-z\mathbf I)^{-1},
\end{align}

its spectral expansion is 

\begin{align}
    \mathcal G_{ss}^{(D)}(z_1,z_2) = \frac{1}{D} \sum_{a,b=1}^D
    \frac{
        m_{ab}^2
    }{
        (\widehat\lambda_a-z_1)(\widehat\lambda_b-z_2),
    }
\end{align}

so the double discontinuity of $\mathcal{G}^{(D)}_{ss}$ extracts the limiting measure of $m_{ab}^2$ over pairs of eigenvalues. Its deterministic equivalent, \(\mathcal G_{ss}(z_1,z_2) := \lim_{D\to\infty} \mathcal G_{ss}^{(D)}(z_1,z_2)\), is obtained from the \textit{two-body stability operator} of the linearized MDE, detailed in the next section.

In the gapless bulk-selection regime the mixed boundary value has a simple diagonal
singularity because, for \(x\) in the bulk,
\begin{align}
    \kappa_{+-}(x,x)
    =
    1.
\end{align}
Indeed, subtracting the scalar equation for \(\xi_+(x)\) from its conjugate
gives
\begin{align}
    \frac{1}{
        \xi_+(x)\xi_-(x)
    }
    =
    \gamma
    \left[
        \rho |g_{s,+}(x)|^2
        +
        (1-\rho)\Delta^2 |g_{n,+}(x)|^2
    \right],
\end{align}
which is precisely the identity \(\kappa_{+-}(x,x)=1\).  Thus the singularity is
a first-order pole of the mixed boundary value through the stability
denominator, rather than a second-order pole in \(x-y\).

To recover \(T_{ss}\), take the double spectral discontinuity of
\(\mathcal G_{ss}\).  For \(x,y\in I\), define
\begin{align}
    \mathcal G_{ss}^{\sigma\tau}(x,y)
    =
    \mathcal G_{ss}(x+i\sigma0,y+i\tau0),
    \qquad
    \sigma,\tau\in\{+,-\}.
\end{align}
The two-point overlap measure is
\begin{align}
    K_2(x,y)
    =
    -\frac{1}{4\pi^2}
    \left(
        \mathcal G_{ss}^{++}
        -
        \mathcal G_{ss}^{+-}
        -
        \mathcal G_{ss}^{-+}
        +
        \mathcal G_{ss}^{--}
    \right)(x,y).
    \label{eq:K2_discontinuity}
\end{align}
Equivalently,
\[
    K_2(x,y)
    =
    \lim_{D\to\infty}
    \frac{1}{D}
    \sum_{a,b}
    m_{ab}^2
    \delta(x-\widehat\lambda_a)
    \delta(y-\widehat\lambda_b).
\]
It decomposes as
\begin{align}
    K_2(x,y)
    =
    d_s(x)\delta(x-y)
    +
    k_{ss}^{\rm reg}(x,y),
\end{align}
with diagonal weight
\begin{align}
    d_s(x)
    =
    \varrho(x)w_s(x)^2
    =
    \rho^2
    \frac{\varrho_s(x)^2}{\varrho(x)}.
    \label{eq:K2_diagonal_weight}
\end{align}
Because the mixed-boundary residue is purely imaginary, no real principal-value
term remains in the overlap measure.  For \(x\neq y\), the smooth off-diagonal
kernel is
\begin{align}
    k_{ss}^{\rm reg}(x,y)
    =
    -\frac{1}{2\pi^2}
    \left[
        \Re \mathcal G_{ss}^{++}(x,y)
        -
        \Re \mathcal G_{ss}^{+-}(x,y)
    \right].
    \label{eq:kreg_two_point}
\end{align}
The diagonal \(x=y\) of \(k_{ss}^{\rm reg}\) is removable; numerically one
should treat the analytic diagonal term \ref{eq:K2_diagonal_weight}
separately and then take the continuous limit of
\ref{eq:kreg_two_point}.

Combining the diagonal and off-diagonal contributions gives
\begin{align}
    T_{ss}
    &=
    \int_I
    d_s(x)\,dx
    +
    \iint_{I\times I}
    k_{ss}^{\rm reg}(x,y)\,dx\,dy
    \nonumber\\
    &=
    \rho^2
    \int_I
    \frac{\varrho_s(x)^2}{\varrho(x)}
    \,dx
    +
    \iint_{I\times I}
    k_{ss}^{\rm reg}(x,y)\,dx\,dy.
    \label{eq:Tss_expected_trace}
\end{align}
This formula uses the exact two-body stability denominator and separates the
diagonal singularity analytically.  It avoids any guessed pole expansion in
\(x-y\).

\subsection{Three-resolvent quantity \(\tau_{sss}\)}

The third-order trace constant is
\begin{align}
    \tau_{sss}
    =
    \frac{1}{D}
    \mathbb E
    \Tr\!\left(
        \mathbf P\mathbf M_s
        \mathbf P\mathbf M_s
        \mathbf P\mathbf M_s
    \right).
\end{align}
Expanding again in empirical eigenvectors gives
\begin{align}
    \tau_{sss}
    =
    \frac{1}{D}
    \sum_{a,b,c:\,\widehat\lambda_a,\widehat\lambda_b,\widehat\lambda_c\in I}
    m_{ab}m_{bc}m_{ca}.
    \label{eq:tau_eigenvector_expansion}
\end{align}
This expression separates into three types of contributions:
\begin{align}
    \tau_{sss}
    &=
    \underbrace{
    \frac{1}{D}
    \sum_{a\in I}
    m_{aa}^3
    }_{\text{all indices equal}}
    \nonumber\\
    &\quad
    +
    \underbrace{
    \frac{3}{D}
    \sum_{\substack{a,b\in I\\a\neq b}}
    m_{aa}m_{ab}^2
    }_{\text{one pair of indices equal}}
    \nonumber\\
    &\quad
    +
    \underbrace{
    \frac{1}{D}
    \sum_{\substack{a,b,c\in I\\a,b,c\ \mathrm{distinct}}}
    m_{ab}m_{bc}m_{ca}
    }_{\text{all indices distinct}}.
    \label{eq:tau_decomposition}
\end{align}
The factor of \(3\) in the second term accounts for the three possible
coincident pairs \((a=b)\), \((b=c)\), and \((c=a)\).

The fully diagonal term is determined by the one-point overlap density:
\begin{align}
    \frac{1}{D}
    \sum_{a\in I}
    m_{aa}^3
    \longrightarrow
    \int_I
    \varrho(x)w_s(x)^3\,dx
    =
    \rho^3
    \int_I
    \frac{\varrho_s(x)^3}{\varrho(x)^2}
    \,dx.
    \label{eq:tau_diagonal_term}
\end{align}
The one-pair-equal contribution is determined by the diagonal overlap \(w_s\)
and the off-diagonal part of the two-point kernel:
\begin{align}
    \frac{3}{D}
    \sum_{\substack{a,b\in I\\a\neq b}}
    m_{aa}m_{ab}^2
    \longrightarrow
    3
    \iint_{I\times I}
    w_s(x)
    k_{ss}^{\rm reg}(x,y)
    \,dx\,dy.
    \label{eq:tau_pair_term}
\end{align}
The remaining all-distinct term is a new three-eigenvector contribution. It
is not determined by the scalar one- and two-point trace constants alone.  At
the deterministic-equivalent level, however, it is obtained from the
length-three multi-resolvent chain generated by the same pairwise stability
operators used above.

The required covariance-side chain is
\begin{align}
    \mathcal G_{sss}^{(D)}(z_1,z_2,z_3)
    =
    \frac{1}{D}
    \Tr\!\left(
        \mathbf M_s\mathbf R(z_1)
        \mathbf M_s\mathbf R(z_2)
        \mathbf M_s\mathbf R(z_3)
    \right).
    \label{eq:three_resolvent_chain}
\end{align}
We use the linearized notation from the previous subsection.  For
\(a,b\in\{1,2,3\}\), define
\[
    \mathcal B_{ab}[\mathbf X]
    =
    \mathbf X
    -
    \boldsymbol\Pi_a
    \mathcal S[\mathbf X]
    \boldsymbol\Pi_b,
\]
and define the pair-dressed signal insertion
\begin{align}
    \mathbf V_{ab}
    =
    \mathcal B_{ab}^{-1}
    \left[
        \widehat{\mathbf M}_s
    \right].
\end{align}
By block symmetry,
\[
    \mathbf V_{ab}
    =
    \operatorname{diag}
    \left(
        v_s^{ab}\mathbf I_{\rho D},
        v_n^{ab}\mathbf I_{(1-\rho)D},
        v_0^{ab}\mathbf I_N
    \right),
\]
where
\begin{align}
    v_s^{ab}
    -
    a_{s,a}a_{s,b}v_0^{ab}
    &=
    1,
    \\
    v_n^{ab}
    -
    \Delta a_{n,a}a_{n,b}v_0^{ab}
    &=
    0,
    \\
    v_0^{ab}
    -
    \gamma b_ab_b
    \left[
        \rho v_s^{ab}
        +
        (1-\rho)\Delta v_n^{ab}
    \right]
    &=
    0.
\end{align}
Equivalently, with
\[
    \kappa_{ab}:=\kappa(z_a,z_b),
\]
one has
\begin{align}
    v_0^{ab}
    &=
    \frac{
        \gamma\rho b_ab_b
    }{
        1-\kappa_{ab}
    },
    \label{eq:v0_pair_solution}
    \\
    v_s^{ab}
    &=
    1
    +
    a_{s,a}a_{s,b}v_0^{ab},
    \\
    v_n^{ab}
    &=
    \Delta a_{n,a}a_{n,b}v_0^{ab}.
\end{align}

The length-three deterministic chain is obtained by dressing the product
\[
    \boldsymbol\Pi_1
    \mathbf V_{12}
    \boldsymbol\Pi_2
    \mathbf V_{23}
    \boldsymbol\Pi_3
\]
with the outer stability operator \(\mathcal B_{13}\).  Define
\begin{align}
    \mathbf Y_{123}
    =
    \mathcal B_{13}^{-1}
    \left[
        \boldsymbol\Pi_1
        \mathbf V_{12}
        \boldsymbol\Pi_2
        \mathbf V_{23}
        \boldsymbol\Pi_3
    \right].
\end{align}
Again,
\[
    \mathbf Y_{123}
    =
    \operatorname{diag}
    \left(
        y_s\mathbf I_{\rho D},
        y_n\mathbf I_{(1-\rho)D},
        y_0\mathbf I_N
    \right).
\]
Introduce the source terms
\begin{align}
    s_s
    &=
    a_{s,1}
    v_s^{12}
    a_{s,2}
    v_s^{23}
    a_{s,3},
    \\
    s_n
    &=
    a_{n,1}
    v_n^{12}
    a_{n,2}
    v_n^{23}
    a_{n,3},
    \\
    s_0
    &=
    b_1
    v_0^{12}
    b_2
    v_0^{23}
    b_3.
\end{align}
Then the block equations for \(\mathbf Y_{123}\) are
\begin{align}
    y_s
    -
    a_{s,1}a_{s,3}y_0
    &=
    s_s,
    \\
    y_n
    -
    \Delta a_{n,1}a_{n,3}y_0
    &=
    s_n,
    \\
    y_0
    -
    \gamma b_1b_3
    \left[
        \rho y_s
        +
        (1-\rho)\Delta y_n
    \right]
    &=
    s_0.
\end{align}
Solving gives
\begin{align}
    y_0
    =
    \frac{
        s_0
        +
        \gamma b_1b_3
        \left[
            \rho s_s
            +
            (1-\rho)\Delta s_n
        \right]
    }{
        1-\kappa_{13}
    },
    \qquad
    y_s
    =
    s_s
    +
    a_{s,1}a_{s,3}y_0.
    \label{eq:y_three_body_solution}
\end{align}
The deterministic equivalent of the three-resolvent covariance-side chain is
therefore
\begin{align}
    \mathcal G_{sss}(z_1,z_2,z_3)
    &:=
    \lim_{D\to\infty}
    \mathcal G_{sss}^{(D)}(z_1,z_2,z_3)
    \nonumber\\
    &=
    \rho
    \frac{
        y_s
    }{
        \zeta_1\zeta_2\zeta_3
    }.
    \label{eq:Gsss_three_body_deterministic_equivalent}
\end{align}
The final expression is independent of the choice of square-root branches
\(\zeta_a^2=z_a\).  All denominators appearing in
\ref{eq:Gsss_three_body_deterministic_equivalent} are pairwise stability
denominators:
\[
    1-\kappa_{12},
    \qquad
    1-\kappa_{23},
    \qquad
    1-\kappa_{13}.
\]
Thus there is no additional cubic or triple stability pole to isolate.

The compact contour representation is
\begin{align}
    \tau_{sss}
    =
    -\frac{1}{(2\pi i)^3}
    \oint_{\Gamma_I}
    \oint_{\Gamma_I}
    \oint_{\Gamma_I}
    \mathcal G_{sss}(z_1,z_2,z_3)
    \,dz_1dz_2dz_3.
    \label{eq:tau_contour_formula}
\end{align}
This formula automatically includes the diagonal, pair-coincident, and
all-distinct contributions.

For completeness, we also give the equivalent real-axis decomposition.  Away
from the pair diagonals \(x=y\), \(y=z\), and \(z=x\), define the all-distinct
three-point kernel by
\begin{align}
    k_{sss}^{\rm reg}(x,y,z)
    =
    \frac{1}{(2\pi i)^3}
    \sum_{\sigma_1,\sigma_2,\sigma_3\in\{+,-\}}
    \sigma_1\sigma_2\sigma_3\,
    \mathcal G_{sss}^{\sigma_1\sigma_2\sigma_3}(x,y,z),
    \label{eq:ksss_reg_discontinuity}
\end{align}
where
\[
    \mathcal G_{sss}^{\sigma_1\sigma_2\sigma_3}(x,y,z)
    =
    \mathcal G_{sss}
    (
        x+i\sigma_1 0,
        y+i\sigma_2 0,
        z+i\sigma_3 0
    ).
\]
The full triple spectral measure has the distributional decomposition
\begin{align}
    K_3(x,y,z)
    &=
    \varrho(x)w_s(x)^3
    \delta(x-y)\delta(y-z)
    \nonumber\\
    &\quad
    +
    w_s(x)k_{ss}^{\rm reg}(x,z)\delta(x-y)
    +
    w_s(y)k_{ss}^{\rm reg}(y,x)\delta(y-z)
    \nonumber\\
    &\quad
    +
    w_s(z)k_{ss}^{\rm reg}(z,y)\delta(z-x)
    +
    k_{sss}^{\rm reg}(x,y,z).
    \label{eq:K3_decomposition}
\end{align}
The only singular surfaces are therefore the pair diagonals, together with their
triple intersection.  These singular pieces are precisely the diagonal and
one-pair-equal contributions already displayed in
\ref{eq:tau_diagonal_term} and \ref{eq:tau_pair_term}.

Integrating \ref{eq:K3_decomposition} over \(I^3\) gives
\begin{align}
    \tau_{sss}
    &=
    \rho^3
    \int_I
    \frac{\varrho_s(x)^3}{\varrho(x)^2}
    \,dx
    \nonumber\\
    &\quad
    +
    3
    \iint_{I\times I}
    w_s(x)
    k_{ss}^{\rm reg}(x,y)
    \,dx\,dy
    \nonumber\\
    &\quad
    +
    \iiint_{I^3}
    k_{sss}^{\rm reg}(x,y,z)
    \,dx\,dy\,dz.
    \label{eq:tau_expected_trace}
\end{align}
The first line is the contribution of a single empirical eigenvector appearing
three times, the second line is the contribution of two distinct empirical
eigenvectors, and the final line is the genuine all-distinct
three-eigenvector contribution computed from the length-three deterministic
equivalent \ref{eq:Gsss_three_body_deterministic_equivalent}.

\subsection{Trace constants used in the ODE closure}

Once
\[
    t,\qquad t_s,\qquad T_{ss},\qquad \tau_{sss}
\]
have been computed, all remaining trace constants needed in
Appendix~\ref{App:TheoryProjected} follow from
\[
    \mathbf M_s+\mathbf M_n=\mathbf I,
    \qquad
    \mathbf P^2=\mathbf P.
\]
For the second-order traces,
\begin{align}
    T_{sn}
    &=
    \frac{1}{D}
    \Tr(\mathbf P\mathbf M_s\mathbf P\mathbf M_n)
    =
    t_s-T_{ss},
    \\
    T_{nn}
    &=
    \frac{1}{D}
    \Tr(\mathbf P\mathbf M_n\mathbf P\mathbf M_n)
    =
    t-2t_s+T_{ss}.
\end{align}
For the third-order traces appearing in the Hilbert--Schmidt closure,
\begin{align}
    V_{ss}
    &=
    \frac{1}{D}
    \Tr(\mathbf P\mathbf M_s\mathbf P\mathbf M_s\mathbf P\mathbf M_s)
    =
    \tau_{sss},
    \\
    V_{sn}
    &=
    \frac{1}{D}
    \Tr(\mathbf P\mathbf M_s\mathbf P\mathbf M_s\mathbf P\mathbf M_n)
    =
    T_{ss}-\tau_{sss},
    \\
    V_{nn}
    &=
    \frac{1}{D}
    \Tr(\mathbf P\mathbf M_s\mathbf P\mathbf M_n\mathbf P\mathbf M_n)
    =
    t_s-2T_{ss}+\tau_{sss}.
\end{align}
The stochastic variance terms in the \(Q_i\) dynamics are therefore
\begin{align}
    G_s
    &=
    T_{ss}+\Delta T_{sn}
    \nonumber\\
    &=
    T_{ss}
    +
    \Delta(t_s-T_{ss}),
    \\
    G_n
    &=
    T_{sn}+\Delta T_{nn}
    \nonumber\\
    &=
    (t_s-T_{ss})
    +
    \Delta(t-2t_s+T_{ss}).
\end{align}
Finally, the Hilbert--Schmidt projection coefficients are obtained by defining
\begin{align}
    \Delta_G
    =
    T_{ss}t-t_s^2
\end{align}
and substituting the deterministic trace limits into
\begin{align}
    \Omega_{ij}^{(s)}
    &=
    \frac{
        (t-t_s)V_{ij}
        -
        (t_s-T_{ss})T_{ij}
    }{
        \Delta_G
    },
    \\
    \Omega_{ij}^{(n)}
    &=
    \frac{
        -t_sV_{ij}
        +
        T_{ss}T_{ij}
    }{
        \Delta_G
    }.
\end{align}
These coefficients are fixed during fine-tuning.  They encode the typical
geometry of the random PCA projector relative to the true signal/noise
decomposition learned during pretraining.

\section{Large-alpha scaling of the representation-limited error}
\label{app:large_alpha_scaling}

We now derive the large-pretraining-sample scaling of the best possible
generalisation error. As before, let
\[
\alpha=\frac{N}{D},
\qquad
\mathbf{\widehat{\Sigma}}= \mathbf{\Sigma}+\mathbf{E},
\qquad
\mathbf \Sigma=\mathbf M_s+\Delta \mathbf M_n,
\]
where \(\mathbf M_s\) and \(\mathbf M_n\) project onto subspaces of dimensions \(\rho D\) and
\((1-\rho)D\), respectively, and \(\mathbf E \) is random matrix (capturing finite sample effects). The population signal and noise eigenvalues are
therefore \(1\) and \(\Delta\), with spectral gap
\[
g=1-\Delta.
\]
We assume throughout that \(0<\Delta<1\) is fixed as \(\alpha\to\infty\), so
that the population eigengap remains nonzero. The empirical PCA projector \(\mathbf P\)
is the rank-\(\rho D\) spectral projector of \(\widehat{\mathbf \Sigma}\) associated
with its upper spectral component, as defined in the main theory. In coordinates adapted to the signal/noise decomposition, write
\begin{align}
\widehat{\mathbf \Sigma}&=\mathbf \Sigma+ \mathbf E \\
\widehat{\mathbf \Sigma}
&=
\begin{pmatrix}
\mathbf I_{\rho D} & 0 \\
0 & \Delta \mathbf I_{(1-\rho)D}
\end{pmatrix}
+
\begin{pmatrix}
\mathbf E_{ss} & \mathbf E_{sn} \\
\mathbf E_{ns} & \mathbf E_{nn}
\end{pmatrix}.
\end{align}

Trivially, $\widehat{\mathbf \Sigma} \rightarrow \mathbf \Sigma$ as $\alpha \rightarrow \infty$. For the Gaussian pretraining model,
\[
\mathbf X
=
\begin{pmatrix}
\mathbf Z_s & \sqrt{\Delta}\, \mathbf Z_n
\end{pmatrix},
\]
where \(\mathbf Z_s\in\mathbb{R}^{N\times \rho D}\) and
\(\mathbf Z_n\in\mathbb{R}^{N\times(1-\rho)D}\) have independent standard Gaussian
entries. Therefore,
\begin{align}
\mathbf E_{ns}
=
\mathbf M_n(\widehat{\mathbf \Sigma}-\mathbf \Sigma)\mathbf M_s
=
\frac{\sqrt{\Delta}}{N}\mathbf Z_n^\top \mathbf Z_s.
\label{eq:gaussiantity_of_Ens}
\end{align}

In the subsequent sections we will find the perturbative correction to the PCA projector \(\mathbf P\) in terms of the off-diagonal block \(\mathbf E_{ns}\). We will then relate it to the principal angles between the empirical PCA subspace and the true signal subspace, and finally compute the recoverable teacher variance in terms of these angles. This allows characterization of the representation-limited error in the large-\(\alpha\) limit, in terms of $\alpha$.

\subsection{Perturbation of the PC projector}
The first-order change in a spectral projector is found from standard
Rayleigh-Schrödinger perturbation theory \citep{schrodinger1926quantisierung}. Perturbation theory tells us how the eigenvectors of $\widehat{\mathbf \Sigma}$ change as a function of the perturbation $\mathbf E$. Since the projection $\mathbf P$ onto the top-$\rho D$ subspace is formed from summing the outer products of the top-$\rho D$ eigenvectors, we can use perturbation theory to compute the change in the projector $\mathbf P$ as a function of $\mathbf E$. This is done by expanding the projector in powers of the perturbation $\mathbf E$. The first-order term gives

\[
\mathbf P
=
\mathbf M_s
+
\frac{\mathbf M_n \mathbf E \mathbf M_s+\mathbf M_s \mathbf E \mathbf M_n}{1-\Delta}
+
O\!\left(
\frac{\norm{\mathbf E}_{\text{op}}^{2}}{(1-\Delta)^2}
\right),
\]
where $\norm{\cdot}_{\text{op}}$ denotes the operator norm.
In particular, the block that rotates the empirical signal subspace into the
population noise subspace is
\begin{align}
\mathbf M_n \mathbf P \mathbf M_s
=
\frac{\mathbf E_{ns}}{1-\Delta}
+
O\!\left(
\frac{\norm{\mathbf E}_{\text{op}}^{2}}{(1-\Delta)^2}
\right).
\label{eq:rotation_block}
\end{align}
\label{eq:large_alpha_scaling}

From Corollary 2 in \cite{koltchinskii2017concentration}, \( \norm{\mathbf E}_{\text{op}}^2 
= O_{\mathbb{P}}(\alpha^{-1})
\). From the Gaussianity of \ref{eq:gaussiantity_of_Ens}, we have \( \norm{\mathbf E_{ns}}_F = O_{\mathbb{P}}(\sqrt{D\alpha^{-1}}) \). Then applying Cauchy-Schwarz to \ref{eq:rotation_block}, we have
\begin{align}
\frac{1}{D}\norm{\mathbf M_n \mathbf P \mathbf M_s}^2_F = \frac{\norm{\mathbf E_{ns}}_F^2}{D(1-\Delta)^2} + O_{\mathbb{P}}(\alpha^{-3/2})
\label{eq:rotation_block_frobenius}.
\end{align}

We will use this result in the next section.
\subsection{Scaling of the First order trace overlap, $t_s$}

Let \(\theta_1,\ldots,\theta_{\rho D}\) denote the principal angles between the
empirical PCA subspace and the true signal subspace. Then
\[
t_s
=
\frac{1}{D}\Tr(\mathbf{PM}_s) =
\frac{1}{D}\Tr(\mathbf{M}_s \mathbf{PM}_s)
=
\frac{1}{D}
\sum_{a=1}^{\rho D}\cos^2\theta_a,
\]

This can be seen by rewriting $\mathbf P$ and $\mathbf M_s$ in their respective eigenbases.

and hence
\begin{align}
\rho-t_s
=
\frac{1}{D}
\sum_{a=1}^{\rho D}\sin^2\theta_a.
\label{eq:rotation_block_sine}
\end{align}

By writing $\mathbf A = \mathbf{M}_n \mathbf{PM}_s$, we can see that $\mathbf A^\top \mathbf A = \mathbf{M}_s \mathbf{PM}_s - (\mathbf{M}_s \mathbf{PM}_s)^2$. Hence $\mathbf A$ has singular values $\cos^2\theta_a - \cos^4\theta_a = \sin^2\theta_a \cos^2\theta_a$. Consequently,
\begin{align}
\norm{\mathbf{M}_n \mathbf{PM}_s}_F^2 = \Tr(\mathbf A^\top \mathbf A)
=
\sum_{a=1}^{\rho D}
\sin^2\theta_a\cos^2\theta_a = \sum_{a=1}^{\rho D}
(\sin^2\theta_a - \sin^4\theta_a).
\end{align}

Using the Davis-Kahan theorem \citep{davis1970rotation}, which bounds the principal angles in terms of the operator norm of the perturbation (for $\Delta \neq 1$), we have \( \text{max}_{1\leq a\leq \rho D} \sin\theta_a = O\left(\frac{\norm{\mathbf E}_{\text{op}}}{1 - \Delta}\right) = O_{\mathbb{P}}(\alpha^{-1/2}) \). Allowing us to write 

\begin{align}
\frac{1}{D}\norm{\mathbf{M}_n \mathbf{PM}_s}_F^2
=
\frac{1}{D}
\sum_{a=1}^{\rho D}
\sin^2\theta_a
+
O_{\mathbb{P}}(\alpha^{-2})
\label{eq:rotation_block_frobenius_angles}
\end{align}

Combining \ref{eq:rotation_block_frobenius}, \ref{eq:rotation_block_sine} and \ref{eq:rotation_block_frobenius_angles} we get

\begin{align}
\rho-t_s
=
\frac{1}{D(1-\Delta)^2}
\norm{\mathbf E_{ns}}_F^2
+
O_{\mathbb P}(\alpha^{-3/2}).
\label{eq:ts_ens_relation}
\end{align}

Since the $O_{\mathbb P}(\alpha^{-3/2})$ term is uniform in the proportional \(D,N\to\infty\) limit, and $t_s$ and $\norm{\mathbf E_{ns}}_F^2$ are self-averaging for fixed $\alpha$ we can easily take the deterministic limit of \ref{eq:ts_ens_relation}. First we now evaluate the Gaussian average. For each signal index \(a\) and noise
index \(b\),
\[
(\mathbf E_{ns})_{ba}
=
\frac{\sqrt{\Delta}}{N}
\sum_{\mu=1}^{N}
(\mathbf Z_n)_{\mu b}(\mathbf Z_s)_{\mu a}.
\]
This variable has zero mean and variance
\[
\E (\mathbf E_{ns})_{ba}^{2}
=
\frac{\Delta}{N}.
\]
There are
\[
\rho(1-\rho)D^2
\]
entries in \(E_{ns}\), so
\[
\E\norm{\mathbf E_{ns}}_F^2
=
\rho(1-\rho)D^2\frac{\Delta}{N}.
\]
Using \(N=\alpha D\), this becomes
\[
\E\norm{\mathbf E_{ns}}_F^2
=
\rho(1-\rho)D\frac{\Delta}{\alpha}.
\]

Therefore,
\[
\rho-t_s
=
\frac{\rho(1-\rho)\Delta}{(1-\Delta)^2}
\frac{1}{\alpha}
+
O(\alpha^{-3/2}),
\]
or equivalently,
\begin{align}
\boxed{
1-\frac{t_s}{\rho}
=
\frac{(1-\rho)\Delta}{(1-\Delta)^2}
\frac{1}{\alpha}
+
O(\alpha^{-3/2})
}.
\label{eq:ts_scaling}
\end{align}

\subsection{Bayes Optimal Classifier}
\label{app:bayes_classifier}

Let $\mathbf x^\prime = \mathbf P \mathbf x$ where $\mathbf x$ denote the projected observation seen by the student. Recall our definition for the teacher aligning field, $\nu = \frac{\mathbf x^\top \mathbf w^*}{\sqrt{D}}$, whose sign is the teacher label. We can characterise the Bayes optimal classifier by considering the conditional expectation of $\nu$ given the projected observation. $(\mathbf x^\prime, \nu)$ is jointly Gaussian, hence

\begin{align}
\mathbb E[\nu|\mathbf x^\prime] = \text{Cov}(\nu, \mathbf x^\prime) \text{Cov}(\mathbf x^\prime)^{+} \mathbf x^\prime = \frac{1}{\sqrt{D}}{\mathbf w^*}^\top \mathbf P (\mathbf{P \Sigma P})^+ \mathbf x^\prime,
\end{align}

where $+$ denotes the Moore-Penrose pseudoinverse. Also, because the conditional variance of \(\nu\mid z\) is independent of \(z\),

\begin{align}
    \text{Pr}(\nu>0\mid \mathbf x^\prime) > \frac{1}{2} \iff \mathbb E[\nu\mid \mathbf x^\prime] > 0.
\end{align}

Hence the Bayes optimal classifier is given by
\begin{align}
    y_B(\mathbf x^\prime) = \text{sgn}(\mathbf w_B \mathbf x^\prime) = \text{sgn}({\mathbf w^*}^\top \mathbf P (\mathbf{P \Sigma P})^+ \mathbf x^\prime),
\end{align}

where \(\mathbf w_B = (\mathbf{P \Sigma P})^+ \mathbf P \mathbf w^*\). Let
\[
q_a=\sin^2\theta_a.
\]
The fraction of teacher-field variance recoverable from the projected observation is
\begin{align}
    \chi = \frac{(\mathbf w_B^\top \mathbf \Sigma \mathbf w^*)^2}{(\mathbf w_B^\top \mathbf \Sigma \mathbf w_B)(\mathbf {w^*}^\top \mathbf \Sigma \mathbf w^*)},
    \label{eq:chi_definition}
\end{align}

Defining $\lambda_B = \frac{\mathbf w_B^\top \mathbf P \mathbf x}{\sqrt{D}} = \frac{\mathbf w_B^\top \mathbf x}{\sqrt{D}}$ and noting that $\mathbf \Sigma \mathbf w^* = \mathbf w^*$ and $\mathbf w_B^\top \mathbf \Sigma \mathbf w_B = \mathbf w_B^\top \mathbf P \mathbf \Sigma \mathbf P \mathbf w_B = \mathbf w_B^\top \mathbf w^*$ (since $\mathbf w_B$ lies in the range of $\mathbf P$), we show that $\text{Cov}(\nu, \lambda_B) = \text{Var}(\lambda_B)$. And \ref{eq:chi_definition} becomes

\begin{align}
    \chi = \frac{(\mathbf w_B^\top \mathbf w^*)^2}{(\mathbf w_B^\top \mathbf w^*)(\mathbf {w^*}^\top \mathbf w^*)} = \frac{\mathbf w_B^\top \mathbf w^*}{\mathbf {w^*}^\top \mathbf w^*} = \frac{\text{Var}(\lambda_B)}{\text{Var}(\nu)},
\end{align}

which is precisely the fraction of teacher-field variance recoverable from the projected observation. We can use \ref{eq:chi_definition} and the noted relationships above to compute \(\chi\) in terms of the principal angles.

\begin{align}
\chi = \frac{\mathbf {w_B}^\top \mathbf \Sigma \mathbf w_B}{\mathbf {w^*}^\top \mathbf w^*} = \frac{1}{SD} \mathbf {w^*}^\top (\mathbf{P \Sigma P})^+ \mathbf w^*
&= \frac{1}{SD} \mathbf {w^*}^\top (\Delta \mathbf P + (1-\Delta)\mathbf P \mathbf M_s \mathbf P)^+ \mathbf w^* \\
&= 
\frac{1}{\rho D}
\sum_{a=1}^{\rho D}
\frac{(\mathbf p_a^\top \mathbf w^*)^2}
{\Delta+(1-\Delta)\cos^2\theta_a},
\end{align}

where $\mathbf p_a$ is the set of orthonormal eigenvectors of $\mathbf{P M_s P}$. For an isotropically distributed teacher in signal space, this becomes

\begin{align}
\chi = \frac{1}{\rho D}
\sum_{a=1}^{\rho D}
\frac{\cos^2\theta_a}
{\Delta+(1-\Delta)\cos^2\theta_a}.
\end{align}

rewriting each term in the summation as

\[
\frac{\cos^2\theta_a}
{\Delta +(1-\Delta)\cos^2\theta_a}
= 1 - \Delta \sin^2\theta_a +O(\sin^4\theta_a).
\]

we can recall the use of the Davis-Kahan \citep{davis1970rotation} theorem to bound the principal angles in terms of the operator norm of the perturbation, which gives us \(\sin^2\theta_a = O_{\mathbb{P}}(\alpha^{-1})\). Therefore, we can write

\[
1-\chi
=
\frac{\Delta}{\rho D}
\sum_{a=1}^{\rho D}\sin^2\theta_a
+
O(\alpha^{-2}).
\]
Using \ref{eq:rotation_block_sine}, we obtain
\[
1-\chi
=
\Delta\left(1-\frac{t_s}{\rho}\right)
+
O(\alpha^{-2}).
\]
Then using \ref{eq:ts_scaling}, we obtain
\begin{align}
\boxed{
1-\chi
=
\frac{(1-\rho)\Delta^2}{(1-\Delta)^2}
\frac{1}{\alpha}
+
O(\alpha^{-3/2})
}.
\label{eq:chi_expansion}
\end{align}

\subsection{Bayes optimal classification error}

The Bayes optimal classifier based only on the projected observation has error
\[
e_{B}
=
\frac{1}{\pi}\arccos\sqrt{\chi}.
\]
Equivalently,
\[
e_{B}
=
\frac{1}{\pi}\arcsin\sqrt{1-\chi}.
\]
Substituting
\[
\arcsin x=x+O(x^3)
\qquad
\text{as }x\to0,
\]
we obtain
\[
e_{B}
=
\frac{1}{\pi}\sqrt{1-\chi}
+
O((1-\chi)^{3/2}) = \frac{1}{\pi}\sqrt{1-\chi}
+
O(\alpha^{-3/2}).
\]
Substituting \ref{eq:chi_expansion} yields
\[
\boxed{
e_{B}
=
\frac{\Delta\sqrt{1-\rho}}
{\pi(1-\Delta)}
\alpha^{-1/2}
+
O(\alpha^{-1})
}.
\]

Thus the recoverable-variance deficit scales as
\[
1-\chi\sim\alpha^{-1},
\]
whereas the classification error scales as
\[
e_{B}\sim\alpha^{-1/2}.
\]

This result holds in the separated, perturbative regime with fixed
\(0<\Delta<1\). It is not uniform as \(\Delta\to1\), because the eigengap
\(1-\Delta\) closes and the Kato expansion ceases to be controlled. A
sufficient qualitative condition for the expansion is
\[
\frac{\sqrt{\Delta/\alpha}}{1-\Delta}\ll1,
\qquad\text{or equivalently}\qquad
\alpha\frac{(1-\Delta)^2}{\Delta}\gg1.
\]

\section{Large-alpha scaling of the Fixed points of the Dynamics}

In this section we derive the asymptotic SFT generalization error directly from the
projected stochastic dynamics. The central observation is that, although the microscopic
student weight does not in general converge to a deterministic vector, one can identify
a linear observable whose stationary drift determines the student--teacher field
correlation exactly.

Recalling the SFT update \ref{eq:sft_update}

\begin{align}
    \mathbf{w}^{(\mu+1)}
    =
    \mathbf{w}^{(\mu)}
    -
    \frac{\eta}{T\sqrt{D}}
    \sum_{t=1}^T
    y_t^{(\mu)}
    \mathbf{P}\mathbf{x}_t^{(\mu)}
    \theta\left(-y_t^{*(\mu)}y_t^{(\mu)}\right),
\end{align}

The student observes only the projected input \(Px\), where \(P\) is the empirical PCA
projector learned during pretraining.  Recall
\begin{align}
    S &= \frac{\|\mathbf{w}^*\|^2}{D},
    &
    \widehat Q &= \frac{1}{D}\mathbf{w}^\top \mathbf{P\Sigma P} \mathbf{w},
    &
    J &= \frac{1}{D}\mathbf{w}^\top \mathbf{P} \mathbf{w}^* .
\end{align}
The student--teacher field correlation is
\begin{align}
    c
    =
    \frac{J}{\sqrt{S\widehat Q}},
\end{align}
and the classification error is
\begin{align}
    E
    \equiv 1-P_{\rm gen}
    =
    \frac{1}{\pi}\arccos c.
    \label{eq:sft_error_from_corr}
\end{align}

Let
\begin{align}
    \beta = 1-2\epsilon,
    \qquad
    k=\sqrt{\frac{2}{\pi}},
\end{align}
where \(\epsilon\) is the probability of flipping the teacher label.  For
\(y=\text{sgn}(\lambda)\) and noisy teacher label \(y^*\), the elementary identity
\begin{align}
    y\,\mathbf 1_{\{y\neq y^*\}}
    =
    \frac{1}{2}(y-y^*)
\end{align}
together with
\begin{align}
    \mathbb E[y^*|\nu]
    =
    (1-2\epsilon)\text{sgn}(\nu)
    =
    \beta\,\text{sgn}(\nu)
\end{align}
implies that the conditional mean SFT update is proportional to
\begin{align}
    \mathbb E[\Delta \mathbf w\,|\,\mathbf w]
    \propto
    -\frac{\eta}{2}
    \left[
        \mathbb E\!\left(\mathbf P \mathbf x\,\text{sgn}\lambda\right)
        -
        \beta\,\mathbb E\!\left(\mathbf P \mathbf x\,\text{sgn}\nu\right)
    \right].
\end{align}
For centred jointly Gaussian variables one has
\begin{align}
    \mathbb E[X\text{sgn}(Y)]
    =
    \sqrt{\frac{2}{\pi}}\,
    \frac{\text{Cov}(X,Y)}{\sqrt{\text{Var}(Y)}}.
    \label{eq:gaussian_sign_identity}
\end{align}

Applying \ref{eq:gaussian_sign_identity} to the Gaussian input \(x\) gives
\begin{align}
    \mathbb E[\mathbf P \mathbf x\,\text{sgn}\lambda]
    &=
    k\,\frac{\mathbf P \mathbf \Sigma \mathbf P \mathbf w}{\sqrt{D\widehat Q}},
    \\
    \mathbb E[\mathbf P \mathbf x\,\text{sgn}\nu]
    &=
    k\,\frac{\mathbf P \mathbf \Sigma \mathbf w^*}{\sqrt{DS}}
    =
    k\,\frac{\mathbf P \mathbf w^*}{\sqrt{DS}}.
\end{align}
Thus, on the usual macroscopic time scale, the conditional drift is
\begin{align}
    \mathbf b(\mathbf w)
    =
    -\frac{\eta k}{2\sqrt{D}}
    \left[
        \frac{\mathbf{P\Sigma P w}}{\sqrt{\widehat Q}}
        -
        \beta\,\frac{\mathbf P \mathbf w^*}{\sqrt S}
    \right].
    \label{eq:exact_sft_weight_drift}
\end{align}
Importantly, one should \emph{not} set \(b(w)=0\): the stochastic SFT dynamics
contain an \(O(\eta^2)\) contribution to quadratic observables, so a stationary
macroscopic state need not correspond to a deterministic fixed weight vector.

Instead, define the projected covariance on \(\text{Ran} \mathbf{P}\),
\begin{align}
    \mathbf A
    :=
    \mathbf{P\Sigma P}\big|_{\text{Ran} \mathbf{P}}.
\end{align}
Since \(\Sigma\succeq \Delta \mathbf I\), \(\mathbf A\) is positive definite on \(\text{Ran} \mathbf{P}\).
Let
\begin{align}
    \mathbf p := \mathbf P \mathbf w^*
\end{align}
and introduce an additional order-parameter (linear in $\mathbf w$)
\begin{align}
    K
    :=
    \frac{1}{D} \mathbf w^\top \mathbf A^{-1} \mathbf p.
\end{align}
Because \(K\) is linear in \(\mathbf w\), its macroscopic drift contains no quadratic
diffusion correction.  Using \ref{eq:exact_sft_weight_drift},
\begin{align}
    \dot K
    &=
    \frac{1}{D}(\mathbf A^{-1}\mathbf p)^\top \mathbf b(\mathbf w)
    \nonumber\\
    &=
    -\frac{\eta k}{2\sqrt{D}}
    \left[
        \frac{J}{\sqrt{\widehat Q}}
        -
        \beta\,\frac{\mathbf p^\top \mathbf A^{-1}\mathbf p}{\sqrt S}
    \right].
\end{align}
At a stationary macroscopic state, \(\dot K=0\), and hence
\begin{align}
    \boxed{
    c_{\rm SFT}^*
    =
    \frac{J^*}{\sqrt{S\widehat Q^*}}
    =
    (1-2\epsilon)H_\alpha
    }
    \label{eq:exact_sft_corr}
\end{align}
with
\begin{align}
    \boxed{
    H_\alpha
    :=
    \frac{
        {w^*}^\top P(P\Sigma P)^+Pw^*
    }{
        {w^*}^\top w^*
    } = \frac{{w_B}^\top Pw^*}{{w^*}^\top w^*} = \frac{{w_B}^\top w^*}{{w^*}^\top w^*} = \chi.
    }
    \label{eq:Halpha_def}
\end{align}

Recall from the definition of $\chi$, we see that \(H_\alpha\) is precisely the fraction of teacher-field variance recoverable from the projected observation. Recalling \ref{eq:chi_expansion}, we have

\begin{align}
    e_{\rm SFT}^*(\alpha,\epsilon) = \frac{1}{\pi}\arccos c_{\rm SFT}^*
    =
    \frac{1}{\pi}
    \arccos\!\left[
        (1-2\epsilon)
        \left(
            1-A(\Delta,\rho)\frac{1}{\alpha}
            +O(\alpha^{-3/2})
        \right)
    \right],
    \label{eq:sft_final_scaling}
\end{align}

with $A(\Delta,\rho) = \frac{(1-\rho)\Delta^2}{(1-\Delta)^2}$. For clean labels, \(\epsilon=0\), using
\(\arccos(1-x)=\sqrt{2x}+O(x^{3/2})\),
\begin{align}
    \boxed{
    e_{\rm SFT}^*(\epsilon=0) =
    \frac{\sqrt{2}}
    {\pi}\sqrt{A(\Delta,\rho)}
    \alpha^{-1/2} + O(\alpha^{-1}) = \sqrt{2}e_B,
    }
\end{align}

where the leading $O(\alpha^{-1})$ term is inherited from the $O(\alpha^{-3/2})$ term in \ref{eq:chi_expansion}, not from the $O(\alpha^{-3/2})$ term in the inverse cosine Taylor expansion.
Thus the SFT exponent is \(-1/2\), identical to the Bayes error exponent. 
For any fixed
\(0<\epsilon<1/2\), and using $\text{arccos}(a-x) = \arccos(a) - \frac{x}{\sqrt{1-a^2}} + O\left(\frac{a}{(1-a^2)^{3/2}}x^2\right)$,
\begin{align}
    e_{\rm SFT}^*
    &=
    \frac{1}{\pi}\arccos(1-2\epsilon)
    +
    \frac{
        (1-2\epsilon)A(\Delta,\rho)
    }{
        \pi\sqrt{1-(1-2\epsilon)^2}
    }
    \alpha^{-1}
    +
    O\left(\epsilon^{-1/2}\alpha^{-3/2} + \epsilon^{-3/2}\alpha^{-2}\right).
    \label{eq:sft_final_scaling_noisy}
\end{align}

Where the $O(\epsilon^{-1/2}\alpha^{-3/2}$ term is inherited from the $O(\alpha^{-3/2})$ term in \ref{eq:chi_expansion}, and the $O(\epsilon^{-3/2}\alpha^{-2})$ term is inherited from the $O\left(\frac{a}{(1-a^2)^{3/2}}x^2\right)$ term in the inverse cosine Taylor expansion.

\begin{align}
    e_{\rm SFT}^* - e_\epsilon
    &=
    \frac{
        \beta A(\Delta,\rho)
    }{
        \pi\sqrt{1-\beta^2}
    }
    \alpha^{-1}
    +
    O\left(\epsilon^{-1/2}\alpha^{-3/2} + \epsilon^{-3/2}\alpha^{-2}\right),
    \label{eq:excess_sft_error}
\end{align}
where $e_\epsilon=\frac{1}{\pi}\arccos(1-2\epsilon)$. \ref{eq:excess_sft_error} is valid only for \(\epsilon\gg\alpha^{-1}\), otherwise the $O(\epsilon^{-3/2}\alpha^{-2})$ term is not subleading. We see the raw error approaches a non-zero noise floor, while the excess error above that floor decays as \(\alpha^{-1}\).  In the joint small-noise, large-\(\alpha\) regime,

\begin{align}
    E_{\rm SFT}^*
    \simeq
    \frac{1}{\pi}
    \sqrt{
        4\epsilon
        +
        \frac{2\beta A(\Delta,\rho)}{\alpha}
    },
\end{align}
which displays explicitly the crossover between the clean
\(\alpha^{-1/2}\) representation-limited regime and the noisy-label error floor.

\section{RL with \texorpdfstring{$n$}{n}-or-more correct decisions}
\label{app:n_or_more}

We now extend the all-correct reward rule to episodes in which at least $n$ of the
$T$ decisions must be correct. The thermodynamic limit, Gaussian averages, and
operator closure are the same as in \cref{App:TheoryProjected}; only the
reward-weighted expectations change. We first calculate these expectations and
then derive the consequences for initial learning and asymptotic performance.
We distinguish the isotropic, realizable model from the projected model, where
information loss and anisotropy can both affect the limiting error.

\subsection{Reward-weighted expectations and projected dynamics}
\label{app:nmore_dynamics}

For an episode of independent observations, define
\begin{align}
    K &= \sum_{t=1}^T \mathbb{1}\{y_t=y_t^*\},
    & R_n &= \mathbb{1}\{K\geq n\},
    \qquad 1\leq n\leq T.
    \label{eq:nmore_reward}
\end{align}
The student weights are held fixed within an episode, and the update is
\begin{align}
    \mathbf w^{(\mu+1)}
    =\mathbf w^{(\mu)}+
    \frac{\eta}{T\sqrt D}R_n^{(\mu)}
    \sum_{t=1}^T y_t^{(\mu)}\mathbf P\mathbf x_t^{(\mu)}.
    \label{eq:nmore_update}
\end{align}
Thus unrewarded episodes give no update. We use clean teacher labels and a
binary positive reward, without a reward baseline or a penalty. As before,
$\xi=\mu/D$ denotes training time, and $D\to\infty$ is taken at fixed $T$
and bounded covariance eigenvalues. The pretraining ratio $\alpha=N/D$
and projector $\mathbf P$ are fixed during finetuning.

Recall that the probability of one correct decision is
\begin{align}
    P=\frac12+\frac1\pi\sin^{-1}c,
    \qquad c=\frac{J_s+J_n}{\sqrt{S\widehat Q}},
    \qquad \widehat Q=Q_s+\Delta Q_n.
\end{align}
Condition on one tagged decision. If it is correct, the other $T-1$ decisions
must contain at least $n-1$ correct answers; if it is incorrect, they must
contain at least $n$. The corresponding conditional reward probabilities are
\begin{align}
    a_n(P)&=\Pr\{\operatorname{Bin}(T-1,P)\geq n-1\},
    & b_n(P)&=\Pr\{\operatorname{Bin}(T-1,P)\geq n\}.
    \label{eq:nmore_tagged_probabilities}
\end{align}
We take a binomial tail to be one when its threshold is zero and zero when
the threshold exceeds the number of trials. It is useful to define
\begin{align}
    H_n(P)&=a_n(P)+b_n(P),\\
    h_n(P)&=a_n(P)-b_n(P)
    =\binom{T-1}{n-1}P^{n-1}(1-P)^{T-n},
    \label{eq:nmore_discrimination}\\
    B_n(P)&=\langle R_n\rangle
    =\sum_{k=n}^T\binom Tk P^k(1-P)^{T-k}
    =Pa_n(P)+(1-P)b_n(P).
    \label{eq:nmore_reward_probability}
\end{align}
The factor $h_n$ is the probability that the other decisions contain exactly
$n-1$ correct answers. Only in this event does changing the tagged decision
change whether the episode is rewarded.

For any field $X$ depending on the tagged observation, averaging over the
remaining observations gives
\begin{align}
    \langle XyR_n\rangle
    &=\left\langle Xy\left[
    a_n\mathbb{1}\{y=y^*\}+b_n\mathbb{1}\{y\ne y^*\}
    \right]\right\rangle\\
    &=\frac{H_n}{2}\langle X\operatorname{sgn}(\lambda)\rangle
    +\frac{h_n}{2}\langle X\operatorname{sgn}(\nu)\rangle.
    \label{eq:nmore_tagged_average}
\end{align}
Here we used $\mathbb{1}\{y=y^*\}=(1+yy^*)/2$ and $y^2=1$.
This calculation does not assume that $X$ is independent of correctness.
The remaining expectations are precisely the Gaussian sign averages already
calculated in \cref{App:TheoryProjected}.

Taking $X=\tilde\nu_i$ and $X=\lambda_i$ in
\cref{eq:nmore_tagged_average}, and substituting into \cref{eq:rl_ode}, gives
\begin{align}
    \frac{dJ_i}{d\xi}
    &=\frac{\eta}{\sqrt{2\pi}}
    \left(H_n\frac{C_i^J}{\sqrt{\widehat Q}}
    +h_n\frac{S_i}{\sqrt S}\right),\\
    \frac{dQ_i}{d\xi}
    &=\eta\sqrt{\frac2\pi}
    \left(H_n\frac{C_i^Q}{\sqrt{\widehat Q}}
    +h_n\frac{J_i}{\sqrt S}\right)
    +\frac{\eta^2}{T}G_iB_n,
    \qquad i\in\{s,n\}.
    \label{eq:nmore_projected_odes}
\end{align}
All reward probabilities in these equations are evaluated at the current $P$.
The quantities $C_i^J,C_i^Q$ and $G_i$ retain their earlier definitions.
In particular, the quadratic contribution contains $B_n$, since $R_n^2=R_n$.
The $T$ diagonal terms in the squared update supply
$\eta^2G_iB_n/T$; cross terms between distinct observations and the effect
of conditioning on the extensive quadratic norm vanish at leading order in
$D$. No additional reward-variance factor appears for this binary update.

The reward averages are exact in the thermodynamic limit. Closing
\cref{eq:nmore_projected_odes} still requires the operator approximation
in \cref{eq:omega_projection}, with the same trace constants and coefficients
as before. The spherical constraint in \cref{app:spherical_constraint} and
the local learning-rate calculation in \cref{App:LocallyOptimalLR} apply
with these new drift and quadratic coefficients. The scaling laws below,
however, are derived for a constant learning rate.

For $n=T$, we have $a_T=P^{T-1}$, $b_T=0$, and $B_T=P^T$.
Thus $H_T=h_T=P^{T-1}$, and \cref{eq:nmore_projected_odes} reduces to
\cref{eq:projected_rl_odes}. At the other endpoint, $n=1$,
$a_1=1$, $b_1=1-(1-P)^{T-1}$, and $B_1=1-(1-P)^T$.

\subsection{Isotropic benchmark and initial learning}
\label{app:nmore_isotropic}

To isolate the effect of the reward threshold, first consider the unprojected,
isotropic model $\mathbf P=\mathbf I$, $\Sigma=\mathbf I$, with
$S=1$, as in the RL perceptron setting of \citet{RLperceptron}.
This is a separate realizable benchmark, rather than an assumption about the
empirical PCA projector. Define
\begin{align}
    J=\frac{\mathbf w^\top\mathbf w^*}{D},\qquad
    Q=\frac{\mathbf w^\top\mathbf w}{D},\qquad
    q=\sqrt Q,\qquad c=\frac Jq,\qquad
    e=1-P=\frac1\pi\arccos c.
\end{align}
Using \cref{eq:nmore_tagged_average} with the teacher and student fields yields
\begin{align}
    \dot J&=\frac{\eta}{\sqrt{2\pi}}(H_nc+h_n),\\
    \dot Q&=\eta q\sqrt{\frac2\pi}(H_n+h_nc)
    +\frac{\eta^2}{T}B_n,\\
    \dot q&=\frac{\eta}{\sqrt{2\pi}}(H_n+h_nc)
    +\frac{\eta^2}{2Tq}B_n.
    \label{eq:nmore_isotropic_norm}
\end{align}
Dots denote derivatives with respect to $\xi$. Differentiating $c=J/q$,
the terms proportional to $H_n$ cancel, leaving
\begin{align}
    \boxed{
    \dot c=\frac{\eta}{q\sqrt{2\pi}}h_n(1-c^2)
    -\frac{\eta^2}{2Tq^2}B_nc.
    }
    \label{eq:nmore_isotropic_angle}
\end{align}
The first term improves alignment with the teacher, while the second is
the angular cost of the stochastic update. A radial rescaling does not
change $c$, so \cref{eq:nmore_isotropic_angle} also applies with a spherical
constraint, with $q$ held fixed. The norm equation itself describes only
the unconstrained dynamics.

At independent random initialization, $c(0)=0$ and $P(0)=1/2$. Hence
\begin{align}
    \dot c_n(0)
    &=\frac{\eta}{q_0\sqrt{2\pi}}
    \binom{T-1}{n-1}2^{-(T-1)},\\
    \boxed{
    \frac{\dot c_n(0)}{\dot c_{n=T}(0)}=\binom{T-1}{n-1}.
    }
    \label{eq:nmore_initial_speedup}
\end{align}
This compares equal initial norms and learning rates. It is an initial-slope
multiplier, rather than an exact rescaling of learning time. In particular,
the multiplier is largest near the middle of the binomial row and returns
to one at $n=1$.

The same multiplier holds for the projected model at independent random
initialization. There, $J_i$ and the covariance-weighted teacher overlaps
$C_i^J$ vanish initially. Since $c=0$, changes in its denominator do not
contribute to its initial derivative, and \cref{eq:nmore_projected_odes} gives
\begin{align}
    \dot c_n(0)
    =\frac{\eta(S_s+S_n)}{S\sqrt{2\pi\widehat Q_0}}
    h_n(1/2).
    \label{eq:nmore_projected_initial_slope}
\end{align}
The geometry changes the common prefactor, but not the threshold multiplier
when that prefactor is nonzero. Merely choosing a specially constructed
student with $J_s+J_n=0$ is insufficient: its higher teacher-weighted overlaps
need not vanish.

\subsection{Scaling laws in the isotropic model}
\label{app:nmore_isotropic_scaling}

Write $m=T-n$ for the number of allowed mistakes and
$C_{T,n}=\binom{T-1}{n-1}$. Transforming
\cref{eq:nmore_isotropic_angle} using $c=\cos(\pi e)$ gives
\begin{align}
    \dot e
    =-\frac{\eta}{\pi q\sqrt{2\pi}}h_n\sin(\pi e)
    +\frac{\eta^2}{2\pi Tq^2}B_n\cot(\pi e).
    \label{eq:nmore_exact_error}
\end{align}
As $e\to0$ at fixed $T,n$,
$h_n=C_{T,n}(1-e)^{n-1}e^m\sim C_{T,n}e^m$ and $B_n\to1$.
Expanding the trigonometric factors therefore gives
\begin{align}
    \dot e\simeq
    -\frac{C_{T,n}}{\sqrt{2\pi}}\frac\eta q e^{m+1}
    +\frac{1}{2\pi^2T}\left(\frac\eta q\right)^2\frac1e.
    \label{eq:nmore_small_error}
\end{align}
The asymptotic balance depends on whether the student norm is allowed to grow.

\paragraph{Unconstrained weights.}
For constant $\eta>0$, on a trajectory approaching $c=1$,
$H_n+h_nc\to2$. This holds both for $n<T$, where $a_n,b_n\to1$,
and for $n=T$, where $a_T\to1$ and $b_T=0$. Thus
\cref{eq:nmore_isotropic_norm} implies
\begin{align}
    q(\xi)\sim\eta\sqrt{\frac2\pi}\,\xi,
    \qquad
    \dot e\simeq-\frac{C_{T,n}}{2\xi}e^{m+1}
    +\frac{1}{4\pi T\xi^2e}.
    \label{eq:nmore_growing_norm}
\end{align}
For $m\geq1$, neglecting the second term gives
\begin{align}
    \frac{d(e^{-m})}{d\xi}\simeq\frac{mC_{T,n}}{2\xi},
    \qquad
    e(\xi)^{-m}\simeq e(\xi_0)^{-m}
    +\frac{mC_{T,n}}2\log\frac\xi{\xi_0}.
\end{align}
Consequently,
\begin{align}
    \boxed{
    e(\xi)\sim
    \left[\frac{T-n}{2}\binom{T-1}{n-1}
    \log\frac\xi{\xi_0}\right]^{-1/(T-n)},
    \qquad n<T.
    }
    \label{eq:nmore_logarithmic_decay}
\end{align}
Here $\xi_0>0$ is a fixed reference time. The approximation is self-consistent:
the ratio of the discarded stochastic term to the alignment term is
proportional to
\[
    \frac{1}{\xi e^{m+2}}
    \sim\frac{(\log\xi)^{(m+2)/m}}{\xi}\longrightarrow0.
\]
Although almost every episode is rewarded late in learning, the reward
discriminates between correct and incorrect decisions only with probability
of order $e^m$. Together with the decreasing effective rate $\eta/q$, this
produces inverse-logarithmic learning. The constant $\eta$ cancels from the
leading asymptotic coefficient, but still affects the transient.

The all-correct boundary $n=T$ must be treated separately. Here $m=0$ and
$C_{T,T}=1$, and both terms in \cref{eq:nmore_growing_norm} contribute.
Setting $u=e^2$ gives
\begin{align}
    \dot u\simeq-\frac u\xi+\frac{1}{2\pi T\xi^2},
    \qquad
    \frac{d(\xi u)}{d\xi}\simeq\frac{1}{2\pi T\xi}.
\end{align}
Integrating yields
\begin{align}
    \boxed{
    e(\xi)\sim\sqrt{\frac{\log(\xi/\xi_0)}{2\pi T\xi}},
    \qquad n=T.
    }
    \label{eq:nmore_all_correct_decay}
\end{align}
Thus the all-correct rule has algebraic decay with a logarithmic correction.

\paragraph{Fixed norm.}
Now hold $q$ fixed and define the effective learning rate $\gamma=\eta/q$.
At a small-error equilibrium, \cref{eq:nmore_small_error} gives
\begin{align}
    \frac{C_{T,n}}{\sqrt{2\pi}}\gamma e_\infty^{m+1}
    \simeq\frac{\gamma^2}{2\pi^2T e_\infty},
\end{align}
and hence
\begin{align}
    \boxed{
    e_\infty\sim
    \left[\frac{\eta/q}
    {\sqrt2\,\pi^{3/2}T\binom{T-1}{n-1}}\right]^{1/(T-n+2)}.
    }
    \label{eq:nmore_fixed_norm_floor}
\end{align}
This is a small-rate asymptotic, $\eta/q\to0$, at fixed $T,n$.
For a finite rate, setting \cref{eq:nmore_exact_error} to zero instead gives
the implicit equilibrium equation on the positive-correlation branch,
\begin{align}
    \frac\eta q
    =\sqrt{\frac2\pi}\,T\,
    \frac{C_{T,n}(1-e_\infty)^{n-1}e_\infty^{T-n}
    \sin^2(\pi e_\infty)}
    {B_n(1-e_\infty)\cos(\pi e_\infty)}.
    \label{eq:nmore_fixed_norm_equilibrium}
\end{align}
Unlike \cref{eq:nmore_logarithmic_decay}, \cref{eq:nmore_fixed_norm_floor}
includes $n=T$, where the exponent is $1/2$; both all-correct results also
include $T=n=1$. Substituting a growing $q(\xi)$ into the fixed-norm floor
gives an instantaneous drift-balance scale, not necessarily the unconstrained
trajectory. These constant-rate results do not determine the asymptotics
under the locally optimal schedule of \cref{App:LocallyOptimalLR}.

\subsection{Information loss and anisotropy under projection}
\label{app:nmore_projection}

We return to a fixed projector $\mathbf P$. Let
$\mathbf U\in\mathbb R^{D\times r}$ have orthonormal columns spanning its
range, so $\mathbf P=\mathbf U\mathbf U^\top$, and define
\begin{align}
    \mathbf z=\mathbf U^\top\mathbf x,\qquad
    \mathbf v=\mathbf U^\top\mathbf w,\qquad
    \mathbf C=\mathbf U^\top\Sigma\mathbf U,\qquad
    \mathbf g=\mathbf U^\top\Sigma\mathbf w^*.
\end{align}
Then
\begin{align}
    \widehat Q=\frac{\mathbf v^\top\mathbf C\mathbf v}{D},\qquad
    J=J_s+J_n=\frac{\mathbf v^\top\mathbf g}{D},\qquad
    q^2=Q_s+Q_n=\frac{\mathbf v^\top\mathbf v}{D}.
    \label{eq:nmore_retained_moments}
\end{align}
In general $\widehat Q\ne q^2$, so the isotropic angular cancellation
cannot be assumed for this model.

\paragraph{The information bound.}
The conditional Gaussian representation from \cref{app:bayes_classifier}
can be written in retained coordinates as
\begin{align}
    \nu=\frac{\mathbf g^\top\mathbf C^{-1}\mathbf z}{\sqrt D}
    +\zeta,\qquad
    \zeta\perp\mathbf z,\qquad
    \operatorname{Var}(\zeta)
    =S-\frac{\mathbf g^\top\mathbf C^{-1}\mathbf g}{D}.
    \label{eq:nmore_conditional_teacher}
\end{align}
The inverse acts on supported covariance directions; it is a pseudoinverse
if the retained covariance is singular. The Bayes direction is
$\mathbf v_B\propto\mathbf C^{-1}\mathbf g$. In the notation of
\cref{eq:chi_definition},
\begin{align}
    \chi=\frac{\mathbf g^\top\mathbf C^{-1}\mathbf g}{DS},\qquad
    \kappa=\sqrt\chi,\qquad
    e_B=\frac1\pi\arccos\kappa.
    \label{eq:nmore_information_floor}
\end{align}
This lower bound applies to any classifier using the projected observation.
It is independent of $n$ and of the learning-rate schedule. A positive
conditional variance in \cref{eq:nmore_conditional_teacher} gives $e_B>0$,
which reducing $\eta/q$ cannot remove.

For the covariance and signal-supported teacher of our model,
\begin{align}
    \mathbf C=\Delta\mathbf I_r+(1-\Delta)\mathbf U^\top\mathbf M_s\mathbf U,
    \qquad \mathbf g=\mathbf U^\top\mathbf w^*.
\end{align}
If $\ell_a$ are the eigenvalues of $\mathbf U^\top\mathbf M_s\mathbf U$
and $g_a$ are the teacher coordinates in this eigenbasis, then
\begin{align}
    \chi=\frac1{DS}\sum_{a=1}^r
    \frac{g_a^2}{\Delta+(1-\Delta)\ell_a}.
\end{align}
This is the same recoverable variance used in the large-$\alpha$ analysis.
The residual variance, rather than subspace misalignment alone, determines
whether the information floor is strictly positive.

\paragraph{The direction selected by RL.}
Applying \cref{eq:nmore_tagged_average} to each retained coordinate gives
the conditional mean drift per unit training time,
\begin{align}
    \left.\frac{d\mathbf v}{d\xi}\right|_{\rm mean}
    =\frac{\eta}{\sqrt{2\pi}}
    \left(H_n\frac{\mathbf C\mathbf v}{\sqrt{\widehat Q}}
    +h_n\frac{\mathbf g}{\sqrt S}\right).
    \label{eq:nmore_retained_mean}
\end{align}
This describes the mean update; the quadratic terms in the macroscopic
dynamics remain necessary at finite effective learning rate. When
$\mathbf C\mathbf v$ is not parallel to $\mathbf v$, the $H_n$ term
also changes direction, reinforcing components of larger retained variance.
In the limit $\eta/q\to0$, a stationary direction must therefore satisfy
\begin{align}
    H_n(P_0)\frac{\mathbf C\mathbf v_0}{\sqrt{\widehat Q_0}}
    +h_n(P_0)\frac{\mathbf g}{\sqrt S}
    =\Lambda\mathbf v_0,
    \label{eq:nmore_stationary_direction}
\end{align}
where $\Lambda$ accounts for radial growth or spherical normalization.
At the Bayes direction $\mathbf v_0\propto\mathbf C^{-1}\mathbf g$,
the left-hand side is proportional to $\mathbf g$. It is generally not
parallel to $\mathbf C^{-1}\mathbf g$ unless the teacher-bearing direction
lies within a single covariance eigenspace. Thus the Bayes classifier need
not be stationary under RL, even when the effective rate vanishes.
If the dynamics converge to a stable stationary branch, its limiting error
$e_0$ obeys $e_0\geq e_B$ and may depend on $n$ through $H_n/h_n$.
This argument does not establish uniqueness or global stability of the branch.

\paragraph{Isotropic retained covariance with missing information.}
To separate information loss from anisotropic drift, consider an auxiliary
projected Gaussian model with $\mathbf C=\mathbf I_r$ and
$0<\kappa<1$. This is a solvable special case, not an assumption about the
empirical PCA covariance in our model. Write
$t=r/D=\operatorname{Tr}(\mathbf P)/D$ for the retained fraction, preserving
the notation for the trace constant, and $\gamma=\eta/q$.
Here $\widehat Q=q^2$ and $\|\mathbf g\|^2/(DS)=\kappa^2$.
Taking the teacher overlap and norm of the update gives
\begin{align}
    \dot J&=\frac{\eta\sqrt S}{\sqrt{2\pi}}(H_nc+h_n\kappa^2),\\
    \dot q&=\frac\eta{\sqrt{2\pi}}(H_n+h_nc)
    +\frac{\eta^2t}{2Tq}B_n,\\
    \boxed{
    \dot c=\frac\gamma{\sqrt{2\pi}}h_n(\kappa^2-c^2)
    -\frac{\gamma^2t}{2T}B_nc.
    }
    \label{eq:nmore_retained_isotropic_angle}
\end{align}
The quadratic term contains $t$ because
$\operatorname{Tr}(\mathbf C)/D=r/D$. The deterministic angular target
is $c=\kappa$, rather than $c=1$.

At fixed norm, let $c_*=\kappa-\delta c$ and set
$P_B=1-e_B$. Denote reward factors evaluated at $P_B$ by
$h_B=h_n(P_B)$, $H_B=H_n(P_B)$, and $B_B=B_n(P_B)$.
Expanding \cref{eq:nmore_retained_isotropic_angle} at small $\gamma$ gives
\begin{align}
    \frac\gamma{\sqrt{2\pi}}h_B\,2\kappa\delta c
    &\simeq\frac{\gamma^2t}{2T}B_B\kappa,
    &\delta c&\simeq\frac{\gamma tB_B\sqrt{2\pi}}{4Th_B}.
\end{align}
Using $de/dc=-1/(\pi\sqrt{1-c^2})$, we obtain
\begin{align}
    \boxed{
    e_\infty=e_B+
    \frac{tB_B}{2T\sqrt{2\pi}\,h_B\sqrt{1-\kappa^2}}
    \frac\eta q+o(\eta/q).
    }
    \label{eq:nmore_projected_linear_floor}
\end{align}
The excess error is linear in the effective rate at fixed
$0<\kappa<1$ and fixed $T,n$. Its coefficient can be large, since
\begin{align}
    h_B=\binom{T-1}{n-1}(1-e_B)^{n-1}e_B^{T-n}.
\end{align}
Thus the small-rate expansion is not uniform as $\kappa\to1$ or as $T$
grows. In particular, a permissive threshold and a small information floor
may require a very small effective rate before the linear regime is reached.

For unconstrained weights and constant $\eta$, a trajectory approaching
$c=\kappa$ instead has
\begin{align}
    q(\xi)\sim\eta L\xi,\qquad
    L=\frac{H_B+\kappa h_B}{\sqrt{2\pi}},\qquad
    \gamma\sim\frac1{L\xi}.
\end{align}
Writing $\delta c=\kappa-c$ and linearizing yields
\begin{align}
    \frac{d\delta c}{d\xi}
    &\simeq-\frac{\beta_n}{\xi}\delta c
    +\frac{tB_B\kappa}{2TL^2\xi^2},
    &\beta_n&=\frac{2\kappa h_B}{H_B+\kappa h_B}.
    \label{eq:nmore_projected_relaxation}
\end{align}
Since $H_B\geq h_B>0$, we have
$0<\beta_n\leq2\kappa/(1+\kappa)<1$.
Multiplication by $\xi^{\beta_n}$ and integration show that the homogeneous
term decays as $\xi^{-\beta_n}$, while the quadratic-update forcing in
this linearized equation contributes a term proportional to $\xi^{-1}$.
For a nonzero leading amplitude,
\begin{align}
    \boxed{
    e(\xi)-e_B\sim A\xi^{-\beta_n},
    }
    \label{eq:nmore_projected_algebraic_decay}
\end{align}
where $A$ depends on the trajectory. Nonlinear corrections can also
contribute subleading powers. For $n=T$, $H_B=h_B$ and
$\beta_T=2\kappa/(1+\kappa)$. For $n<T$, a small $h_B$ can make
the relaxation extremely slow.

At a strictly positive information floor the angular fixed point can be
linearized because $h_B>0$. As $e_B\to0$, $h_B\to0$ for $n<T$ and
the exponent $\beta_n$ tends to zero. The nonlinear terms leading to
\cref{eq:nmore_logarithmic_decay} then become essential. The long-time and
vanishing-information-loss limits are therefore nonuniform: the realizable
law may describe a crossover when $e\gg e_B$, but it is not the final
relaxation law at fixed positive $e_B$ in this special case.

\subsection{Fixed points of the projected four-order-parameter dynamics}
\label{app:nmore_projected_fixed_points}

For the general covariance in our model, we use the same operator closure
as in \cref{eq:omega_projection}. Introduce normalized order parameters
\begin{align}
    q^2=Q_s+Q_n,\qquad
    j_i=J_i/q,\qquad u_i=Q_i/q^2,\qquad
    u_s+u_n=1,\qquad \widehat u=u_s+\Delta u_n.
\end{align}
There are three independent coordinates,
$\mathbf x=(j_s,j_n,u_s)^\top$, with $u_n=1-u_s$, and
\begin{align}
    e(\mathbf x)=\frac1\pi\arccos
    \left(\frac{j_s+j_n}{\sqrt{S\widehat u}}\right),
    \qquad P(\mathbf x)=1-e(\mathbf x).
    \label{eq:nmore_normalized_error}
\end{align}
To construct their dynamics explicitly, define the normalized closed moments
\begin{align}
    \mathcal C_i^J(\mathbf x)
    &=\sum_{k\in\{s,n\}}
    (\Omega_{is}^{(k)}+\Delta\Omega_{in}^{(k)})j_k,\\
    \mathcal C_i^Q(\mathbf x)
    &=\sum_{k\in\{s,n\}}
    (\Omega_{is}^{(k)}+\Delta\Omega_{in}^{(k)})u_k.
\end{align}
Within the closure, $C_i^J=q\mathcal C_i^J$ and
$C_i^Q=q^2\mathcal C_i^Q$. Consequently,
\cref{eq:nmore_projected_odes} takes the form
\begin{align}
    \dot J_i=\eta\mathcal A_i(\mathbf x),\qquad
    \dot Q_i=\eta q\mathcal D_i(\mathbf x)+\eta^2\mathcal B_i(\mathbf x),
\end{align}
where
\begin{align}
    \mathcal A_i
    &=\frac1{\sqrt{2\pi}}
    \left(H_n\frac{\mathcal C_i^J}{\sqrt{\widehat u}}
    +h_n\frac{S_i}{\sqrt S}\right),\\
    \mathcal D_i
    &=\sqrt{\frac2\pi}
    \left(H_n\frac{\mathcal C_i^Q}{\sqrt{\widehat u}}
    +h_n\frac{j_i}{\sqrt S}\right),
    &\mathcal B_i&=\frac{G_i}{T}B_n.
    \label{eq:nmore_normalized_coefficients}
\end{align}
Let $\mathcal D=\mathcal D_s+\mathcal D_n$ and
$\mathcal B=\mathcal B_s+\mathcal B_n$. The unconstrained radial equation is
\begin{align}
    \dot q=\frac\eta2\mathcal D+\frac{\eta^2}{2q}\mathcal B.
    \label{eq:nmore_general_radial}
\end{align}
Differentiating $j_i=J_i/q$ and $u_i=Q_i/q^2$ now gives
\begin{align}
    \dot j_i&=\gamma\left(\mathcal A_i-\frac{j_i}{2}\mathcal D\right)
    -\gamma^2\frac{j_i}{2}\mathcal B,\\
    \dot u_i&=\gamma(\mathcal D_i-u_i\mathcal D)
    +\gamma^2(\mathcal B_i-u_i\mathcal B),
    \qquad \gamma=\eta/q.
\end{align}
Thus the normalized dynamics can be written as
\begin{align}
    \dot{\mathbf x}
    &=\gamma\mathbf F_{T,n}(\mathbf x)
    +\gamma^2\mathbf G_{T,n}(\mathbf x),
    \label{eq:nmore_normalized_dynamics}\\
    \mathbf F_{T,n}
    &=\begin{pmatrix}
    \mathcal A_s-j_s\mathcal D/2\\
    \mathcal A_n-j_n\mathcal D/2\\
    \mathcal D_s-u_s\mathcal D
    \end{pmatrix},
    &\mathbf G_{T,n}
    &=\begin{pmatrix}
    -j_s\mathcal B/2\\
    -j_n\mathcal B/2\\
    \mathcal B_s-u_s\mathcal B
    \end{pmatrix}.
\end{align}
The same normalized equations apply with the spherical constraint, since
the radial corrections in \cref{app:spherical_constraint} leave
$j_i,u_i$ unchanged. In that case $q$, and hence $\gamma$ for constant
$\eta$, is fixed.

Consider a fixed-norm stationary branch at small $\gamma$. Its limiting
state solves $\mathbf F_{T,n}(\mathbf x_0)=0$ on the physically admissible
domain; we do not impose zero error. Suppose this state is locally
asymptotically stable, its Jacobian
$\mathbf M=\mathrm D\mathbf F_{T,n}(\mathbf x_0)$ is invertible in the
three independent coordinates, and the coefficient functions are smooth
nearby. Substituting
$\mathbf x_*(\gamma)=\mathbf x_0+\gamma\mathbf x_1+O(\gamma^2)$
into $\mathbf F_{T,n}(\mathbf x_*)+\gamma\mathbf G_{T,n}(\mathbf x_*)=0$
gives
\begin{align}
    \mathbf M\mathbf x_1+\mathbf G_{T,n}(\mathbf x_0)=0,
    \qquad
    \mathbf x_*(\gamma)=\mathbf x_0
    -\gamma\mathbf M^{-1}\mathbf G_{T,n}(\mathbf x_0)
    +O(\gamma^2).
\end{align}
Provided the error is also smooth at $\mathbf x_0$,
\begin{align}
    \boxed{
    e_\infty(\gamma)=e_0
    -\gamma\nabla e(\mathbf x_0)^\top
    \mathbf M^{-1}\mathbf G_{T,n}(\mathbf x_0)
    +O(\gamma^2),\qquad e_0=e(\mathbf x_0).
    }
    \label{eq:nmore_general_floor}
\end{align}
Thus a regular projected fixed point has an $O(\eta/q)$ correction, whose
coefficient can vanish, about its limiting error. The conclusion is local
to the selected stable branch; degenerate fixed points, branch changes,
and the singular realizable boundary require separate analysis.

At fixed norm, local relaxation is determined by the eigenvalues of
$\gamma\mathrm D\mathbf F_{T,n}(\mathbf x_*)+
\gamma^2\mathrm D\mathbf G_{T,n}(\mathbf x_*)$.
For unconstrained weights, suppose a trajectory approaches $\mathbf x_0$
with $\mathcal D(\mathbf x_0)>0$. Then
\cref{eq:nmore_general_radial} gives
$q\sim\eta L_0\xi$, where $L_0=\mathcal D(\mathbf x_0)/2$.
To leading order, the perturbation satisfies
\begin{align}
    \frac{d\delta\mathbf x}{d\xi}
    \simeq\frac{\mathbf M}{L_0\xi}\delta\mathbf x
    +\frac{\mathbf G_{T,n}(\mathbf x_0)}{L_0^2\xi^2}.
\end{align}
A homogeneous eigenmode with Jacobian eigenvalue $\lambda_a$ therefore
scales as $\xi^{\lambda_a/L_0}$. Quadratic-update forcing, resonances,
nonlinear corrections, and the projection of these modes onto the error
can affect the leading observable relaxation. There is no universal
late-time exponent depending only on $T-n$ for the anisotropic model.

The fixed points and relaxation rates in this subsection are predictions
of the four-order-parameter closure. In contrast, the information bound
$e\geq e_B$ is exact and independent of that approximation. A closed-ODE
prediction below this bound would indicate a failure of the closure to
represent the physical dynamics.

\section{Compute cost calculation details}

\subsection{Pretraining cost}
\label{app:pretraining_cost}
For an LAE with encoder dimension \(K\), data dimension \(D\) and batch size \(B\), usingf the convention that one multiply-add costes 2 FLOPs, the cost of a single forward pass is \(2KBD + 2KBD = 4KBD\). For the backward pass, the cost is decoder weight gradient + gradient into hidden layer + encoder weight gradient, which is \(2KBD + 2KBD + 2KBD = 6KBD\). Therefore the total cost per step is \(C_{\rm LAE} = 10KBD\). The total pretraining cost is then 

\[
    C_{\rm PT} = 10 N_{\rm steps} KBD = 10 N_{\rm steps} \alpha \rho D^3 = \mathcal{O}(N_{\rm steps} \alpha \rho D^3),
\]

where \(N_{\rm steps}\) is the number of training steps taken to convergence.

\subsection{SFT and RL step costs}

Each step requires projecting observed data into the encoder subspace, which is a dense matrix-vector multiplication costing \(2KD\) FLOPs. The SFT/RL steps then requires a forward pass through the student model, this costs \(2K\) FLOPs + \(K\) for the update. This must be calculated for each observation in an episode of length \(T\). Therefore the total cost per SFT/RL step is 
\[
    C_{\rm SFT/RL} = 2\rho T D^2 + 3\rho T D = \mathcal{O}(\rho T D^2).
\]

In simulation of the stochastic updates to obtain the deterministic ODEs limit, $D$ subupdates per epoch are made, so we can write

\[
C_{\rm SFT/RL} = \mathcal{O}(\rho T D^3).
\]

\section{Experimental Details of LAE training}
\label{supp:LAE training}

For each batchsize $N$ 20 LAEs were trained to convergence, the overlap of the projection of the final encoders with the signal subspace was then calculated and averaged over runs. The simple LAE had an encoder $\mathbf{E} \in \mathbb{R}^{(D \times K)}$ and a decoder $\mathbf{D} \in \mathbb{R}^{(K \times D)}$, with $D=2000$ and $K=400$.
Weights were initialised from zero mean Gaussians  with variance $1/D$ and $1/K$ for the encoder and decoder respectively. Adam optimizer with learning rate 0.0001, and 10000 steps were used ensuring convergence. JAX with float32 precision was used.

\section{Optimal Compute Allocation Experimental details}
\label{App:BudgetAllocation}

In addition to solving the projected ODEs for fixed hyperparameters, we also
optimize how a fixed computational budget is allocated between pretraining data
selection, SFT updates, and RL updates.  The optimization is performed over the
deterministic ODE dynamics rather than over stochastic simulations.  This makes
it possible to evaluate many candidate allocations efficiently.

Let \(B_{\rm tot}\) denote the total available budget.  We allocate this budget
between three stages:
\[
    C_{PT},
    \qquad
    C_{\rm SFT},
    \qquad
    C_{\rm RL},
\]
where \(C_{PT}\) is the budget spent on choosing the pretraining level
\(\alpha\), \(C_{\rm SFT}\) is the budget spent on SFT updates, and
\(C_{\rm RL}\) is the budget spent on RL updates.  The allocation satisfies
\[
    C_{PT} + C_{\rm SFT}+C_{\rm RL}=B_{\rm tot}.
\]
The cost of choosing a pretraining level \(\alpha\) is taken to be linear,
\[
    C_{PT} = c_\alpha \alpha,
\]
where \(c_\alpha\) is the cost per unit \(\alpha\).  The remaining budget is
\[
    B_{\rm ODE}(\alpha)
    =
    B_{\rm tot}-c_\alpha\alpha.
\]
For the combined SFT--RL setting, we parameterize the split of this remaining
budget by a fraction \(f_{\rm SFT}\in[0,1]\):
\[
    C_{\rm SFT}=f_{\rm SFT}B_{\rm ODE}(\alpha),
    \qquad
    C_{\rm RL}=(1-f_{\rm SFT})B_{\rm ODE}(\alpha).
\]
If the cost per SFT step is \(a\), and the cost per RL step is \(b\), then the
corresponding numbers of ODE steps are
\[
    n_{\rm SFT}
    =
    \frac{C_{\rm SFT}}{a},
    \qquad
    n_{\rm RL}
    =
    \frac{C_{\rm RL}}{b}.
\]
The objective is to maximize the final clean teacher--student agreement
probability after applying \(n_{\rm SFT}\) SFT steps followed by \(n_{\rm RL}\)
RL steps:
\[
    \max_{\alpha,f_{\rm SFT}}
    \;
    P
    \left(
        R_{\rm RL}^{n_{\rm RL}}
        \circ
        R_{\rm SFT}^{n_{\rm SFT}}
        (\mathbf q_0(\alpha))
    \right),
    \label{eq:budget_objective}
\]
where
\[
    \mathbf q=(J_s,J_n,Q_s,Q_n)
\]
denotes the ODE state, \(\mathbf q_0(\alpha)\) is the initial condition induced
by the PCA projector learned at pretraining level \(\alpha\), and
\(p_{\rm gen}\) is the probability of agreement with the clean teacher.

\subsection{Precomputation and interpolation over $\alpha$}

For each value of \(\alpha\), the initial condition and closure coefficients are
computed from the random-matrix trace constants described in
Appendix~\ref{App:ExpectedTraceConstants}.  Recomputing these constants at every
candidate value of \(\alpha\) would be expensive.  We therefore first evaluate
them on a fixed interpolation grid
\[
    \alpha_1,\ldots,\alpha_M,
\]
with \(M=\texttt{interp\_alpha\_res}\).  In the experiments reported here,
\(\texttt{interp\_alpha\_res}=200\) unless otherwise stated.

For an arbitrary candidate \(\alpha\), the initial condition
\[
    (J_s,J_n,Q_s,Q_n)
\]
and the associated constants
\[
    S,\quad S_s,\quad S_n,\quad
    \Omega_{ij}^{(s)},\quad \Omega_{ij}^{(n)},\quad
    T_{ss},\quad T_{sn},\quad T_{nn},\quad R
\]
are obtained by linear interpolation between the precomputed values.  Here
\[
    R=Q_s+Q_n
\]
is the projected student norm used when the spherical constraint is imposed.

This interpolation step allows the allocation search to vary \(\alpha\)
continuously over the search interval while avoiding repeated evaluations of the
random-matrix formulas.

\subsection{ODE evaluation for a fixed allocation}

For a fixed pair \((\alpha,f_{\rm SFT})\), the algorithm first computes the
available ODE budget
\[
    B_{\rm ODE}(\alpha)
    =
    B_{\rm tot}-c_\alpha\alpha.
\]
It then converts the SFT and RL budget shares into fractional step counts:
\[
    n_{\rm SFT}
    =
    \frac{f_{\rm SFT}B_{\rm ODE}(\alpha)}{a},
    \qquad
    n_{\rm RL}
    =
    \frac{(1-f_{\rm SFT})B_{\rm ODE}(\alpha)}{b}.
\]

The SFT trajectory is computed once per value of \(\alpha\).  Starting from
\(\mathbf q_0(\alpha)\), the SFT ODE update is iterated up to a maximum static
number of steps,
\[
    n_{\max}
    =
    \min
    \left\{
        \left\lfloor
        \frac{
            B_{\rm tot}-c_\alpha\alpha_{\min}
        }{
            \min(a,b)
        }
        \right\rfloor
        +10,
        \;
        n_{\rm cap}
    \right\},
\]
where \(n_{\rm cap}=\texttt{max\_ode\_steps}\).  The additional \(+10\) buffer
prevents interpolation errors near the largest affordable step count.  The code
stores the full SFT prefix
\[
    \mathbf q_0,
    \mathbf q_1^{\rm SFT},
    \ldots,
    \mathbf q_{n_{\max}}^{\rm SFT}.
\]
For a fractional value of \(n_{\rm SFT}\), the state after SFT is obtained by
linear interpolation between the two neighboring integer SFT iterates.

Starting from this interpolated SFT state, the RL ODE update is applied for
\(\lfloor n_{\rm RL}\rfloor\) integer steps.  If \(n_{\rm RL}\) is fractional,
one additional RL step is computed and the final state is linearly interpolated
between the integer-step and next-step states.  The final performance is then
computed using the analytic expression for the clean generalization probability,
\[
    P
    =
    \frac{1}{2}
    +
    \frac{1}{\pi}
    \sin^{-1}
    \left(
        \frac{J_s+J_n}
        {
        \sqrt{S(Q_s+\Delta Q_n)}
        }
    \right).
\]

\subsection{Coarse-to-fine grid search}

The budget-allocation problem is solved by a coarse-to-fine grid search.  For a
given total budget, the largest affordable pretraining level is
\[
    \alpha_{\max}^{\rm aff}
    =
    \min
    \left\{
        \alpha_{\max},
        \frac{B_{\rm tot}}{c_\alpha}
    \right\}.
\]
If \(\alpha_{\max}^{\rm aff}<\alpha_{\min}\), the budget is declared infeasible.

The coarse search evaluates a grid of candidate \(\alpha\) values and SFT
fractions:
\[
    \alpha \in
    \{\alpha_1^{\rm coarse},\ldots,\alpha_{N_\alpha}^{\rm coarse}\},
    \qquad
    f_{\rm SFT}\in
    \{f_1^{\rm coarse},\ldots,f_{N_f}^{\rm coarse}\}.
\]
The \(\alpha\)-grid may be linearly spaced or warped according to
\[
    \alpha(u)
    =
    \alpha_{\min}
    +
    (\alpha_{\max}^{\rm aff}-\alpha_{\min})u^p,
    \qquad
    u\in[0,1],
\]
where \(p=\texttt{alpha\_grid\_power}\).  This allows more resolution near small
\(\alpha\) when desired.

After evaluating the full coarse grid, the algorithm identifies the best
\(\texttt{top\_k}\) coarse \(\alpha\)-regions according to
\[
    \max_{f_{\rm SFT}}
    p_{\rm gen}(\alpha,f_{\rm SFT}).
\]
Each selected region is then refined.  The refined \(\alpha\)-interval is formed
by taking a small neighborhood around the selected coarse \(\alpha\)-cell, with
width controlled by \(\texttt{refine\_alpha\_half\_width\_cells}\).  Within each
selected region, the algorithm evaluates a finer grid of
\(\texttt{refine\_alpha\_res}\) values of \(\alpha\).  The SFT fraction is
refined over the full interval \(f_{\rm SFT}\in[0,1]\), rather than only near the
coarse optimum.  This is useful because the optimal SFT--RL split can shift
substantially as \(\alpha\) changes, especially near the boundary where the
remaining ODE budget changes rapidly.

The best refined candidate over all selected regions is returned.  The reported
allocation is
\[
    (C_{PT},C_{\rm SFT},C_{\rm RL})
    =
    \left(
        c_\alpha\alpha,
        f_{\rm SFT}B_{\rm ODE}(\alpha),
        (1-f_{\rm SFT})B_{\rm ODE}(\alpha)
    \right),
\]
together with the corresponding final generalization performance.

\subsection{Single-stage baselines}

We also evaluate two single-stage baselines.  In the SFT-only baseline,
\[
    f_{\rm SFT}=1,
    \qquad
    C_{\rm RL}=0.
\]
The search is then one-dimensional over \(\alpha\), and the full remaining ODE
budget is spent on SFT:
\[
    n_{\rm SFT}
    =
    \frac{B_{\rm tot}-c_\alpha\alpha}{a}.
\]
In the RL-only baseline,
\[
    f_{\rm SFT}=0,
    \qquad
    C_{\rm SFT}=0,
\]
and the remaining budget is spent entirely on RL:
\[
    n_{\rm RL}
    =
    \frac{B_{\rm tot}-c_\alpha\alpha}{b}.
\]
Both baselines use the same coarse-to-fine search over \(\alpha\) as the full
SFT--RL allocation problem.

\subsection{Implementation details}

The allocation search is implemented in JAX and run on CPU with 64-bit floating
point arithmetic enabled.  The ODE update maps for SFT and RL are JIT-compiled.
The coarse and refined grids are evaluated using vectorized JAX loops.  For each
candidate \(\alpha\), the SFT prefix trajectory is computed once and reused for
all candidate values of \(f_{\rm SFT}\).  This reuse is the main computational
saving in the two-stage search, because each candidate fraction only requires
interpolating along the stored SFT trajectory before running the RL segment.

Invalid numerical evaluations, such as candidates producing NaN values, are
assigned performance \(-\infty\) and are therefore excluded from the optimum.
For each run, the code saves a JSON file containing the training mode, total
budget, task parameters \(T\) and \(\epsilon\), runtime, maximum ODE step count,
best allocation, best performance, selected \(\alpha\), selected
\(f_{\rm SFT}\), refined search regions, and the full search configuration.  The
saved search configuration includes the coarse and refined grid resolutions,
the number of top coarse regions refined, the \(\alpha\)-grid warping parameter,
and the interpolation resolution used for the precomputed random-matrix
constants.

\section{Skill Composition}
\label{app:skill_composition}

We adopt linear composition of multiple RL perceptrons following \citet{lee2024animals}. In this section, we make a short summary of the framework to make the paper self-contained. Total $K$ of RL perceptrons, which is an abstraction of $K$ independent skills are linearly combined with compositional context vector $\mathbf{V}\in\mathbb{R}^K$. Each element $V_{k}$ indicates the contribution of $k_{\text{th}}$ skill. We We restrict our model with $\lVert{\mathbf{V}}\rVert = 1$. We notate $\mathbf{V}^*$ as teacher and $\mathbf{V}$ as a student, and $\mathbf{V}$ is updated as the agent gets learning signal. The skill composition task is defined as following:
\begin{align}
\tilde{y}_t^* = \text{sign}\left(\sum^K_{k=1} \mathbf{V}^*_k \frac{{\mathbf{w}^*_k}^\intercal \mathbf{x}_{k,t}}{\sqrt{N}}\right),
\end{align}
where we use `` $\Tilde{~}$ '' to notate skill composition task. The updates of the student $\mathbf{w}$ and $\mathbf{V}$ becomes

\begin{align}
    \Delta \mathbf{w}_k^{n+1} = \frac{\eta_{ \mathbf{w}}}{\sqrt{N}}\Biggl(\frac{1}{T}\sum^{T}_{t=1}\Tilde{y}_t V_k\mathbf{x}_{k,t}\Biggr) \mathbb{I}(R(\Tilde{y}_{1:T}, \Tilde{y}^*_{1:T})) \\
    \Delta V_k^{n+1} = \frac{\eta_{ v}}N\Biggl(\frac{1}{T}\sum^{T}_{t=1}\Tilde{y}_t \frac{{\mathbf{w}_k}^\intercal x_{k,t}}{\sqrt{N}}\Biggr)\mathbb{I}(R(\Tilde{y}_{1:T}, \Tilde{y}^*_{1:T})).
\end{align}

In thermodynamic limit, the ODEs of the order parameters are as following,

\begin{align}
    \label{eq: Q_ode_skill}
    \frac{dQ_{k}}{d\xi} & = \frac{2\eta_w}{\sqrt{2\pi}}\Biggl(\frac{V_k^2 Q_{k}}{\sqrt{\sum_{i=1}^K{V_i^2Q_{i}}}} + \frac{V_kV^*_k R_{k}}{\sqrt{\sum_{i=1}^K{V_i^{*2}{S_{i}}}}} \Biggr )\Tilde{P}^{T-1} \nonumber + \frac{\eta_w^2}{T}V_k^2 \Tilde{P}^T \\
    \frac{dR_{k}}{d\xi} & = \frac{\eta_w}{\sqrt{2\pi}}\Biggl (\frac{V_k^2R_{k}}{\sqrt{\sum_{i=1}^K{V_i^2Q_{i}}}} + \frac{V_kV^*_k S_{k}}{\sqrt{\sum_{i=1}^K{V_i^{*2}S_{i}}}} \Biggr )\Tilde{P}^{T-1} \\
    \frac{dV_k}{d\xi} & = \frac{\eta_v}{\sqrt{2\pi}}\Biggl( \frac{V_k Q_{k}}{\sqrt{\sum^K_{i=1} V_i^2 Q_{i} }} + \frac{V_k^* R_{k}}{\sqrt{\sum_{i=1}^K V^{*2}_i S_{i}}} \Biggr )\Tilde{P}^{T-1},
\end{align}

where $\eta_{w}$ and $\eta_{v}$ denote learning rates for $\mathbf{w}$ and $\mathbf{V}$, respectively.

\begin{figure}[t]
    \centering
\includegraphics[width=\linewidth]{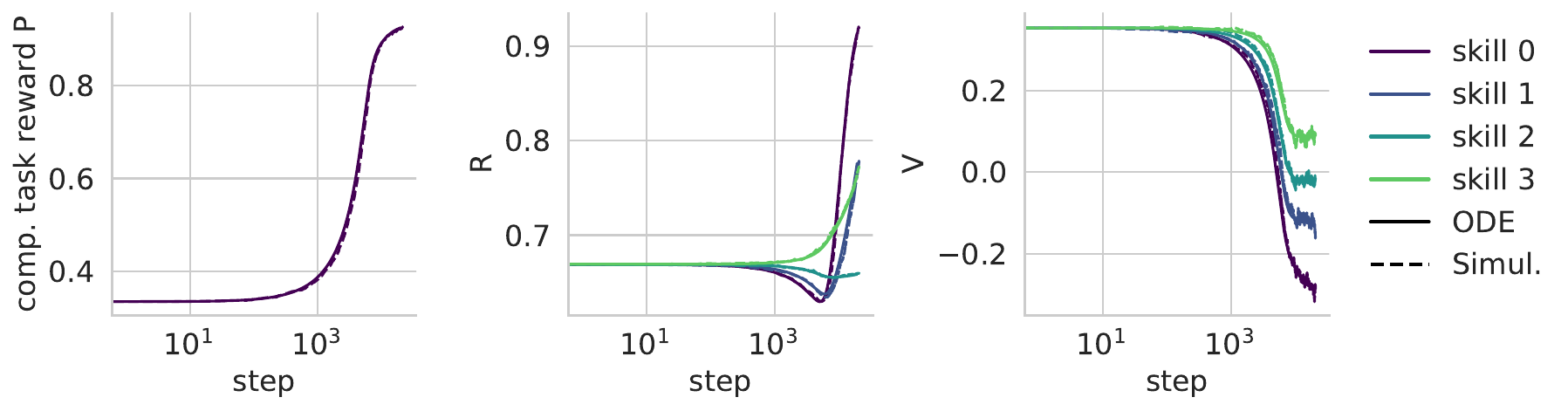}
        \caption{Comparison of ODEs and simulation from skill compositions. We show dynamics of compositional task reward probability (left), each skill level ($R_{0-3}$, middle) and composition context ($V_{0-3}$, right). The integration of ODEs and simulation show good match. ($T=8$, $K=8$ (we visualize first 4 skills out of 8 skills in total), $\eta_{w}=\eta{v}=1.0$.)}
        \label{supp_fig:ode_sim_comp}
\end{figure}

We vary the initial $R_{k}$ for different pre-trained skill level between 0 and maximum value that could be reach given $T$.

In~\cref{supp_fig:ode_sim_comp}, we report the number of steps that reached the 99\% of the minimum generalization error by numerical integration of ODEs. We used $K=8, T=8$ for the experiments and varied the number of junk skills among 8 skills. We used the same learning rate $1.0$ for $\eta_{w}$ and $\eta_{v}$.

\section{Details on LLM Experiments}
\label{app:olmo_exp}

We fine-tuned OLMo-2 1B base model~\citep{olmo20242} on GSM8K~\citep{cobbe2021training} math word problems across 3 random seeds for each setup. The GSM8K training set was split 90/10 into an RL partition and an SFT partition. In the SFT + RL conditions, the base model was first fine-tuned via supervised learning on the SFT partition (10\% of training data, ~747 examples) using the full chain-of-thought reasoning traces from GSM8K official dataset for 20, 60, 100, or 300 steps (lr = 2e-4, batch size 8), and the resulting checkpoint was then used to initialize GRPO training. In the RL-only baseline, GRPO training was applied directly to the base model with no SFF a priori. All GRPO runs trained for 1000 steps on the 90\% RL partition (lr = 2e-5, batch size 8, 8 generations per prompt, max completion length 768 tokens), using a binary accuracy reward over the GSM8K test format. All models used 4-bit NF4 quantization with LoRA adapters (rank 32) applied to all attention and MLP projection layers. Evaluation reward was computed every 50 steps on a held-out subset of 128 GSM8K test split examples.

Both SFT and RL training used the same plain text prompt format applied directly to the base model. Each prompt consisted of a brief instruction:
\begin{framed}
"Solve the math problem. The last line must be exactly: Final answer: $\text{\\boxed} \{\text{NUMBER}\}$", 
\end{framed}
followed by the problem under a "Problem:" header and a "Solution:" prefix that cued the model to begin its response.

We used this format to work with a raw base model rather than an instruction-tuned one, providing just enough structure to elicit a parseable answer while keeping the format simple. We verified that before any finetuning, the model can often generate coherent answers with the correct format, but wrong answers. 

For SFT, the full target sequence was appended. We used the GSM8K dataset's full answer trace, for example, 

\begin{framed}
Natalia sold 48/2 = <<48/2=24>>24 clips in May.

Natalia sold 48+24 = <<48+24=72>>72 clips altogether in April and May.
\end{framed}
followed by the formatted final answer. In this way, we expect the model learned to produce step-by-step solutions ending in the required boxed format. 

For GRPO, only the prompt portion was provided and the model generated completions freely, with a binary accuracy reward assigned based on whether the final boxed answer matched the ground truth.

\section{Compute Resources}
Only experiments that required a considerable compute resources was LLM experiment with OlMO 1B model and GSM8k dataset in~\cref{fig:3}b.
All of the LLM experiments were run on a single NVIDIA H100 GPU. A RL run with our setup took ~8 hours, whereas SFT completed in under 1 hour.
\section{LLM Usage}
We used LLM assistance during manuscript preparation to improve clarity and consistency. In particular, LLMs were used to help reorganize mathematical notation, double-check the presentation of proofs, and improve the manuscript for grammar, style, and readability. 
\newpage
\section*{NeurIPS Paper Checklist}

\begin{enumerate}

\item {\bf Claims}
    \item[] Question: Do the main claims made in the abstract and introduction accurately reflect the paper's contributions and scope?
    \item[] Answer: \answerYes{} % Replace by \answerYes{}, \answerNo{}, or \answerNA{}.
    \item[] Justification: Our abstract summarizes the contribution of our work: a simple analytically tractable model of multi-stage training.

\item {\bf Limitations}
    \item[] Question: Does the paper discuss the limitations of the work performed by the authors?
    \item[] Answer: \answerYes{} % Replace by \answerYes{}, \answerNo{}, or \answerNA{}.
    \item[] Justification: We have a dedicated limitation section at the end of the Discussion.

\item {\bf Theory assumptions and proofs}
    \item[] Question: For each theoretical result, does the paper provide the full set of assumptions and a complete (and correct) proof?
    \item[] Answer: \answerYes{} % Replace by \answerYes{}, \answerNo{}, or \answerNA{}.
    \item[] Justification: We provide a theory, its derivation steps and the simulation which directly tests the correctness of the theory.

    \item {\bf Experimental result reproducibility}
    \item[] Question: Does the paper fully disclose all the information needed to reproduce the main experimental results of the paper to the extent that it affects the main claims and/or conclusions of the paper (regardless of whether the code and data are provided or not)?
    \item[] Answer: \answerYes{} % Replace by \answerYes{}, \answerNo{}, or \answerNA{}.
    \item[] Justification: We provide the experimental details and hyperparameters used for generating the results. The code for the experiments will be publicly available upon acceptance.

\item {\bf Open access to data and code}
    \item[] Question: Does the paper provide open access to the data and code, with sufficient instructions to faithfully reproduce the main experimental results, as described in supplemental material?
    \item[] Answer: \answerNo{} % Replace by \answerYes{}, \answerNo{}, or \answerNA{}.
    \item[] Justification: We provide the experimental details and hyperparameters used for generating the results. The code for the experiments will be publicly available upon acceptance.

\item {\bf Experimental setting/details}
    \item[] Question: Does the paper specify all the training and test details (e.g., data splits, hyperparameters, how they were chosen, type of optimizer) necessary to understand the results?
    \item[] Answer: \answerYes{} % Replace by \answerYes{}, \answerNo{}, or \answerNA{}.
    \item[] Justification: We include the details of the hyperparameters for all the results we include in the paper.

\item {\bf Experiment statistical significance}
    \item[] Question: Does the paper report error bars suitably and correctly defined or other appropriate information about the statistical significance of the experiments?
    \item[] Answer: \answerYes{} % Replace by \answerYes{}, \answerNo{}, or \answerNA{}.
    \item[] Justification: Most of our experiment is for the validation of the theory that provides exact learning dynamics. For one experiment with LLM finetuning to compare the characteristics of learning curves, we used 3 seeds and notated mean and standard deviation. 

\item {\bf Experiments compute resources}
    \item[] Question: For each experiment, does the paper provide sufficient information on the computer resources (type of compute workers, memory, time of execution) needed to reproduce the experiments?
    \item[] Answer: \answerYes{}  % Replace by \answerYes{}, \answerNo{}, or \answerNA{}.
    \item[] Justification: We provide additional section in appendix for compute resources providing the relevant details for the experiments that required considerable compute. 
    
\item {\bf Code of ethics}
    \item[] Question: Does the research conducted in the paper conform, in every respect, with the NeurIPS Code of Ethics \url{https://neurips.cc/public/EthicsGuidelines}?
    \item[] Answer: \answerYes{} % Replace by \answerYes{}, \answerNo{}, or \answerNA{}.
    \item[] Justification: We follow the Code of Ethics.

\item {\bf Broader impacts}
    \item[] Question: Does the paper discuss both potential positive societal impacts and negative societal impacts of the work performed?
    \item[] Answer: \answerNA{} % Replace by \answerYes{}, \answerNo{}, or \answerNA{}.
    \item[] Justification: The goal of this work is to advance the theoretical understanding of the foundation model with a complex training pipeline. We do not foresee any direct societal impacts of the work.
    
\item {\bf Safeguards}
    \item[] Question: Does the paper describe safeguards that have been put in place for responsible release of data or models that have a high risk for misuse (e.g., pre-trained language models, image generators, or scraped datasets)?
    \item[] Answer: \answerNA{} % Replace by \answerYes{}, \answerNo{}, or \answerNA{}.
    \item[] Justification: N/A

\item {\bf Licenses for existing assets}
    \item[] Question: Are the creators or original owners of assets (e.g., code, data, models), used in the paper, properly credited and are the license and terms of use explicitly mentioned and properly respected?
    \item[] Answer: \answerYes{} % Replace by \answerYes{}, \answerNo{}, or \answerNA{}.
    \item[] Justification: We used OLMo-2 1B and GSM8K and cited them.

\item {\bf New assets}
    \item[] Question: Are new assets introduced in the paper well documented and is the documentation provided alongside the assets?
    \item[] Answer: \answerNA{} % Replace by \answerYes{}, \answerNo{}, or \answerNA{}.
    \item[] Justification: NA

\item {\bf Crowdsourcing and research with human subjects}
    \item[] Question: For crowdsourcing experiments and research with human subjects, does the paper include the full text of instructions given to participants and screenshots, if applicable, as well as details about compensation (if any)? 
    \item[] Answer: \answerNA{} % Replace by \answerYes{}, \answerNo{}, or \answerNA{}.
    \item[] Justification: NA

\item {\bf Institutional review board (IRB) approvals or equivalent for research with human subjects}
    \item[] Question: Does the paper describe potential risks incurred by study participants, whether such risks were disclosed to the subjects, and whether Institutional Review Board (IRB) approvals (or an equivalent approval/review based on the requirements of your country or institution) were obtained?
    \item[] Answer: \answerNA{} % Replace by \answerYes{}, \answerNo{}, or \answerNA{}.
    \item[] Justification: NA

\item {\bf Declaration of LLM usage}
    \item[] Question: Does the paper describe the usage of LLMs if it is an important, original, or non-standard component of the core methods in this research? Note that if the LLM is used only for writing, editing, or formatting purposes and does \emph{not} impact the core methodology, scientific rigor, or originality of the research, declaration is not required.
    %this research? 
    \item[] Answer: \answerYes{} % Replace by \answerYes{}, \answerNo{}, or \answerNA{}.
    \item[] Justification: We included section of LLM usage declaration in the appendix.

\end{enumerate}

\end{document}